\documentclass[10pt,twocolumn,letterpaper]{article}

\usepackage[pagenumbers]{cvpr} % To force page numbers, e.g. for an arXiv version

\usepackage{graphicx}
\usepackage{amsmath}
\usepackage{amssymb}
\usepackage{amsmath,bm}
\usepackage{booktabs}
\usepackage{amsfonts}
\usepackage{float}
\usepackage{multirow}
\usepackage{stfloats}
\usepackage{appendix}

\usepackage[pagebackref,breaklinks,colorlinks]{hyperref}

\usepackage[capitalize]{cleveref}
\crefname{section}{Sec.}{Secs.}
\Crefname{section}{Section}{Sections}
\Crefname{table}{Table}{Tables}
\crefname{table}{Tab.}{Tabs.}

\def\cvprPaperID{10370} % *** Enter the CVPR Paper ID here
\def\confName{CVPR}
\def\confYear{2023}

\begin{document}

\title{ProxyFormer: Proxy Alignment Assisted Point Cloud Completion with Missing Part Sensitive Transformer}

\author{Shanshan Li, ~Pan Gao\thanks{Corresponding author.}, ~Xiaoyang Tan, ~Mingqiang Wei\\
College of Computer Science and Technology, Nanjing University of Aeronautics and Astronautics\\
{\tt\small \{markli, pan.gao, x.tan, mqwei\}@nuaa.edu.cn} \\
}
\maketitle

% Due to the limited resolution of 3D scanning equipment such as Lidar, the mutual occlusion between targets, and the transparency of the target surface material, the collected 3D point cloud data is often incomplete. 

%%%%%%%%% ABSTRACT
\begin{abstract}
    Problems such as equipment defects or limited viewpoints will lead the captured point clouds to be incomplete. Therefore, recovering the complete point clouds from the partial ones plays an vital role in many practical tasks, and one of the keys lies in the prediction of the missing part. In this paper, we propose a novel point cloud completion approach namely ProxyFormer that divides point clouds into existing (input) and missing (to be predicted) parts and each part communicates information through its proxies. Specifically, we fuse information into point proxy via feature and position extractor, and generate features for missing point proxies from the features of existing point proxies. Then, in order to better perceive the position of missing points, we design a missing part sensitive transformer, which converts random normal distribution into reasonable position information, and uses proxy alignment  to refine the missing proxies. It makes the predicted point proxies more sensitive to the features and positions of the missing part, and thus make these proxies more suitable for subsequent coarse-to-fine processes. Experimental results show that our method outperforms state-of-the-art completion networks on several benchmark datasets and has the fastest inference speed. Code is available at \url{https://github.com/I2-Multimedia-Lab/ProxyFormer}.
\end{abstract}

\vspace{-0.3cm}
%%%%%%%%% BODY TEXT
\section{Introduction}
\label{sec:intro}

3D data is used in many different fields, including autonomous driving, robotics, remote sensing, and more \cite{chen2022pseudo,zhang2022adversarial,james2022coarse,liu2022cdgnet,li2022exploiting}. Point cloud has a very uniform structure, which avoids the irregularity and complexity of composition. However, in practical applications, due to the occlusion of objects, the difference in the reflectivity of the target surface material, and the limitation of the resolution and viewing angle of the visual sensor, the collected point cloud data is often incomplete. The resultant missing geometric and semantic information will affect the subsequent 3D tasks \cite{uddin2022incomplete}. Therefore, how to use a limited amount of incomplete data to complete point cloud and restore the original shape has become a hot research topic, and is of great significance to downstream tasks \cite{wu2020multimodal,chen2022pointtree,chen2022imlovenet,hou2022hitpr,xu2022bico,liu2022pvnas}. 

%     \begin{figure}[t]
% 		\centering
% 		%\setlength{\abovecaptionskip}{-0.03cm}
% 		\includegraphics[width=8cm]{pictures/Main_Idea.pdf}\\
% 		\caption{\textbf{Illustration of our main idea.} \emph{ProxyFormer} is used for the generation of missing parts of point clouds. There are two types of proxies: the existing proxies (\emph{EP}) and the missing proxies (\emph{MP}). The \emph{MP} in turn contains \emph{true-MP} from true missing point cloud and \emph{pre_MP} predicted by transformer.}
% 		\label{Fig.1.}
% 	\end{figure}
	
	\begin{figure}[t]
		\centering
		\includegraphics[width=8.5cm]{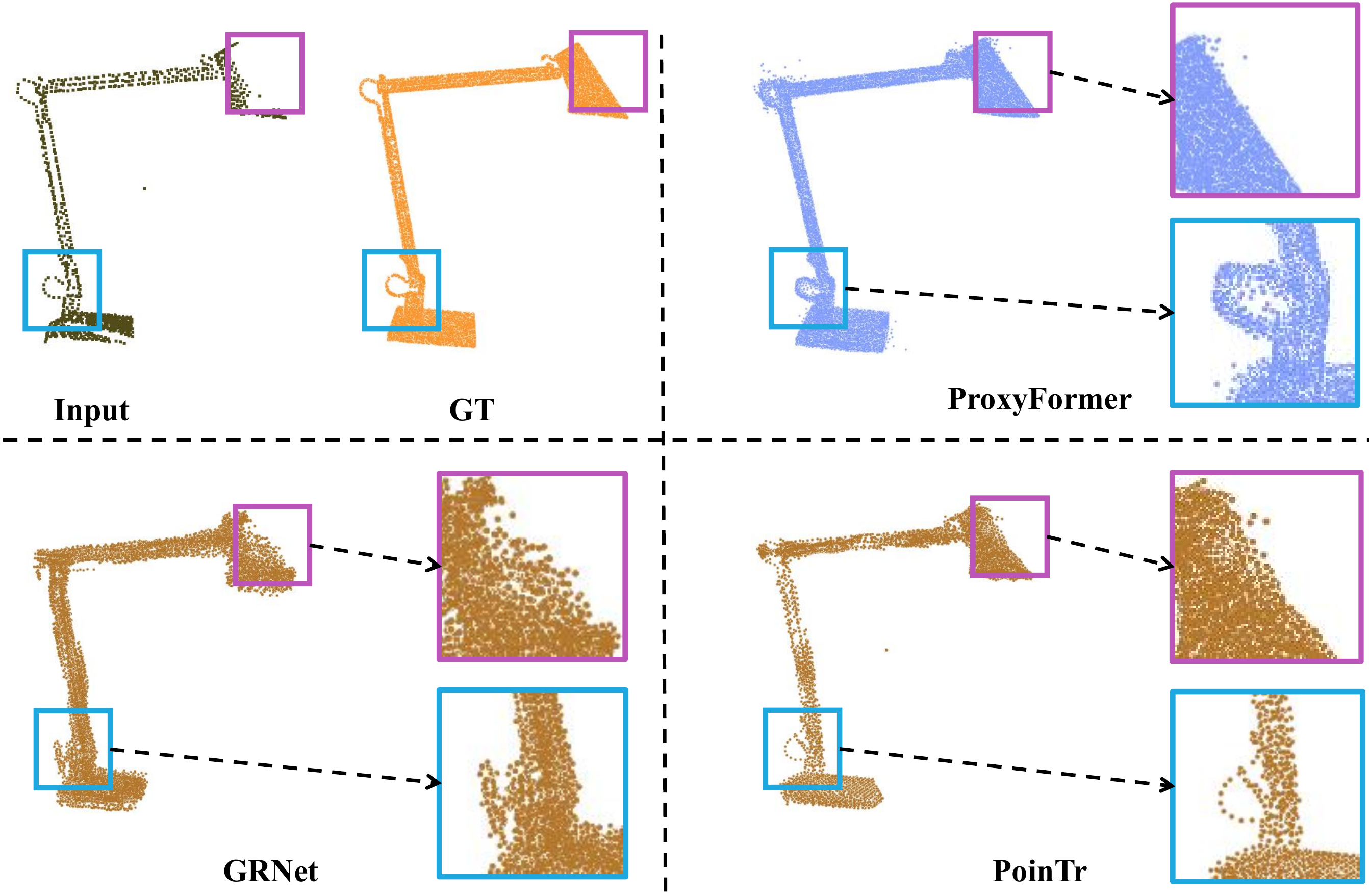}\\
		\caption{\textbf{Visual comparison of point cloud completion results.} Compared with GRNet \cite{xie2020grnet} and PoinTr \cite{yu2021pointr}. \emph{ProxyFormer} completely retains the partial input (blue bounding box) and restores the missing part with details (purple bounding box).}
		\label{main_Fig.1.}
	\end{figure}
	
With the tremendous success of PointNet \cite{qi2017pointnet} and PointNet++ \cite{qi2017pointnet++}, direct processing of 3D coordinates has become the mainstream of point cloud analysis. In recent years, there have been many point cloud completion methods \cite{achlioptas2018learning,yuan2018pcn,huang2020pf,xie2020grnet,yu2021pointr,xiang2021snowflakenet,zhou2022seedformer}, and the emergence of these networks has also greatly promoted the development of this area. Many methods \cite{achlioptas2018learning,yuan2018pcn,xie2020grnet} adopt the common encoder-decoder structure, which usually get global feature from the incomplete input by pooling operation and map this feature back to the point space to obtain a complete one. This kind of feature can predict the approximate shape of the complete point cloud. However, there are two drawbacks: (1) The global feature is extracted from partial input and thus lack the ability to represent the details of the missing part; (2) These methods discard the original incomplete point cloud and regenerate the complete shape after extracting features, which will cause the shape of the original part to deform to a certain extent. Methods like \cite{huang2020pf,yu2021pointr} attempt to predict the missing part separately, but they do not consider the feature connection between the existing  and the missing parts, which  are still not good solutions to the first drawback. The results of GRNet \cite{xie2020grnet} and PoinTr \cite{yu2021pointr} in Fig. \ref{main_Fig.1.} illustrate the existence of these problems. GRNet failed to keep the ring on the light stand while PoinTr incorrectly predicted the straight edge of the lampshade as a curved edge. Besides, some methods \cite{yu2021pointr,xiang2021snowflakenet,lin2021pctma,zhou2022seedformer} are based on the transformer structure and use the attention mechanism for feature correlation calculation. However, this also brings up two other problems: (3) In addition to the feature, the position encoding also has a great influence on the effect of the transformer. Existing transformer-based methods, either directly using 3D coordinates \cite{guo2021pct,zhou2022seedformer} or using MLP to upscale the coordinates \cite{yu2021pointr,xiang2021snowflakenet}, the position information of the point cloud cannot be well represented; (4) It also leads to the problem of excessive parameters or calculation. Furthermore, we also note that most of the current supervised methods do not make full use of the known data. During the training process, the point cloud data we can obtain includes incomplete input and Ground Truth (GT). This pair of data can indeed undertake the point cloud completion task well, but in fact, we can obtain a new data through these two data, that is, the missing part of the point cloud, so as to increase our prior knowledge.
	
In order to solve the above-mentioned problems, we propose a novel point cloud completion network named \emph{ProxyFormer}, which completely preserves the incomplete input and has better detail recovery capability as shown in Fig. \ref{main_Fig.1.}. Firstly, we design a feature and position extractor to convert the point cloud to proxies, with a particular attention to the representation of point position. Then, we let the proxies of the partial input interact with the generated missing part proxies through a newly proposed missing part sensitive transformer, instead of using the global feature extracted from incomplete input alone as in prior methods. After mapping proxies back to the point  space, we splice it with the incomplete input points to 100\%  preserve  the original data. During training, we use the true missing part of the point cloud to increase prior knowledge and for prediction error refinement. Overall, the main contributions of our work are as follows:
\vspace{-0.2cm}
\begin{itemize}
    \item We design a Missing Part Sensitive Transformer, which focuses on the geometry structure and details of the missing part. We also propose a new position encoding method that aggregates both the coordinates and features from neighboring points.
    \vspace{-0.2cm}
    \item We introduce Proxy Alignment into the training process. We convert the true missing part into proxies, which are used to enhance the prior knowledge while refining the predicted missing proxies.
    \vspace{-0.2cm}
    \item Our proposed method \emph{ProxyFormer} discards the transformer decoder adopted in most transformer based completion methods such as PointTr, which achieves SOTA performance compared to various baselines while having considerably few parameters and the fastest inference speed in terms of GFLOPs.
\end{itemize}

\section{Related Work}
\label{sec:related work}

\noindent{\bfseries 3D shape completion.} Traditional shape completion work mainly includes two categories: geometric rule completion \cite{zhao2007robust,pauly2008discovering,mitra2013symmetry} and template matching completion \cite{nan2012search,kim2013learning,li2015database}. However, these methods require the input to be as complete as possible, and thus are not robust to new objects and environmental noise. VoxelNet \cite{zhou2018voxelnet} attempts to divide the point cloud into voxel grids and applies convolutional neural networks, but the voxelization will lose the details of the point cloud, and the increasing resolution of the voxel grid will significantly increase the memory consumption. Yuan \emph{et al.} \cite{yuan2018pcn} designed PCN, which proposed a coarse-to-fine method based on the PointNet \cite{qi2017pointnet} and FoldingNet \cite{yang2018foldingnet}, but its decoder often fails to recover rare geometries of objects such as seat backs with gaps, \emph{etc.} Therefore, after PCN, many other methods \cite{tchapmi2019topnet,huang2020pf,wang2020cascaded,xia2021asfm} focus on multi-step point cloud generation, which is helpful to recover a final point cloud with fine-grained details. Furthermore, following DGCNN \cite{wang2019dynamic}, some researchers developed graph-based methods \cite{wu2021point,zhu2021towards,wu2021lra} which consider regional geometric details. Although these methods provided better feature extractors and decoders, none of them considered the feature connection between the incomplete input and the missing part, which affects the quality of the completion result. Our proposed \emph{ProxyFormer} is not limited to the partial input but also incorporates true missing points during training. We generate features separately for the  missing part and explore the correlation with the features extracted from the partial input via self-attention.

\noindent{\bfseries Transformers.} The transformer structure originated in the field of natural language processing, which is proposed by Vaswani \emph{et al.} \cite{vaswani2017attention} and applied to machine translation tasks. Recently, this structure was introduced into point cloud processing tasks due to its advantage in extracting correlated features between points. Guo \emph{et al.} \cite{guo2021pct} proposed PCT and optimized the self-attention module, making the transformer structure more suitable for point cloud learning, and achieved good performance in shape classification and part segmentation. Point Transformer \cite{zhao2021point} designs a vector attention for point cloud feature processing. PointTr \cite{yu2021pointr} and SeedFormer \cite{zhou2022seedformer} treat the point cloud completion as a set-to-set translation problem that share similar ideas as \emph{ProxyFormer}. PoinTr designs a geometry-aware block that explicitly simulates local geometric relations to facilitate transformers to use better inductive bias. However, it adopts a transformer encoder-decoder structure for point cloud completion, which results in a large amount of parameters. SeedFormer designs an upsample transformer by extending the transformer structure into a basic operation in point generators that effectively incorporates spatial and semantic relationships between neighboring points. However, the upsample transformer runs throughout its network, resulting in excessive computation. Differently, \emph{ProxyFormer} discards the transformer decoder to reduce the number of parameters, and modifies the query of transformer to make it more suitable for the prediction of the missing part. In the coarse-to-fine process, we still adopt the Foldingnet \cite{yang2018foldingnet}, which greatly reduces the amount of calculation.

\section{Method}
\label{sec:method}

%     \begin{figure*}
% 		\centering
%         \includegraphics[width=17.3cm]{pictures/Network_Architecture.pdf}\\
% 		\caption{The pipeline of \emph{ProxyFormer} is shown in the upper part. The completion of the point cloud is divided into two steps. First, we simply convert the incomplete seed feature into a predicted coarse missing part. Second, we send the predicted missing proxies and the coarse part into FoldingNet \cite{yang2018foldingnet} to obtain the predicted dense missing part. True missing part point cloud is used for training only so that we block its backpropagation and directly employ the pretrained FAPE for the incomplete point cloud to it. (a) The feature and position extractor is applied to obtain seed feature and position encoding which are combined to proxies. $P$ represents the count of points in the input point cloud. $C_1$ and $C_2$ is the dimensions of point cloud features. (b) The missing feature generator is used to generate predicted seed feature from incomplete seed feature. $N$ and $M$ means the count of the incomplete seed feature and predicted seed feature. $C$ means the dimensions of the seed feature and is divided into $U$ groups to speed up operations.}
% 		\label{Fig.2.}
% 	\end{figure*}

    \begin{figure*}
      \centering
      \includegraphics[width=17.3cm]{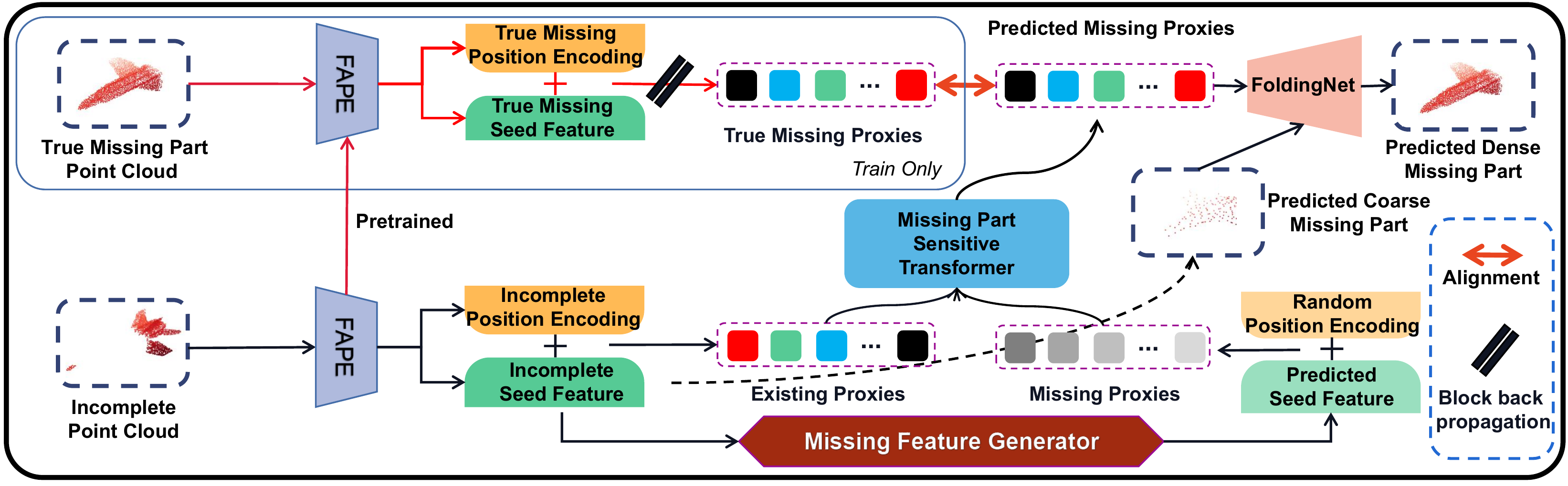}\\
      \quad
      \begin{subfigure}{0.48\linewidth}
        \includegraphics[width=8.5cm]{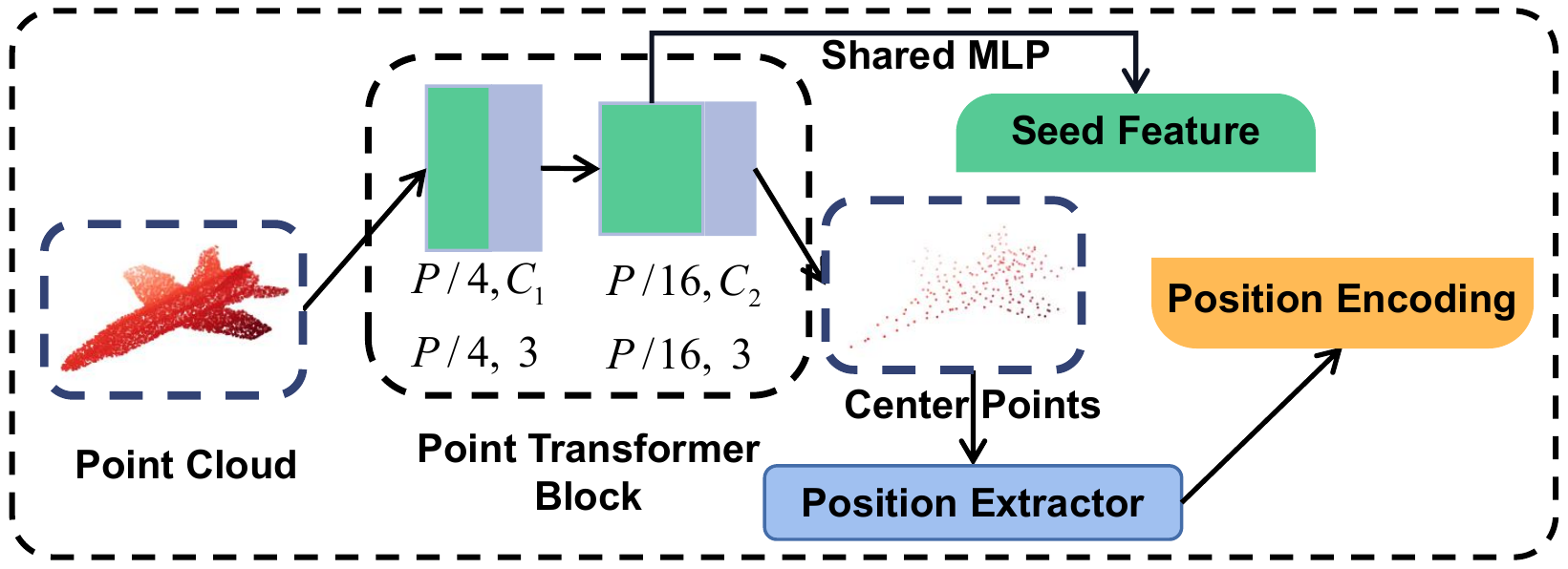}
        \caption{Feature And Position Extractor (FAPE).}
        \label{main_Fig.2-a.}
      \end{subfigure}
      \hfill
      \begin{subfigure}{0.48\linewidth}
        \includegraphics[width=8cm]{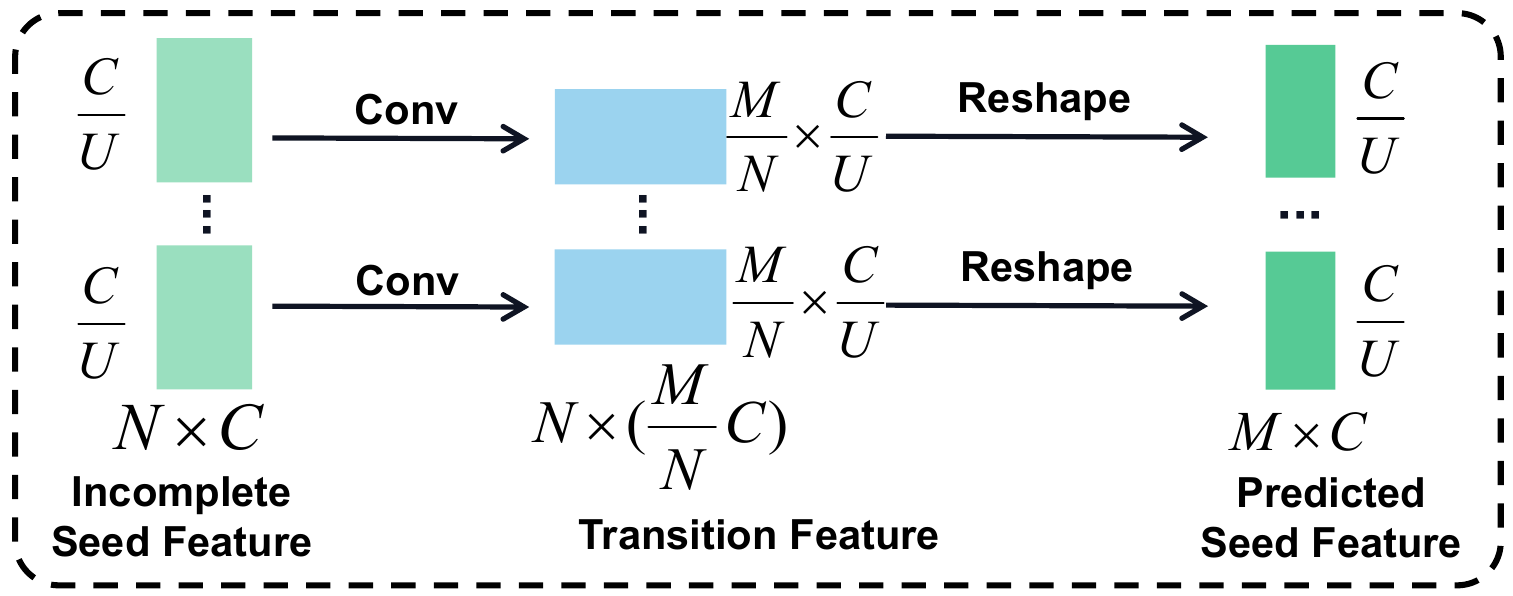}
        \caption{Missing Feature Generator.}
        \label{main_Fig.2-b.}
      \end{subfigure}
      \caption{The pipeline of \emph{ProxyFormer} is shown in the upper part. The completion of the point cloud is divided into two steps. First, we simply convert the incomplete seed feature into a predicted coarse missing part. Second, we send the predicted missing proxies and the coarse part into FoldingNet \cite{yang2018foldingnet} to obtain the predicted dense missing part. True missing part  is used for training only so that we block its back-propagation and directly employ the pretrained FAPE on the incomplete point cloud to generate the proxy. (a) The feature and position extractor is applied to obtain seed feature and position encoding which are combined to the so-called proxies. $P$ represents the count of points in the input point cloud. $C_1$ and $C_2$ is the dimensions of point cloud features. (b) The missing feature generator is used to generate predicted seed feature from incomplete seed feature. $N$ and $M$ means the point count of the incomplete seed feature and predicted seed feature. $C$ means the dimensions of the seed feature and is divided into $U$ groups to speed up operations.}
      \label{main_Fig.2.}
    \end{figure*}

% As introduced in Section \ref{sec:intro}, the structure mainly includes four parts: (1). feature and position extractor (FAPE), which is used to extract the feature and position of point cloud; (2). missing feature generator (Missing Feature Generator), which is used to generate missing part feature. (3). missing part sensitive transformer (Missing Part Sensitive Transformer), used to predict missing part proxies. (4). proxy alignment (PA), used to correct predicted missing part proxies. 

% We first input the incomplete point cloud and missing part point cloud into \emph{FAPE} to get \emph{EP} and \emph{true-MP}. At the same time, the incomplete seed feature of \emph{EP} is used to predict coarse missing part as well as to convert by missing feature generator to predicted seed feature. The predicted feature is added with random position distribution to form \emph{MP}. Secondly, \emph{Missing Part Sensitive Transformer} (Sec. \ref{sec:MPST}) is used to receive \emph{EP} and \emph{MP} to get \emph{pre-MP}. After alignment (Sec. \ref{sec:PA}) with \emph{true-MP}, the \emph{pre-MP} is sent to FoldingNet \cite{yang2018foldingnet} to complete the coarse-to-dense quest for the predicted coarse missing part. 
The overall network structure of \emph{ProxyFormer} is shown in Fig. \ref{main_Fig.2.}. We will introduce our method in detail as follows.

%     \begin{figure}[h]
% 		\centering
% 		%\setlength{\abovecaptionskip}{-0.03cm}
% 		\includegraphics[width=8cm]{pictures/FAPE.pdf}\\
% 		\caption{Datails of \emph{FAPE}. It uses point transformer block for downsampling and feature extraction, and uses a newly designed point cloud position extractor for position extraction.}
% 		\label{Fig.3.}
% 	\end{figure}

\subsection{Proxy Formation}
\label{sec:PF}

\noindent{\bfseries Proxy introduction.} A proxy represents a local region of the point clouds. All the proxies in this paper fuse two information: \textbf{feature} and \textbf{position}. The types of proxies are defined as follows:

\begin{itemize}
\vspace{-0.2cm}
  \item \textbf{Existing Proxies} (\emph{EP}): It combines incomplete seed feature and incomplete position encoding. (obtained by FAPE).
  \vspace{-0.2cm}
  \item \textbf{Missing Proxies} (\emph{MP}): It combines predicted seed feature and random position encoding. During the training process, \emph{MP} is also divided into:
  \begin{itemize}
  \vspace{-0.2cm}
      \item[-] \textbf{Predicted Missing Proxies} (\emph{pre-MP}): It is obtained by Missing Part Sensitive Transformer (Sec. \ref{sec:MPST}).
      \vspace{-0.15cm}
      \item[-] \textbf{True Missing Proxies} (\emph{true-MP}): It combines true missing seed feature and true missing position encoding. (obtained by pre-trained FAPE). It is only used for Proxy Alignment (Sec. \ref{sec:PA}).
  \end{itemize}
\end{itemize}

For clarity, we next explain how to obtain the information these proxies need.

%we try to enhance the connection between points while downsampling.
% we found that directly concatenate the point coordinates with extracted features \cite{qi2017pointnet++,zhao2019pointweb,wang2019dynamic} or simply uses MLP to upgrade the three-dimensional coordinates \cite{yu2021pointr} are ineffective. So 

\noindent{\bfseries Feature and position extractor (FAPE).} For \textbf{feature} extraction, as shown in Fig. \ref{main_Fig.2-a.}, the point cloud of dimension $\left(P, 3\right)$ is sent to point transformer block \cite{zhao2021point}, and the center point cloud of $\left(\frac{P}{16}, 3\right)$ is obtained by farthest point sampling twice. The feature of $\left(\frac{P}{16}, C_2\right)$ is obtained through two vector attention calculations \cite{zhao2021point}. After that, we use a shared MLP to convert the feature to final seed feature.

    \begin{figure}[t]
		\centering
		\includegraphics[width=8.5cm]{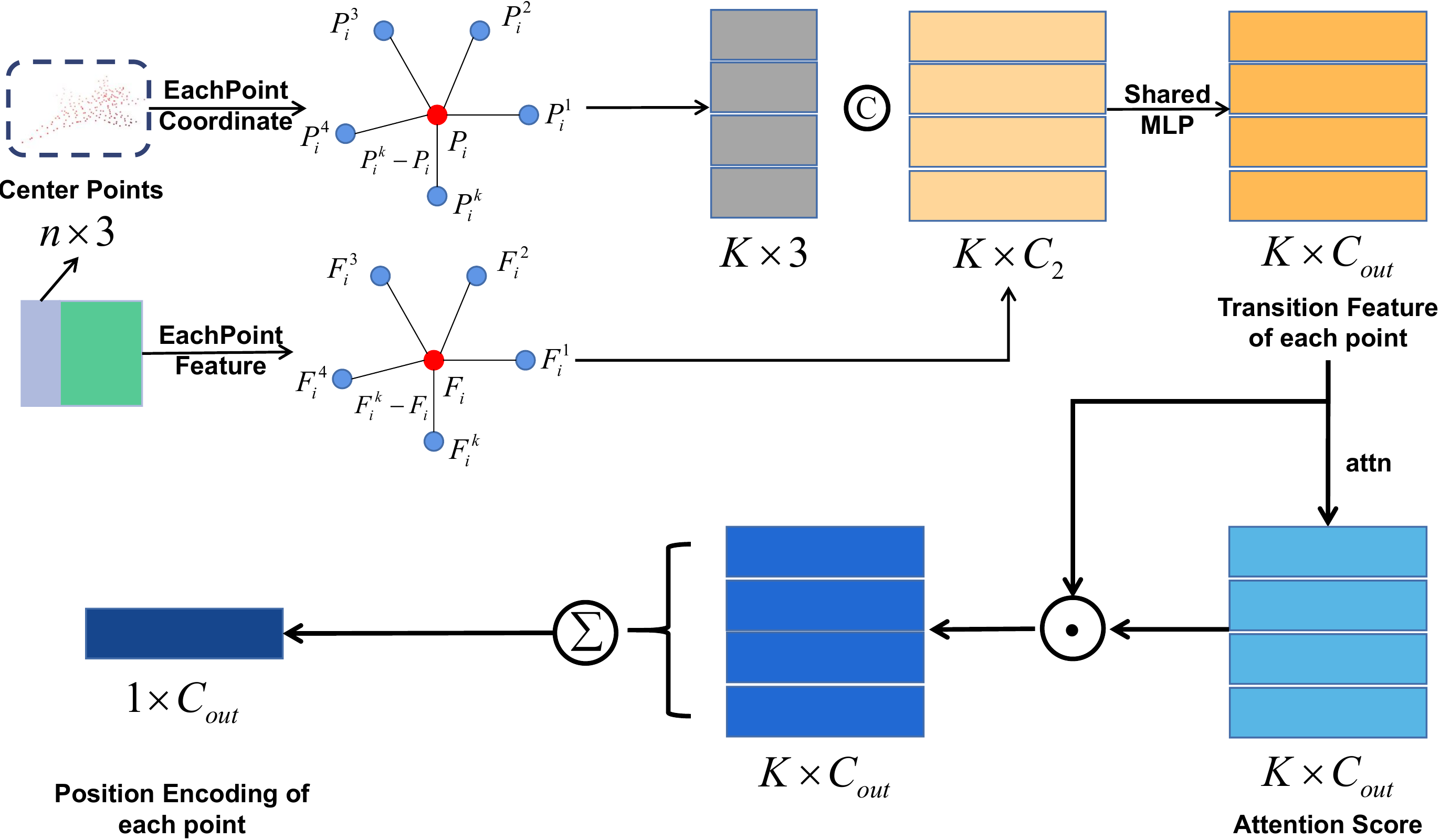}\\
		\caption{Details of position extractor. $n = \frac{P}{16}$. $K$ is the count of neighbor points. $C_2$ and $C_{out}$ are the dimension of point features. $Shared~MLP$ means shared multi-layer perceptron and $attn$ means attention score calculation.}
		\label{main_Fig.3.}
	\end{figure}

For \textbf{position} encoding, we found that directly concatenating the point coordinates with extracted features \cite{qi2017pointnet++,zhao2019pointweb,wang2019dynamic} or simply using MLP to upgrade the three-dimensional coordinates \cite{yu2021pointr} are ineffective. So we design a new position extractor (as shown in Fig. \ref{main_Fig.3.}) to improve this. The coordinates and features of the center points after feature extraction are used as input. For each point, we take its adjacent $K$ points, and use $\tilde{p}_i^k=p_i^k-p_i$ to calculate the relative position of the point. $p_i^k\in{\mathbb{P}=\left\{p_i^1,p_i^2,...,p_i^K\right\}}$, which means the neighbor point coordinates of $p_i$. We also perform neighbor points subtraction in the feature dimension using $\tilde{f}_i^k=|f_i^k-f_i|$. Similarly, $f_i^k\in{\mathbb{F}=\left\{f_i^1,f_i^2,...,f_i^K\right\}}$, which means the neighbor point features of $p_i$. After that, we get the coordinate information of $K \times 3$ and the feature information of $K \times C_2$. Then we concatenate them and transform feature from $K \times (3+C_2)$ to $K \times C_{out}$ (the transition feature ${TF}_i^k$ for each neighboring point) using a shared MLP. After obtaining ${TF}_i^k$, we use attention mechanism to learn a unique attention score for each channel of point features and then aggregate them. The attention score is calculated and channel-wise multiplied with the feature and summed to obtain the final $PE$ (\emph{i.e.} position encoding) of each point. This process can be represented by Eq. (\ref{main_eq.1.}).

    \begin{equation}
        \begin{aligned} PE=\sum_{k=1}^K\left(attn\left(\left\{{TF}_i^k\right\}\right)\cdot\left\{{TF}_i^k\right\}\right)
        \end{aligned} ~ ,
        \label{main_eq.1.}
    \end{equation}
where $\left\{{TF}_i^k\right\}$ is the set of transition feature of $K$-neighbor points and $attn()$ is a shared function (per-point MLPs) with learnable weights $W_{attn}$.

Incomplete point cloud and true missing part point cloud are sent into FAPE to get \emph{EP} and \emph{true-MP}, and the imcomplete seed feature of \emph{EP} is used to predict coarse missing part at the same time.

\subsection{Missing Part Sensitive Transformer}
\label{sec:MPST}

\begin{figure}[h]
		\centering
		\includegraphics[width=8cm]{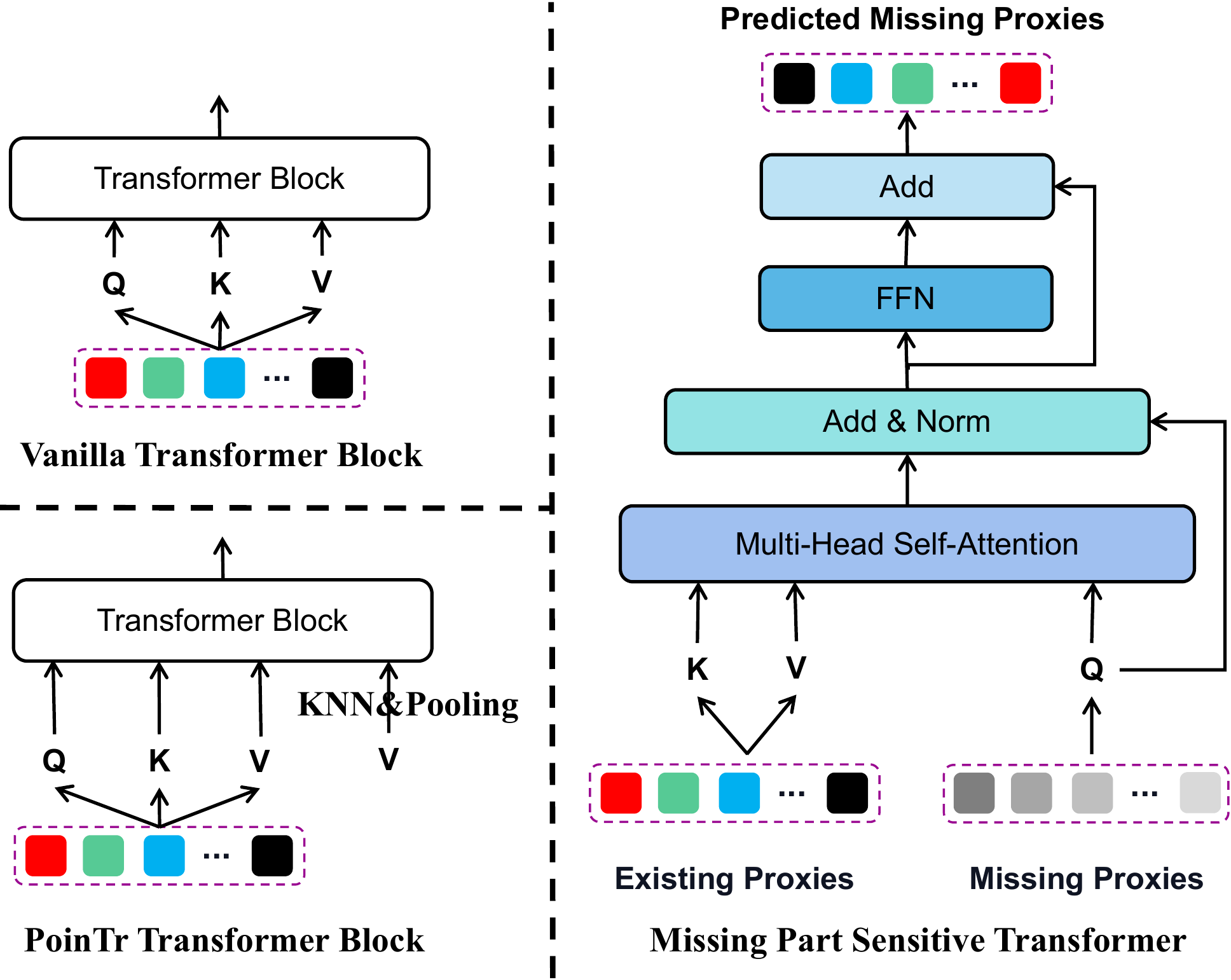}\\
		\caption{Details of Missing Part Sensitive Transformer. Compared with vanilla transformer block and PoinTr transformer block, we change the source of query to make it more suitable for the prediction of the missing part.}
		\label{main_Fig.4.}
	\end{figure}

Usually, query, key and value come from the same input. Many methods \cite{yu2021pointr,xiang2021snowflakenet,zhou2022seedformer} attempt to modify the source of value to adapt to the specific tasks  (the left part of Fig. \ref{main_Fig.4.}). Differently, we change the query source, taking the \emph{MP} with random position encoding as query conditions, to maximally mine the representation of the missing part from the features and positions of existing proxies via self-attention mechanism. 

In order to change the query source to \emph{MP}, we  propose a Missing Feature Generator, which is specially used to learn the missing part features from the existing features. The generation process is shown in Fig. \ref{main_Fig.2-b.}. Specifically, incomplete seed feature of $N \times C$ is used as input, and the $C$-dimensional channels are divided into $U$ equal length groups. Then, the change dimension of the convolution is determined by the point number $M$ of the predicted coarse missing part, which means that we convert each $\frac{C}{U}$ to $\frac{M}{N} \times \frac{C}{U}$. Lastly, we transform the transition feature into predicted seed feature of $M \times C$ through the reshape operation. All channel groups use convolutional layers with shared parameters, reducing the amount of parameters and computation. Predicted missing feature is added with random position encoding to get \emph{MP}.

In Sec. \ref{sec:PF}, we have obtained \emph{EP} and \emph{true-MP}, and through the missing feature generator  described above, we have obtained \emph{MP}. Then we design a Missing Part Sensitive Transformer to further explore their relationship and learn the representation of the missing points for subsequent completion work. Its structure is illustrated in Fig. \ref{main_Fig.4.}, which receives \emph{EP} and \emph{MP} as input and outputs \emph{pre-MP}. \emph{EP} is a matrix $\mathbb{E}$ of $N \times C$, and \emph{MP} is a matrix $\mathbb{M}$ of $M \times C$. Output \emph{pre-MP} is a matrix $\mathbb{P}$ of $M \times C$. 

We use multi-head self-attention mechanism and add residual connections to obtain \emph{pre-MP}:

    \begin{equation}
    	\begin{gathered}
    	    T=\mu \left(Q+\xi (Q,K,V)\right) ~ ,\\
    	    \emph{\text{pre-MP}}=\sigma(T)+T ~ ,
        \end{gathered}
    	 \label{main_eq.2.}
    \end{equation}
where $Q=\mathbb{M} \times W^{Q}$, $K=\mathbb{E} \times W^{KV}$, $V=\mathbb{E} \times W^{KV}$. $\mu$ means layer normalization. $\xi$ means multi-head attention calculation. $\sigma$ means feed forward network. The attention of each head is calculated with $Q_i$, $K_i$, $V_i$ on head $i$:

    \begin{equation}
    	\begin{gathered}
    	    attn_i=softmax\left(\frac{Q_i\left(K_i\right)^T}{\sqrt{d_k}}\right)V_i
        \end{gathered} ~ .
    	 \label{main_eq.3.}
    \end{equation}

This method predicts the proxies of the missing points, which not only discovers feature associations between missing and existing parts, but also converts random positions into meaningful position information. The \emph{pre-MP} is next used for proxy alignment with \emph{true-MP}.

\subsection{Proxy Alignment}
\label{sec:PA}

In this subsection, we describe the Proxy Alignment strategy and how this operation assists us in the point cloud completion task. 

The detailed computational graph of \emph{ProxyFormer} is plotted in Fig. \ref{main_Fig.8.}. Therefore, \emph{pre-MP} and \emph{true-MP} can be formulated as:

    \begin{equation}
    	\begin{gathered}
    	    \emph{\text{pre-MP}}=T\left({PE}_{R}\oplus\theta\left(\omega\left(C_i\right)\right)\right)\\
    	    \emph{\text{true-MP}}={PE}_{T}\oplus\omega\left(C_m\right)
        \end{gathered} ~ .
    	 \label{main_eq.4.}
    \end{equation}

In order to refine the prediction error in \emph{pre-MP}, the proxy alignment constraint is imposed on the model, which can be formulated as:

    \begin{equation}
        \begin{gathered}
        	l_p=MSE\left(\emph{\text{pre-MP}},\emph{\text{true-MP}}\right)
        \end{gathered} ~ ,
    	\label{main_eq.5.}
    \end{equation}
where $l_p$ means the alignment loss that we will apply to our training loss (Sec. \ref{sec:training loss}) and $MSE$ means the mean squared error. After correcting the \emph{pre-MP}, it is used as a feature and sent to FoldingNet \cite{yang2018foldingnet} for coarse-to-fine conversion, and then combined with the previously predicted coarse missing part to obtain the  dense missing part.

\subsection{Training Loss}
\label{sec:training loss}
\noindent{\bfseries Chamfer Distance.} We use the average Chamfer Distance(CD) \cite{fan2017point} as the first type of our completion loss. 

% Suppose two point clouds are represented by S1 and S2 respectively, CD can be calculated with Eq. \ref{eq.10.}.

%     \begin{small}
%         \begin{equation}
%         	\begin{gathered}
%         	    d_{\mathrm{CD}}\left(S_{1}, S_{2}\right)=\frac{1}{|S_{1}|} \sum_{x \in S_{1}} \min _{y \in S_{2}}\|x-y\|_{2}^{2}+\frac{1}{|S_{2}|} \sum_{y \in S_{2}} \min _{x \in S_{1}}\|y-x\|_{2}^{2}
%             \end{gathered}
%     	 \label{eq.10.}
%         \end{equation}
%     \end{small}

\noindent{\bfseries proxy alignment Loss.} We use the $MSE$ loss between \emph{pre-MP} and \emph{true-MP} as the second type of loss.
    
\begin{figure}[t]
    	\centering
    	\includegraphics[width=8.5cm]{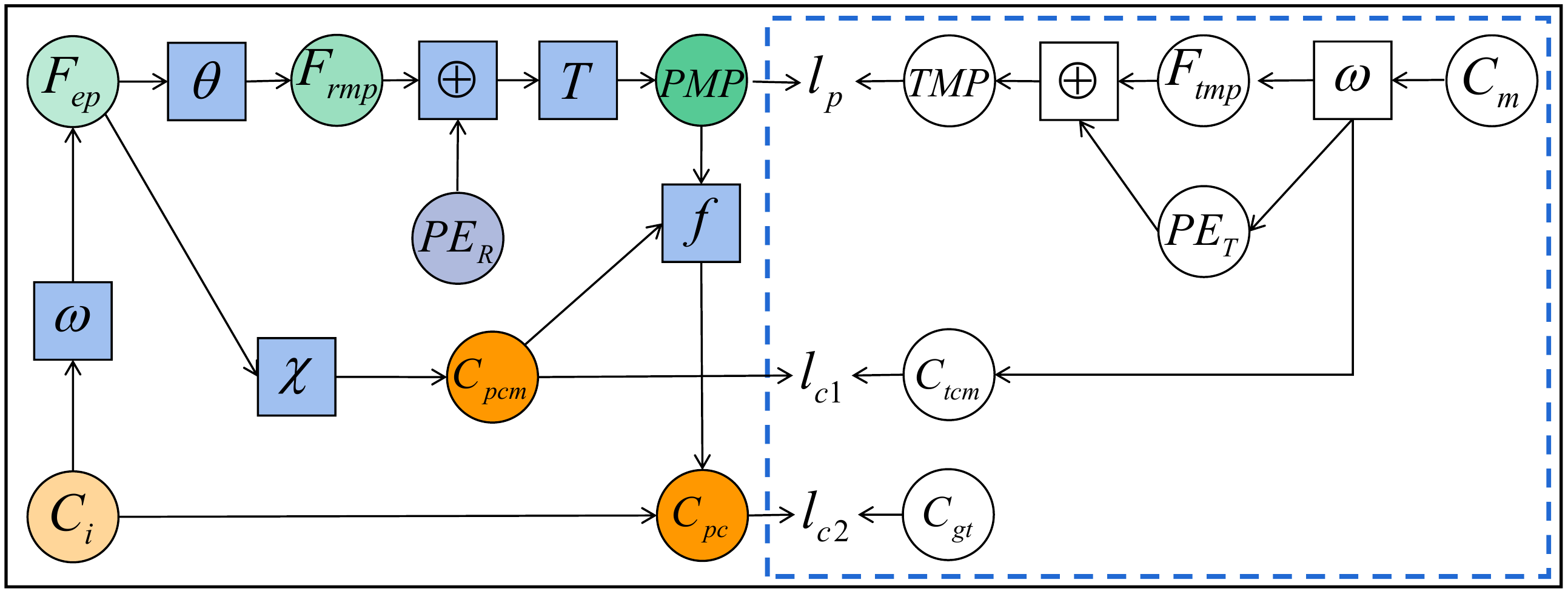}
    	\caption{The computational graph for \emph{ProxyFormer}. The part framed by the blue dotted line is used for training only. For the left part, we input incomplete point cloud $C_i$ and use \emph{FAPE} ($\omega$) to get feature $F_{ep}$ in \emph{EP}. $F_{ep}$ is not only sent to a linear projection layer $\chi$ to generate predicted coarse missing part $C_{pcm}$, but also sent to missing feature generator ($\theta$) to generate feature $F_{rmp}$. $F_{rmp}$ is added ($\oplus$) with random position distribution ${PE}_{R}$ and sent to missing part sensitive transformer ($T$) to get \emph{pre-MP}. Then \emph{pre-MP} and $C_{pcm}$ are sent to FoldingNet ($f$), and the result is spliced with input $C_i$ to form predicted complete point cloud $C_{pc}$. For the right part, we input true missing part point cloud $C_m$ and use the same \emph{FAPE} ($\omega$) to get feature $F_{tmp}$ and position ${PE}_{T}$ in \emph{true-MP}. \emph{pre-MP} is aligned with \emph{true-MP} for correcting deviation values. We also retain the center point $C_{tcm}$ obtained after $C_m$ is downsampled by \emph{FAPE}. $l_p$, $l_{c1}$ and $l_{c2}$ are the losses we use, which will be detailed in the next subsection.}
    	\label{main_Fig.8.}
    \end{figure}

To sum up, as shown in Fig. \ref{main_Fig.8.}, the loss used in this paper consists three parts: (1). $l_{c1}$, the CD between the predicted coarse missing part $C_{pcm}$ and the true center point $C_{tcm}$ of the missing part; (2). $l_{c2}$, the CD between the predicted complete point cloud $C_{pc}$ and the GT $C_{gt}$; (3) $l_{p}$, the alignment loss between \emph{pre-MP} and \emph{ture-MP}. We use the weighted sum of these three terms for network training (we set $\gamma$ to $1.5$ in experiments):

    \begin{equation}
        \begin{gathered}
        	L=l_{c1}+l_{c2}+\gamma l_p
        \end{gathered} ~ .
    	\label{main_eq.6.}
    \end{equation}

\section{Experiments}
In this section, we use \emph{ProxyFormer} for two common point cloud completion benchmarks PCN \cite{yuan2018pcn} and KITTI \cite{geiger2013vision} to evaluate the completion ability of the network, and then we also train and test on two other datasets, ShapeNet-55 and ShapeNet-34 proposed by PoinTr \cite{yu2021pointr}. Finally, through ablation experiments, we demonstrate the effectiveness of each module in the proposed \emph{ProxyFormer}.

\subsection{Point Cloud Completeion on PCN Dataset}
\noindent{\bfseries Dataset and evaluation metric.} The PCN dataset \cite{yuan2018pcn} is created from ShapeNet dataset \cite{chang2015shapenet}, including eight categories with a total of 30974 CAD models. When preparing the data, we use missing part extractor to extract the missing part of the point cloud from the complete point cloud and then downsample it to 3584 points as the true missing part (This process is described in detail in the supplementary material). 

We use the L1 CD to evaluate the results of each methods. In addition, We also use density-aware chamfer distance (DCD) \cite{wu2021density} as a quantitative evaluation criterion, which can retain the measurement ability similar to CD but also better judge the visual effect of the result. 
% The formula for DCD is as follows:

    % \begin{equation}
    %     \begin{aligned}
    %     	d_{D C D}\left(S_1, S_2\right)=\frac{1}{2}\left(\frac{1}{\left|S_1\right|} \sum_{x \in S_1}\left(1-\frac{1}{n_{\hat{y}}} e^{-\alpha\|x-\hat{y}\|_2}\right)\right)\\
    %     	+\frac{1}{2}\left(\frac{1}{\left|S_2\right|} \sum_{y \in S_2}\left(1-\frac{1}{n_{\hat{x}}} e^{-\alpha\|y-\hat{x}\|_2}\right)\right)
    %     \end{aligned}
    % 	\label{eq.13.}
    % \end{equation}

% where $\hat{y}=min_{y\in S_2}\|x-y\|_2, \hat{x}=min_{y\in S_1}\|y-x\|_2$, and $\alpha$ denotes a temperature scalar, which we set to 1000, just as the original text describes.

    \begin{figure*}[t]
		\centering
		\includegraphics[width=17cm]{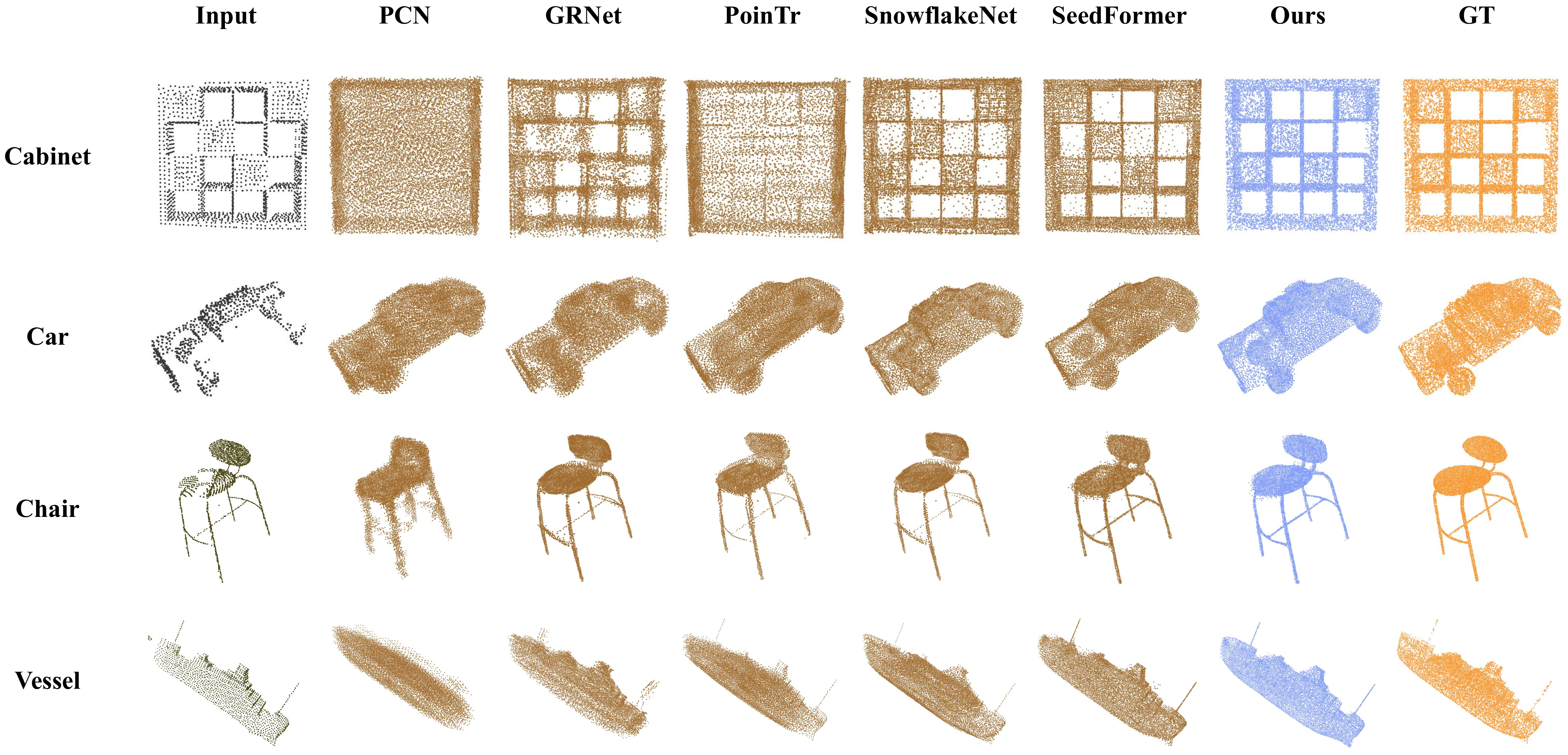}\\
		\caption{The visualization results of each method on the PCN dataset, showing Cabinet, Car, Chair and Vessel from top to bottom.}
		\label{main_Fig.9.}
	\end{figure*}

\noindent{\bfseries Quantiative comparison.} According to the results in Table \ref{main_tab.1.}, on the PCN dataset, our method has substantially surpassed PoinTr \cite{yu2021pointr} and reaches lowest CD in the cabinet, car, sofa and table categories. Further, as can be seen from the DCD shown in Table \ref{main_tab.2.}, our method outperforms state-of-the-art in all the categories, which means that our method is more able to take into account the rationality of shape and distribution density while complementing the object.

    \begin{table}[h]
		\centering
		\caption{Quantitative comparison of PCN dataset. Point resolutions for the output and GT are 16384. For CD, lower is better.}
		\scalebox{0.6}{
		\begin{tabular}{c|c|c|c|c|c|c|c|c|c}
			\cline{1-10}
			\multirow{2}{*}{Methods} & \multicolumn{9}{c}{Chamfer Distance($10^{-3}$)}
			\\
			\cline{2-10}
			& Air & Cab & Car & Cha & Lam  & Sof & Tab & Ves & Ave \\
			\cline{1-10}
			FoldingNet \cite{yang2018foldingnet} & 9.49 & 15.80 & 12.61 & 15.55 & 16.41 & 15.97  & 13.65 & 14.99 & 14.31 \\
			TopNet \cite{tchapmi2019topnet} & 7.61 & 13.31  & 10.90 & 13.82 & 14.44 & 14.78 & 11.22 & 11.12  & 12.15 \\
			AtlasNet \cite{groueix2018papier} & 6.37 & 11.94 & 10.10 & 12.06 & 12.37 & 12.99 & 10.33 & 10.61  & 10.85 \\
			PCN \cite{yuan2018pcn} & 5.50 & 22.70 & 10.63 & 8.70 & 11.00 & 11.34 & 11.68 & 8.59 & 9.64  \\
			GRNet \cite{xie2020grnet} & 6.45 & 10.37 & 9.45 & 9.41 & 7.96 & 10.51 & 8.44 & 8.04 & 8.83  \\
			CRN \cite{wang2020cascaded} & 4.79 & 9.97 & 8.31 & 9.49 & 8.94 & 10.69 & 7.81 & 8.05 & 8.51 \\
			NSFA \cite{zhang2020detail} & 4.76 & 10.18 & 8.63 & 8.53 & 7.03 & 10.53 & 7.35 & 7.48 & 8.06 \\
			PMP-Net \cite{wen2021pmp} & 5.65 & 11.24 & 9.64 & 9.51 & 6.95 & 10.83 & 8.72 & 7.25 & 8.73 \\
			PoinTr \cite{yu2021pointr} & 4.75 & 10.47 & 8.68 & 9.39 & 7.75 & 10.93 & 7.78 & 7.29 & 8.38 \\
			PMP-Net++ \cite{wen2022pmp} & 4.39 & 9.96 & 8.53 & 8.09 & 6.06 & 9.82 & 7.17 & 6.52 & 7.56 \\
			SnowflakeNet \cite{xiang2021snowflakenet} & 4.29 & 9.16 & 8.08 & 7.89 & 6.07 & 9.23 & 6.55 & 6.40 & 7.21 \\
			SeedFormer \cite{zhou2022seedformer} & \textbf{3.85} & 9.05 & 8.06 & \textbf{7.06} & \textbf{5.21} & 8.85 & 6.05 & \textbf{5.85} & \textbf{6.74} \\
			\cline{1-10}
			ProxyFormer(Ours) & 4.01 & \textbf{9.01} & \textbf{7.88} & 7.11 & 5.35 & \textbf{8.77} & \textbf{6.03} & 5.98 & 6.77 \\
			\cline{1-10}
		\end{tabular}}
		\label{main_tab.1.}
	\end{table}

    \begin{table}[h]
		\centering
		\caption{Quantitative comparison of PCN dataset. Point resolutions for the output and GT are 16384. For DCD, lower is better.}
		\scalebox{0.6}{
		\begin{tabular}{c|c|c|c|c|c|c|c|c|c}
			\cline{1-10}
			\multirow{2}{*}{Methods} & \multicolumn{9}{c}{Density-aware Chamfer Distance}
			\\
			\cline{2-10}
			& Air & Cab & Car & Cha & Lam  & Sof & Tab & Ves & Ave \\
			\cline{1-10}
			GRNet \cite{xie2020grnet} & 0.688 & 0.582 & 0.610 & 0.607 & 0.644 & 0.622 & 0.578 & 0.642 & 0.622  \\
			PoinTr \cite{yu2021pointr} & 0.574 & 0.611 & 0.630 & 0.603 & 0.628 & 0.669 & 0.556 & 0.614 & 0.611 \\
			SnowflakeNet \cite{xiang2021snowflakenet} & 0.560 & 0.597 & 0.603 & 0.582 & 0.598 & 0.633 & 0.521 & 0.583 & 0.585 \\
			SeedFormer \cite{zhou2022seedformer} & 0.557 & 0.592 & 0.598 & 0.579 & 0.585 & \textbf{0.626} & 0.520 & 0.605 & 0.583 \\
			\cline{1-10}
			ProxyFormer(Ours) & \textbf{0.555} & \textbf{0.590} & \textbf{0.597} & \textbf{0.571} & \textbf{0.562} & \textbf{0.626} & \textbf{0.518} & \textbf{0.507} & \textbf{0.577} \\
			\cline{1-10}
		\end{tabular}}
		\label{main_tab.2.}
	\end{table}

\noindent{\bfseries Qualitative comparison.} In Fig. \ref{main_Fig.9.}, we visualize the completion results of different methods on the PCN dataset. Compared with other methods, the results show that our method can perceive the position of the missing points while completing, and reduce the noisy points in the process of refinement. For example, as can be seen from the chair in the third row, except that the chair generated by PCN \cite{yuan2018pcn} has been deformed to a large extent, the other methods have successfully recovered the complete chair, but there are many noisy points around it. The chair completed by our method is more visually plausible. In addition, it can be evident from the chair leg and back that the chair completed by our method are more prominent in detail.

\subsection{Point Cloud Completion on ShapeNet-55/34}

    \begin{table*}[t]
        \centering
        \caption{Quantitative comparison on ShapeNet-55. For L2 CD $\times1000$ and DCD, lower is better. For F1-Score@1\%, higher is better.}
        \scalebox{0.8}{
        \begin{tabular}{c|ccccc|cccccc|ccc} 
        \hline
        Methods    & Table & Chair & Plane & Car  & Sofa & CD-S & CD-M & CD-H & DCD-S & DCD-M & DCD-H & CD-Avg & DCD-Avg & F1     \\ 
        \hline
        FoldingNet \cite{yang2018foldingnet} & 2.53  & 2.81  & 1.43  & 1.98 & 2.48 & 2.67 & 2.66 & 4.05 & - & - & - & 3.12 & - & 0.082  \\
        PCN \cite{yuan2018pcn}        & 2.13  & 2.29  & 1.02  & 1.85 & 2.06 & 1.94 & 1.96 & 4.08 & 0.570 & 0.609 & 0.676 & 2.66   & 0.618 & 0.133  \\
        TopNet \cite{tchapmi2019topnet}     & 2.21  & 2.53  & 1.14  & 2.18 & 2.36 & 2.26 & 2.16 & 4.30  & - & - & - & 2.91  & - & 0.126  \\
        PFNet \cite{huang2020pf}      & 3.95  & 4.24  & 1.81  & 2.53 & 3.34 & 3.83 & 3.87 & 7.97 & - & - & - & 5.22 & - & 0.339  \\
        GRNet \cite{xie2020grnet}      & 1.63  & 1.88  & 1.02  & 1.64 & 1.72 & 1.35 & 1.71 & 2.85 & 0.545 & 0.581 & 0.650 & 1.97 & 0.592 & 0.238  \\
        PoinTr \cite{yu2021pointr}     & 0.81  & 0.95  & 0.44  & 0.91 & 0.79 & 0.58 & 0.88 & 1.79 & 0.525 & 0.562 & 0.637 & 1.09  & 0.575 & 0.464  \\
        SeedFormer \cite{zhou2022seedformer} & 0.72  & \textbf{0.81}  & 0.40   & 0.89 & 0.71 & 0.50  & 0.77 & \textbf{1.49} & 0.513 & 0.549 & 0.612 & \textbf{0.92} & 0.558 & 0.472  \\ 
        \hline
        Ours       & \textbf{0.70}   & 0.83  & \textbf{0.34}  & \textbf{0.78} & \textbf{0.69} & \textbf{0.49} & \textbf{0.75} & 1.55 & \textbf{0.502} & \textbf{0.536} & \textbf{0.608} & 0.93 & \textbf{0.549} & \textbf{0.483}  \\
        \hline
        \end{tabular}}
        \label{main_tab.3.}
    \end{table*}

    \begin{table*}[h]
        \centering
        \caption{Quantitative comparison on ShapeNet-34. For L2 CD $\times1000$ and DCD, lower is better. For F1-Score@1\%, higher is better.}
        \scalebox{0.6}{
        \begin{tabular}{c|ccccccccc|ccccccccc} 
        \hline
        \multirow{2}{*}{Methods} & \multicolumn{9}{c|}{34~seen~categories} & \multicolumn{9}{c}{21~unseen~categories}  \\
                                 & CD-S & CD-M & CD-H & DCD-S & DCD-M & DCD-H & CD-Avg  & DCD-Avg & F1 & CD-S & CD-M & CD-H & DCD-S & DCD-M & DCD-H  & CD-Avg & DCD-Avg & F1         \\ 
        \hline
        FoldingNet  & 1.86 & 1.81 & 3.38  & - & - & - & 2.35 & - & 0.139 & 2.76 & 2.74 & 5.36 & - & - & - & 3.62 & - & 0.095      \\
        PCN & 1.87 & 1.81 & 2.97 & 0.571 & 0.617 & 0.683 & 2.22  & 0.624 & 0.150 & 3.17 & 3.08 & 5.29  & 0.601 & 0.638 & 0.692 & 3.85 & 0.644 & 0.101      \\
        TopNet  & 1.77 & 1.61 & 3.54 & - & - & - & 2.31 & - & 0.171     & 2.62 & 2.43 & 5.44 & - & - & - & 3.50  & - & 0.121      \\
        PFNet   & 3.16 & 3.19 & 7.71 & - & - & - & 4.68  & - & 0.347     & 5.29 & 5.87 & 13.33 & - & - & - & 8.16  & - & 0.322      \\
        GRNet  & 1.26 & 1.39 & 2.57 & 0.550 & 0.594 & 0.656 & 1.74  & 0.600 & 0.251     & 1.85 & 2.25 & 4.87 & 0.583 & 0.623 & 0.670 & 2.99 & 0.625 & 0.216      \\
        PoinTr  & 0.76 & 1.05 & 1.88 & 0.533 & 0.570 & 0.622 & 1.23  & 0.575 & 0.421     & 1.04 & 1.67 & 3.44 & 0.558 & 0.608 & 0.647 & 2.05 & 0.604 & 0.384      \\
        SeedFormer  & 0.48 & 0.70  & \textbf{1.30} & 0.513 & 0.561 & 0.608 & 0.83 & 0.561 & 0.452     & 0.61 & \textbf{1.07} & \textbf{2.35} & 0.541 & 0.587 & 0.629 & \textbf{1.34} & 0.586 & 0.402      \\ 
        \hline
        Ours  & \textbf{0.44} & \textbf{0.67} & 1.33 & \textbf{0.506} & \textbf{0.557} & \textbf{0.606} & \textbf{0.81} & \textbf{0.556} & \textbf{0.466}     & \textbf{0.60}  & 1.13 & 2.54 & \textbf{0.538} & \textbf{0.584} & \textbf{0.627} & 1.42 & \textbf{0.583} & \textbf{0.415}      \\
        \hline
        \end{tabular}}
        \label{main_tab.4.}
    \end{table*}

\noindent{\bfseries Dataset and evaluation metric.} We also evaluate our model on two other datasets, ShapeNet-55 and ShapeNet-34, proposed in PoinTr \cite{yu2021pointr}. In the two datasets, the input incomplete point cloud has 2048 points, and the complete point cloud contains 8192 points. Like  \cite{yu2021pointr}, we randomly select a viewpoint during training, and select a value from 2048 to 6144 to delete the corresponding points (25\% to 75\% of the complete point cloud), and then downsample the remaining points to 2048, as the input for model training. For the deleted part, we downsample it to 1536 points, as the true missing part. During testing, we choose 8 fixed viewpoints, and set the count of incomplete points to 2048, 4096 or 6144 (25\%, 50\% or 75\% of the complete point cloud), corresponding to three difficulty levels (simple, moderate and hard) during testing.

We use L2 CD, DCD and F-Score as evaluation metrics.

\noindent{\bfseries Quantitative comparison.} We list the quantitative performance of several methods on ShapeNet-55 and ShapeNet-34 in Tables \ref{main_tab.3.} and \ref{main_tab.4.}, respectively. We use CD-S, CD-M and CD-H to represent the CD results under the simple, moderate and hard settings. The same goes for DCD (the short line in the table indicates that the DCD value of this method is not competitive). It can be seen from Table \ref{main_tab.3.} that \emph{ProxyFormer} achieves the best performance on most listed categories and make lowest CD on simple and moderate settings. As for DCD, our method has the lowest  values on all the three settings, which proves that the objects completed by \emph{ProxyFormer} have the closest density distribution to GT. In terms of F1-Score, our method improved by 2.3\% compared with SeedFormer, reaching the highest value. Similarly, in Table \ref{main_tab.4.}, we can also see that the three indicators of CD, DCD and F1-Score of \emph{ProxyFormer} in the 34 visible categories have greatly exceeded PoinTr \cite{yu2021pointr}, and in all the three settings, we still get the lowest  DCD. Among 21 unseen categories, \emph{ProxyFormer} also make the lowest DCD and the highest F1-Score, which demonstrates the generalization performance of \emph{ProxyFormer}. (More results will be presented in the supplementary material)

% \noindent{\bfseries Qualitative comparison.} In Figure \ref{Fig.10.}, we show the visualization results of point cloud completion on ShapNet-55 by PoinTr \cite{yu2021pointr}, SeedFormer \cite{zhou2022seedformer}, and \emph{ProxyFormer}. We can intuitively see that \emph{ProxyFormer} can better complete the point cloud completion task. For example, for the table in the first row, although the resulting shape of PoinTr is complete, it fails to generate the details of the bottom of the table well, and SeedFormer introduces some noise points in the completion process, which affects the final result. But our method is able to take both shape and detail into account. Another example is the car in the fourth row. The point cloud contains not only a car, but also a person. There is a problem with the results of PoinTr and SeedFormer completion, that is, the distribution of point clouds is uneven, while \emph{ProxyFormer} can better perceive the distribution of missing points, thereby generating a more reasonable complete point cloud.

%     \begin{figure}[h]
% 		\centering
% 		%\setlength{\abovecaptionskip}{-0.03cm}
% 		\includegraphics[width=8cm]{pictures/ShapeNet55_Results.pdf}\\
% 		\caption{The visualization results of each method on ShapeNet-55, showing Table, Chair, Plane, Car and Sofa from top to bottom.}
% 		\label{Fig.10.}
% 	\end{figure}

\subsection{Point Cloud Completion on KITTI}
\noindent{\bfseries Dataset and evaluation metric.} To further evaluate our proposed model, we test it on the real-scanned dataset KITTI \cite{geiger2013vision}, which have no GT values as a reference, and some of the data are very sparse.

We use Fidelity Distance and Minimal Matching Distance (MMD) as evaluation metric.

    \begin{figure}[h]
		\centering
		\includegraphics[width=8.5cm]{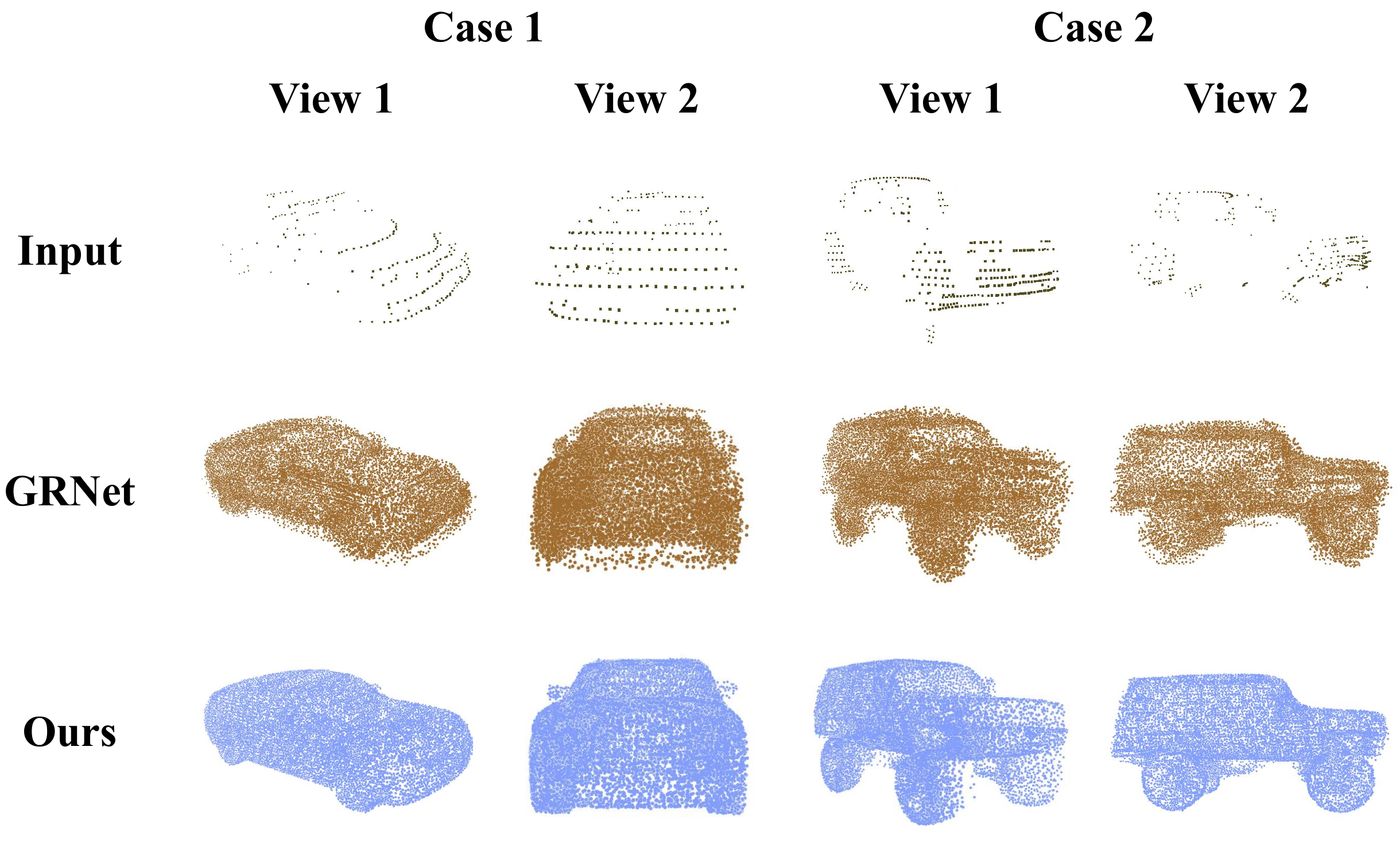}\\
		\caption{The visualization results on KITTI dataset. To better show the effect of completion, we provide two views for each car.}
		\label{main_Fig.11.}
	\end{figure}

\noindent{\bfseries Quantitative and Qualitative comparison.} Following GRNet \cite{xie2020grnet}, we fine-tune our pretrained model on ShapeNetCars (cars in the ShapeNet dataset) and evaluate it on the KITTI dataset, and the evaluation results are shown in Table \ref{main_tab.5.}. From this we can see that our method achieves the state-of-the-art on MMD (since both our method and PoinTr\cite{yu2021pointr} merge input into the final result, the Fidelity Distance is both $0$). As shown in Fig. \ref{Fig.11.}, our method performs well on such real scan data, and even if the input point cloud is very sparse, our method can restore its shape well, and by comparing with the results of GRNet, it can be seen that the point cloud generated by \emph{ProxyFormer} is softer, with less noisy points, and is more ornamental.

     \begin{table*}
        \centering
        \setlength{\abovecaptionskip}{-0.01cm}
        \caption{Quantitative comparison on KITTI dataset. For Fidelity Distance and Minimal Matching Distance (MMD), lower is better.}
        \scalebox{0.68}{
        \begin{tabular}{c|cccccccccc|c} 
        \hline
                 & AtlasNet \cite{groueix2018papier} & PCN \cite{yuan2018pcn}   & FoldingNet \cite{yang2018foldingnet} & TopNet \cite{tchapmi2019topnet} & MSN \cite{liu2020morphing}   & NSFA \cite{zhang2020detail} & CRN \cite{wang2020cascaded}   & GRNet \cite{xie2020grnet} & PoinTr \cite{yu2021pointr} & SeedFormer \cite{zhou2022seedformer} & Ours   \\ 
        \hline
        Fidelity & 1.759    & 2.235 & 7.467      & 5.354  & 0.434 & 1.281 & 1.023 & 0.816 & \textbf{0.000}  & 0.151      & \textbf{0.000}  \\
        MMD      & 2.108    & 1.366 & 0.537      & 0.636  & 2.259 & 0.891 & 0.872 & 0.568 & 0.526  & 0.516      & \textbf{0.508}  \\
        \hline
        \end{tabular}}
        \label{main_tab.5.}
    \end{table*}
    
\subsection{Ablation Studies}
In this subsection, we conduct ablation experiments for \emph{ProxyFormer} on the PCN dataset \cite{yuan2018pcn} to demonstrate the effectiveness of our proposed components.

\noindent{\bfseries Model Design Analysis.} The results of removing each component are listed in Table \ref{main_tab.6.}. The baseline model A only uses Point Transformer for feature extraction, and then send this feature into vanilla transformer encoder to get the feature for FoldingNet. We then add the position extractor to extract the position encoding for each point (Model B). It can be seen that the position extractor we designed reduces the CD of the baseline by $1.06$. After using missing part sensitive transformer for missing proxies prediction (Model C), we can observe that the CD drops significantly to $7.74$. When proxy alignment comes into play, the CD value drops a further $0.97$.

    \begin{table}[h]
        \centering
        \caption{Ablation study of each component. We add components including Position Extractor (PE), Missing Part Sensitive Transformer (Sensitive) and Proxy Alignment (PA) step by step.}
        \begin{tabular}{c|ccl|c} 
        \hline
        Model & PE & Sensitive & PA & CD     \\ 
        \hline
        A     &    &           &    & 11.08  \\
        B     & \checkmark  &           &    & 10.02   \\
        C     & \checkmark  & \checkmark  &    & 7.74   \\
        D     & \checkmark  & \checkmark  & \checkmark  & 6.77   \\
        \hline
        \end{tabular}
        \label{main_tab.6.}
    \end{table}

% \noindent{\bfseries Feature Extractor.} We respectively replace the point transformer with (1) a multilayer perceptron (MLP) composed of ordinary convolutional layers; (2) lightweight DGCNN; (3) traditional transformer using scalar attention. From Table \ref{tab.6.}, we can know that the point transformer can better capture the features of point clouds in the network structure proposed in this paper.

%     \begin{table}[h]
%         \centering
%         %\setlength{\abovecaptionskip}{-0.03cm}
%         \caption{Ablation study on Feature Extractor of FAPE Module.}
%         \scalebox{0.8}{\begin{tabular}{c|c} 
%         \hline
%         Method                      & CD-Avg  \\ 
%         \hline
%         w/~MLP                     & 8.08    \\
%         w/~DGCNN                   & 7.18    \\
%         w/~Traditional Transformer (scalar attention) & 7.53    \\
%         w/~Point Transformer (vector attention)      & \textbf{6.77}    \\
%         \hline
%         \end{tabular}}
%         \label{tab.6.}
%     \end{table}

After conducting ablation experiments on the proposed module, we further demonstrate the irreplaceability of position extractor through one more ablation experiments.

\noindent{\bfseries Position Extractor.} Our position extractor can synthesize the coordinates and feature information of the point cloud to more accurately represent the correlation and similarity between points. In this experiment, we compare our proposed position encoding method with two method: (1) directly use 3D coordinates as position encoding; (2) MLP-style position encoding method. which performs a simple upscaling operation on the 3D coordinates of the point cloud to form position encoding. The results in Table \ref{main_tab.7.} show that the direct use of 3D coordinates provides very limited position information and MLP cannot extract the positional of the point cloud well. Our proposed position encoding method can perceive the geometric structure of the point cloud well, and in this process, it is optimal to fuse the coordinates and feature information of 16 nearby points.

More ablation experiments and analysis will be given in the supplementary material.

    \begin{table}[h]
        \centering
        \setlength{\abovecaptionskip}{-0.00cm}
        \caption{Ablation study of Position Extractor of FAPE Module.}
        \scalebox{0.9}{\begin{tabular}{c|c|c} 
        \hline
        Methods    & Attempts & CD-Avg  \\ 
        \hline
        \multirow{2}{*}{w/o~Position~Extractor}     
        &  w/~3D coordinates    & 9.63             \\ 
        &  w/~MLP               & 7.83             \\ 
        \hline
        \multirow{3}{*}{w/~Position~Extractor} 
        & num~of~neighbor~=~8           & 6.86             \\
        & num~of~neighbor~=~16          & \textbf{6.77}    \\
        & num~of~neighbor~=~32          & 6.92             \\
        \hline
        \end{tabular}}
        \label{main_tab.7.}
    \end{table}

\subsection{Complexity Analysis}
\label{sec:complexity}

Our method achieves the best performance on many metrics on PCN dataset, ShapeNet-55, ShapeNet-34 and KITTI datasets. In Table \ref{main_tab.8.}, we list the number of parameters (Params), theoretical computation cost (FLOPs), the average chamfer distances (CD-Avg) and the average density-aware chamfer distances (DCD-Avg) of our method and other six methods. It can be seen that our method can obtain the lowest DCD-Avg while having the smallest FLOPs, and it is the second best only a litter inferior to SeedFormer \cite{zhou2022seedformer} in terms of CD. Since the transformer decoder part was no longer needed in \emph{ProxyFormer}, the number of parameters is also greatly reduced compared to PoinTr \cite{yu2021pointr}, which also shows that our method can better balance computational cost and performance.
\vspace{-0.2cm}
    \begin{table}[h]
        \centering
        \caption{Complexity analysis. We show the the number of parameter (Params) and FLOPs of our method and six existing methods. We also provide the distance metrics  CD-Avg and  DCD-Avg on PCN dataset.}
        \scalebox{0.85}{\begin{tabular}{c|cc|cc} 
        \hline
        Methods      & Params    & FLOPs   & CD-Avg    & DCD-Avg     \\ 
        \hline
        FoldingNet \cite{yang2018foldingnet}  & \textbf{2.41M} & 27.65G         & 14.31    & 0.688     \\
        PCN \cite{yuan2018pcn}         & 6.84M          & 14.69G         & 9.64     & 0.651      \\
        GRNet \cite{xie2020grnet}       & 76.71M         & 25.88G         & 8.83     & 0.622      \\
        PoinTr \cite{yu2021pointr}      & 30.9M          & 10.41G         & 8.38     & 0.611      \\
        SnowflakeNet \cite{xiang2021snowflakenet} & 19.32M         & 10.32G         & 7.21     & 0.585    \\
        SeedFormer \cite{zhou2022seedformer}  & 3.20M          & 29.61G         & \textbf{6.74} & 0.583 \\ 
        \hline
        Ours         & 12.16M         & \textbf{9.88G} & 6.77   & \textbf{0.577}        \\
        \hline
        \end{tabular}}
        \label{main_tab.8.}
    \end{table}
\vspace{-0.5cm}
\section{Conclusion}
In this paper, we propose a new point cloud completion framework named \emph{ProxyFormer}, which designs a missing part sensitive transformer to generate missing proxies. We extract feature and position for the missing points and form point proxies. We regularize the distribution of predicted point proxies through proxy alignment, so as to better complete the input partial point clouds. Experiments also show that our method achieves state-of-the-art performance on multiple metrics on several challenging benchmark datasets and has the fastest inference speed.

\section*{Acknowledgement}
This work was supported in part by the National Key Research and Development Program of China under Grant 2021ZD0113203 and the Natural Science Foundation of China under Grant 62272227.

%%%%%%%%% REFERENCES
{\small
\bibliographystyle{ieee_fullname}
\bibliography{egbib}
}

% Due to the limited resolution of 3D scanning equipment such as Lidar, the mutual occlusion between targets, and the transparency of the target surface material, the collected 3D point cloud data is often incomplete. 

\begin{appendices}

\section{Overview}
In this supplementary material, we provide additional information to complement the manuscript. First, we provide the details of missing part extractor in Sec. \ref{sec:missing}. Second, we present additional implementation details and experimental settings of \emph{ProxyFormer} (Sec. \ref{sec:implementation}). At last, we do more ablation studies and visualize more qualitative results of our method on the PCN dataset and ShapeNet-55/34. We also present detailed quantitative results on ShapeNet-55 and ShapeNet-34. (Sec. \ref{sec:experimental}).

\section{Missing Part Extractor}
\label{sec:missing}

\begin{figure*}
		\centering
		\includegraphics[width=16cm]{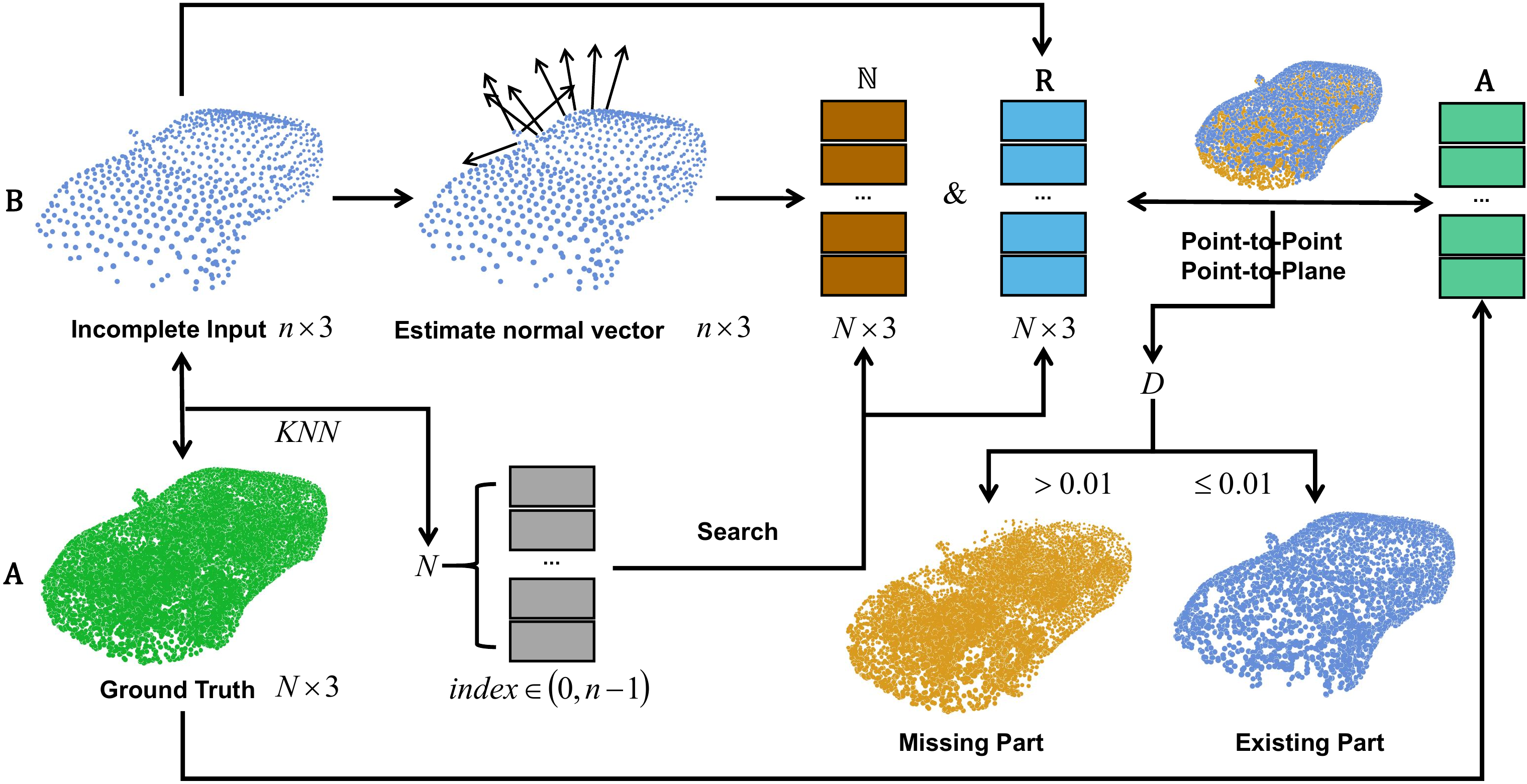}\\
		\caption{The pipeline of Missing Part Extractor. $N$ is the count of points in ground truth and $n$ is the count of points in incomplete input. Firstly, normal vectors are estimated for the incomplete input $\mathbf{B}$. Secondly, the KNN algorithm is used to obtain the index with the shortest distance in $\mathbf{B}$ corresponding to each point in the ground truth $\mathbf{A}$. A set of points $\mathbf{R}$ and a set of normal vectors $\mathbb{N}$ are obtained according to the index. Finally, we use $\mathbf{R}$, $\mathbb{N}$ and $\mathbf{A}$ to calculate the point-to-point distance and the point-to-plane distance. The weighted sum $D$ of these two distances is used for comparison with a pre-defined threshold $0.01$. If the distance is less than or equal to the threshold, the point belongs to the existing part, otherwise it belongs to the missing part.}
		\label{Fig.1.}
	\end{figure*}

The given complete point cloud and incomplete point cloud of some point cloud completion common datasets such as PCN are not in the same scale of coordinate systems. So it is not possible to simply use the set difference operation to obtain the missing part. So we design a missing part extractor based on point-to-point and point-to-plane distances \cite{tian2017geometric}. Fig. \ref{Fig.1.} is a detailed description of the missing part extractor. 

Let $\textbf{A}$ and $\textbf{B}$ denote the Ground Truth (GT) and the incomplete point cloud, respectively. Firstly, we use the normal vector estimation method \cite{tombari2010unique} to obtain the normal vector of each point in $\textbf{B}$. For each point $\bm{p}_i$, we assume the covariance matrix $\mathcal{C}$ as follows:

	\begin{small}
	\begin{equation}
	    \begin{aligned}
    	   &\mathcal{C}=\frac{1}{k} \sum_{i=1}^k \cdot\left(\bm{p}_i-\overline{\bm{p}}\right) \cdot\left(\bm{p}_i-\overline{\bm{p}}\right)^T, \\
    	   &\mathcal{C} \cdot \overrightarrow{\mathbf{v}_j}=\lambda_j \cdot \overrightarrow{\mathbf{v}_j}, j \in\{0,1,2\} ~ .
	   \end{aligned}
	   \label{eq.1.}
    \end{equation}
    \end{small}
	
Here $k$ refers to the $k$ points closest to $\bm{p}_i$. $\overline{\bm{p}}$ is the centroid of the nearest neighbor. $\lambda_j$ is the $j^{th}$ eigenvalue and $\overrightarrow{\mathbf{v}_j}$ is the $j^{th}$ eigenvector. Then, the eigenvector corresponding to the largest eigenvalue is selected and normalized as the normal vector of the point.

    \begin{figure}[h]
		\centering
		\includegraphics[width=8cm]{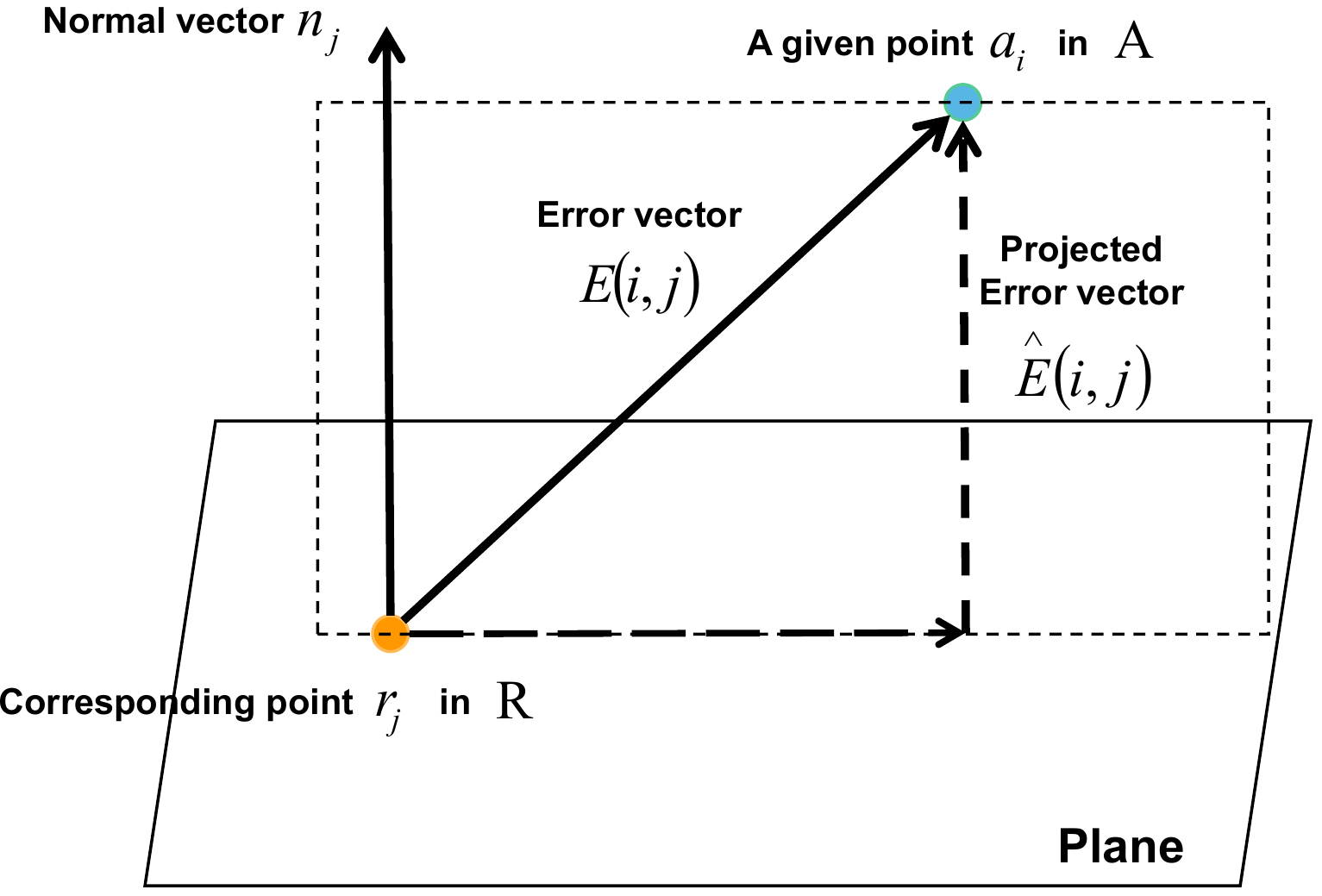}\\
		\caption{Point-to-point and point-to-plane distance. We use vector calculations to get the two distances. $a_i$ is a point in point cloud $\mathbf{A}$ and $r_j$ is the corresponding point in $\mathbf{R}$. $n_j$ is the normal vector of point $r_j$. The error vector $E(i, j)$ is computed by connecting point $r_j$ to point $a_i$. The projected error vector $\hat{E}\left(i, j\right)$ is the new error vector after $E(i,j)$ is projected along the normal vector $n_j$.}
		\label{Fig.2.}
	\end{figure}

After the calculation of the normal vector, the KNN algorithm is used to obtain the index of the corresponding adjacent points between the GT and the incomplete point cloud, which can be expressed as: $idx=KNN\left(\rm{\mathbf{A}},\rm{\mathbf{B}},\rm{K}=1\right)$. According to this index, we go to $\textbf{B}$ to find the points $\bm{b}_{i}$ ($i \in idx$) and normal vectors $\bm{n}_{i}$ ($i \in idx$) of the adjacent points. Here we denote the point cloud composed of $b_i$ as $\mathbf{R}$ (it contains $N$ points and the value of each point comes from the incomplete input, but they are one-to-one correspondence with points in GT), and denote the set of normal vectors $n_i$ corresponding to each point in the point cloud $\mathbf{R}$ as $\mathbb{N}$.

% $\mathbb{R}$ and $\mathbb{N}$ represent the set of $\bm{p}_n$ and the set of $\bm{n}_n$, respectively. 

Then as shown in Fig. \ref{Fig.2.}, we calculate the point-to-point distance and the point-to-plane distance as follows.

(1) For a point $a_i$ in point cloud $\mathbf{A}$, \emph{i.e.}, the blue point in the figure, a corresponding point $r_j$ in the point cloud $\mathbf{R}$ can be found, \emph{i.e.}, the orange point in the figure. Vice versa.

(2) Similarly, we can get the corresponding normal vector $n_j$ from $\mathbb{N}$.

(3) We connect point $r_j$ to point $a_i$ to calculate an error vector whose length is the point-to-point distance, \emph{i.e.},

    \begin{equation}
        \begin{gathered}
            D_{a_i,r_j}^{c2c} = || E\left(i,j\right) ||_2~.
        \end{gathered}
        \label{eq.2.}
    \end{equation}

(4) We project the error vector $E\left(i, j\right)$ along the direction of normal vector $n_j$ to get another error vector $\hat{E}\left(i, j\right)$ whose length is the point-to-plane distance, \emph{i.e.},

    \begin{equation}
        \begin{gathered}
            D_{a_i,r_j}^{c2p} = || \hat{E}(i,j) ||_2 = \frac{E(i,j) \cdot n_j}{||n_j||_2} = || E\left(i,j\right) \cdot n_j ||_2~.
        \end{gathered}
        \label{eq.3.}
    \end{equation}

% use the formula $D_{\rm{\mathbf{A}},\rm{\mathbf{B}}}^{c2c}=\sqrt{\left(\rm{\mathbf{A}}-\mathbb{R}\right)^2}$ and $D_{\rm{\mathbf{A}},\rm{\mathbf{B}}}^{c2p}=\sqrt{\left(\left(\rm{\mathbf{A}}-\mathbb{R}\right)\cdot\mathbb{N}\right)^2}$ to calculate point-to-point and point-to-plane distances. 

(5) We use the weighted sum of the two distances as a comparison condition,

    \begin{equation}
        \begin{gathered}
            D=\alpha D_{a_i,r_j}^{c2c}+\beta D_{a_i,r_j}^{c2p} ~,
        \end{gathered}
        \label{eq.4.}
    \end{equation}
where $\alpha = 0.2$ and $\beta = 0.8$. 

We perform the above operations on each point in the GT $\mathbf{A}$. If the calculated distance of the point is less than or equal to the preset threshold of $0.01$, the point is stored in the set of existing parts, otherwise, it is stored in the set of missing parts. This process can be formulated as,

    \begin{equation}
        \begin{gathered}
            a_i \in \left\{
            \begin{array}{l}
            Existing ~ part ~~~~ if ~~~~ D <= 0.01\\Missing ~ part  ~~~~~~~~~~~~~~~else \\
            \end{array} \right. ~, \forall a_i \in \textbf{A}~ .
        \end{gathered}
        \label{eq.5.}
    \end{equation}

\begin{figure}[h]
		\centering
		\includegraphics[width=8.5cm]{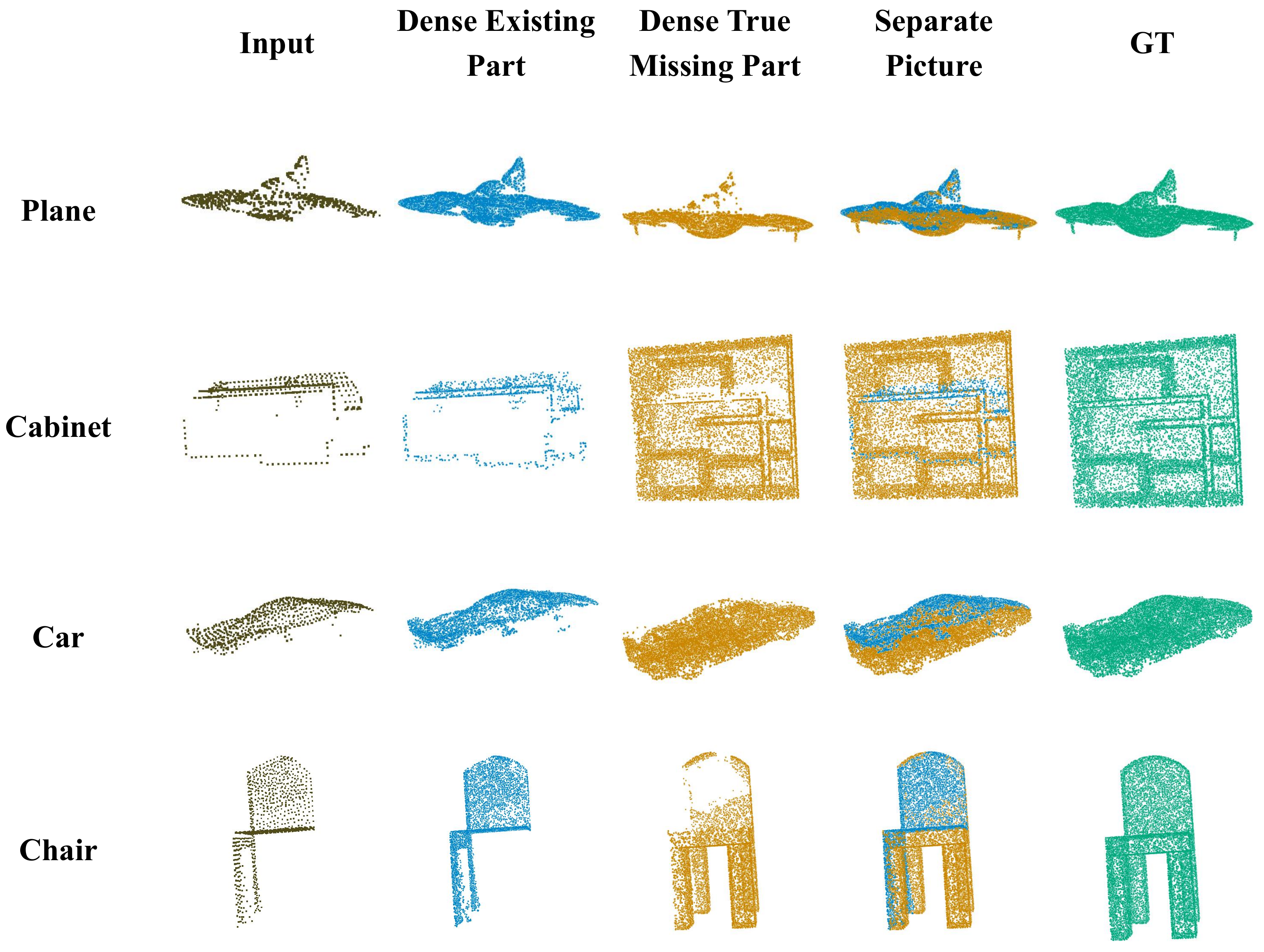}\\
		\caption{Some results of missing part extractor. From left to right are input, dense existing part, dense missing part, the splicing of the two and GT. From top to bottom are cases in four categories plane, cabinet, car and chair.}
		\label{Fig.3.}
	\end{figure}

The missing part is separated from the GT by the above method, and then downsampled to the same number as the incomplete input point for subsequent use. In Fig. \ref{Fig.3.}, we show some results of missing part extractor. 
% Only the input, the dense true missing part and the GT are used during training. What's more, the dense true missing part is only used to be downsampled to the missing part point cloud used for training (in the PCN dataset, it is downsampled to 3584 points), and to calculate the chamfer distance with coarse predicted missing part to provide the loss $d_{CD}\left(C_{pm}, C_{rm}\right)$ mentioned in Sec. 3.5 of the main text.

%%%%%%%%% BODY TEXT
\section{Implementation Details}
\label{sec:implementation}

In this section, we provide additional implementation details of the proposed \emph{ProxyFormer}. 

\subsection{Basic experimental setup}

We use Pytorch \cite{paszke2019pytorch} for our implementation. All network are trained on a single NVIDIA GeForce RTX 3090 graphics card. AdamW optimizer \cite{loshchilov2018fixing} is uesd to train the network with initial learning rate as $5e-4$ and weight decay as $5e-4$. When training on the PCN dataset, we set the batch size to $64$ and train the model for $400$ epochs with the continuous learning rate decay of $0.95$ for every $20$ epochs. When training on ShapeNet-55/34, we set the batch size to $128$ and train the model for $300$ epochs with the continuous learning rate decay of $0.83$ for every $20$ epochs.

\subsection{Four types of proxies} 

There are four types of proxies during training, existing proxies (\emph{EP}), missing proxies (\emph{MP}), true missing proxies (\emph{true-MP}), predicted missing proxies (\emph{pre-MP}). \emph{true-MP} is not available during testing. When training on the PCN dataset \cite{yuan2018pcn}, \emph{EP} is a $128 \times 384$ matrix and \emph{MP}, \emph{true-MP} and \emph{pre-MP} are both $224 \times 384$ matrices. When training on the ShapeNet-55/34 \cite{yu2021pointr}, \emph{EP} is a $128 \times 384$ matrix and \emph{MP}, \emph{true-MP} and \emph{pre-MP} are both $96 \times 384$ matrices.

\subsection{Feature and Position Extractor}

\noindent{\bfseries Feature Extraction.} We use farthest point sampling \cite{qi2017pointnet++} (FPS) to downsample the point cloud and use point transformer \cite{zhao2021point} (PT) to help exchange information between localized feature vectors. First we use an shared MLP to upscale the 3D coordinates of the point cloud to 32-dim as features. Suppose the set of input point cloud is $P_i$, at each downsampling stage: (1) perform farthest point sampling in $P_i$ and form a new set $P_o$ ($P_o \subset P_i$); (2) a $k$NN graph is performed on the $P_i$ (we use $k=16$ in our experiments); (3) local max pooling is used to aggregate the features of nearby $k$ points to the current center point; (4) enhance features with point transformer block. 

On both PCN and ShapeNet-55/34 datasets, the number of incomplete point cloud points input is $P = 2048$. Detailed network architecture is as follows: 

incomplete input ($2048 \times 3$) $\rightarrow$ shared-MLP ($2048 \times 3$ to $2048 \times 32$) $\rightarrow$ FPS ($2048 \times 32$ to $512 \times 128$) $\rightarrow$ PT ($512 \times 128$ to $512 \times 128$) $\rightarrow$ FPS ($512 \times 128$ to $128 \times 384$) $\rightarrow$ PT ($128 \times 384$ to $128 \times 384$).

On PCN \textbf{(ShapeNet-55/34)} dataset, the number of missing part points input is $P = 3584 \textbf{(1536)}$. Bold numbers correspond to the settings on ShapNet-55/34. The process of FAPE is: 

missing part ($3584 \textbf{(1536)} \times 3$) $\rightarrow$ shared-MLP ($3584 \textbf{(1536)} \times 3$ to $3584 \textbf{(1536)} \times 32$) $\rightarrow$ FPS ($3584 \textbf{(1536)} \times 32$ to $896 \textbf{(384)} \times 128$) $\rightarrow$ PT ($896 \textbf{(384)} \times 128$ to $896 \textbf{(384)} \times 128$) $\rightarrow$ FPS ($896 \textbf{(384)} \times 128$ to $224 \textbf{(96)} \times 384$) $\rightarrow$ PT ($224 \textbf{(96)} \times 384$ to $224 \textbf{(96)} \times 384$).

\noindent{\bfseries Position Extraction.} In order to keep the final position dimension consistent with the feature dimension extracted in the previous step, we set both $C_2$ and $C_{out}$ in the main text to $384$. Suppose the set of center points is $P_c$ and the final feature of feature extraction is $F_c$. Perform $k$-neighbor (we use $k=16$ in our experiments) subtraction ($k$NN-sub) and aggregation ($k$NN-agg) operations on $P_c$ and $F_c$, respectively. 

On both PCN and ShapeNet-55/34 datasets, the number of incomplete center points input is $n = 128$. Detailed network architecture is as follows:

$P_c$ ($128 \times 3$) $\rightarrow$ $k$NN-sub ($128 \times 3$ to $128 \times 16 \times 3$) $\rightarrow$ $k$NN-agg ($128 \times 16 \times 3$ to $128 \times 3$)~,

$F_c$ ($128 \times 384$) $\rightarrow$ $k$NN-sub ($128 \times 384$ to $128 \times 16 \times 384$) $\rightarrow$ $k$NN-agg ($128 \times 16 \times 384$ to $128 \times 384$)~,

[$P_c$, $F_c$] ($128 \times 387$) $\rightarrow$ shared-MLP ($128 \times 387$ to $128 \times 384$)~. 

On PCN \textbf{(ShapeNet-55/34)} dataset, the number of missing part center points input is $n = 224 \textbf{(96)}$. Bold numbers correspond to the settings on ShapNet-55/34. Detailed network architecture is as follows: 

$P_c$ ($224 \textbf{(96)} \times 3$) $\rightarrow$ $k$NN-sub ($224 \textbf{(96)} \times 3$ to $224 \textbf{(96)} \times 16 \times 3$) $\rightarrow$ $k$NN-agg ($224 \textbf{(96)} \times 16 \times 3$ to $224 \textbf{(96)} \times 3$)~,

$F_c$ ($224 \textbf{(96)} \times 384$) $\rightarrow$ $k$NN-sub ($224 \textbf{(96)} \times 384$ to $224 \textbf{(96)} \times 16 \times 384$) $\rightarrow$ $k$NN-agg ($224 \textbf{(96)} \times 16 \times 384$ to $224 \textbf{(96)} \times 384$)~,

[$P_c$, $F_c$] ($224 \textbf{(96)} \times 387$) $\rightarrow$ shared-MLP ($224 \textbf{(96)} \times 387$ to $128 \times 384$)~. 

Subsequent attention score calculations do not affect the dimension of this feature.

\subsection{Missing Part Prediction}

In the main text we have mentioned using the incomplete seed feature to generate coarse missing part. When training on PCN dataset, the predicted coarse missing part contains 224 points and the predicted dense missing part contains 14336 points. When training on ShapeNet-55/34, the predicted coarse missing part contains 96 points and the predicted dense missing part contains 6144 points.

\subsection{Missing Feature Generator} 

In Mssing Feature Generator, we set $N$ to $128$ and $M$ to $224$ (on PCN dataset) or $96$ (on ShapeNet-55/34). $C=384$ and we divide the feature dimension equally into $U=16$ groups.

\subsection{Missing Part Sensitive Transformer}

Like other tansformer-based methods, we stack the missing part sensitive transformer and set its depth to 8.

\noindent{\bfseries Multi-Head Self-Attention.} This structural design allows each attention mechanism to map to different spaces through Query, Key and Value to learn features. In this way, the different feature parts of each proxy are optimized, so as to  make the proxies contain more diverse representations. In all our experiments, we set the number of multi-head attention heads to $8$.

\noindent{\bfseries Feed-Forward Network (FFN).} Referring to \cite{vaswani2017attention} and \cite{yu2021pointr}, we set up the feed-forward network as two linear layers with ReLU activation function and dropout.

\subsection{The calculation of DCD}

Wu \emph{et al.} \cite{wu2021density} study the limitations of CD, believe that CD is not the optimal indicator for evaluating the visual quality of point cloud completion tasks, and proposed the density-aware chamfering distance (DCD), which can retain the measurement ability similar to CD and can also better judge the visual effect of the point cloud completion result. The formula for DCD is as follows:

    \begin{equation}
        \begin{aligned}
        	d_{D C D}\left(S_1, S_2\right)=\frac{1}{2}\left(\frac{1}{\left|S_1\right|} \sum_{x \in S_1}\left(1-\frac{1}{n_{\hat{y}}} e^{-\alpha\|x-\hat{y}\|_2}\right)\right)\\
        	+\frac{1}{2}\left(\frac{1}{\left|S_2\right|} \sum_{y \in S_2}\left(1-\frac{1}{n_{\hat{x}}} e^{-\alpha\|y-\hat{x}\|_2}\right)\right) ~ ,
        \end{aligned}
    	\label{eq.6.}
    \end{equation}
where $\hat{y}=min_{y\in S_2}\|x-y\|_2, \hat{x}=min_{y\in S_1}\|y-x\|_2$, and $\alpha$ denotes a temperature scalar, which we set to 1000, just as the original text describes.

\section{Addtional Ablation Studies and Experimental Results}
\label{sec:experimental}

In this section, we first conduct more ablation experiments on some of the components used in \emph{ProxyFormer}. Then, we present more qualitative and quantitative experimental results to further demonstrate the effectiveness of our method.

\subsection{More ablation sutdies and analysis}

\noindent{\bfseries Feature Extractor.} We respectively replace the point transformer with (1) a multilayer perceptron (MLP) composed of ordinary convolutional layers; (2) lightweight DGCNN; (3) traditional transformer using scalar attention. From Table \ref{tab.1.}, we can know that the point transformer can better capture the features of point clouds in the network structure proposed in this paper.

    \begin{table}[h]
        \centering
        \caption{Ablation study on Feature Extractor of FAPE Module.}
        \scalebox{1}{\begin{tabular}{c|c} 
        \hline
        Method                      & CD-Avg  \\ 
        \hline
        w/~MLP                     & 8.08    \\
        w/~DGCNN                   & 7.18    \\
        w/~Traditional Transformer (scalar attention) & 7.53    \\
        w/~Point Transformer (vector attention)      & \textbf{6.77}    \\
        \hline
        \end{tabular}}
        \label{tab.1.}
    \end{table}

\noindent{\bfseries Missing Feature Generator.} Many methods use the tiled copy of global feature as the feature of each point to generate a complete point cloud. In this experiment, we make two attempts to the network after removing the missing feature generator: (1) use random feature as the feature of \emph{MP}; (2) use the global feature (incomplete seed feature) as the feature of \emph{MP}. From Table \ref{tab.2.}, we can see that the effect of using random feature and global feature is not ideal. Using Missing Feature Generator, a more reasonable feature corresponding to the missing point can be generated from the feature of the existing points. Furthermore, experiments show that in the process of feature generation, evenly dividing the features into 16 groups can reduce the training time while reducing the CD of the final result to the lowest.

    \begin{table}[h]
        \centering
        \caption{Ablation study of Missing Feature Generator}
        \scalebox{0.8}{\begin{tabular}{c|c|c} 
        \hline
        Methods &  Attempts  & CD-Avg \\ 
        \hline
        \multirow{2}{*}{w/o~Missing Feature Generator}   & random~feature    & 7.90  \\
        & copy~global~feature     & 8.49             \\ 
        \hline
        \multirow{4}{*}{\begin{tabular}[c]{@{}c@{}}w/~Missing Feature Generator\\\end{tabular}} & \begin{tabular}[c]{@{}c@{}}num~of~feature~patch~=~1\\\end{tabular}  & 6.83             \\
        & \begin{tabular}[c]{@{}c@{}}num~of~feature~patch~=~8\\\end{tabular} & 6.80             \\
        & \begin{tabular}[c]{@{}c@{}}num~of~feature~patch~=16\\\end{tabular}  & \textbf{6.77}    \\
        & \begin{tabular}[c]{@{}c@{}}num~of~feature~patch~=~32\\\end{tabular} & 6.85             \\
        \hline
        \end{tabular}}
        \label{tab.2.}
    \end{table}

    \begin{table*}[t]
		\centering
		\caption{Ablation study on feature used for generating predicted coarse missing part. For CD, lower is better.}
		\begin{tabular}{c|c|c|c|c|c|c|c|c|c}
			\cline{1-10}
			\multirow{2}{*}{Methods} & \multicolumn{9}{c}{Chamfer Distance($10^{-3}$)}
			\\
			\cline{2-10}
			& Air & Cab & Car & Cha & Lam  & Sof & Tab & Ves & Ave \\
			\cline{1-10}
			ProxyFormer(using predicted missing seed feature) & 4.13 & 9.22 & 8.01 & 7.57 & 5.60 & 9.11 & 6.43 & 6.28 & 7.04  \\
			\cline{1-10}
			ProxyFormer(using incomplete seed feature) & \textbf{4.01} & \textbf{9.01} & \textbf{7.88} & \textbf{7.11} & \textbf{5.35} & \textbf{8.77} & \textbf{6.03} & \textbf{5.98} & \textbf{6.77} \\
			\cline{1-10}
		\end{tabular}
		\label{tab.3.}
	\end{table*}

    \begin{table*}[t]
		\centering
		\caption{Ablation study on feature used for generating predicted coarse missing part. For DCD, lower is better.}
		\scalebox{0.9}{
		\begin{tabular}{c|c|c|c|c|c|c|c|c|c}
			\cline{1-10}
			\multirow{2}{*}{Methods} & \multicolumn{9}{c}{Density-aware Chamfer Distance}
			\\
			\cline{2-10}
			& Air & Cab & Car & Cha & Lam  & Sof & Tab & Ves & Ave \\
			\cline{1-10}
			ProxyFormer(using predicted missing seed feature) & 0.572 & 0.604 & 0.611 & 0.597 & 0.609 & 0.641 & 0.530 & 0.579 & 0.593 \\
			\cline{1-10}
			ProxyFormer(using incomplete seed feature) & \textbf{0.555} & \textbf{0.590} & \textbf{0.597} & \textbf{0.571} & \textbf{0.562} & \textbf{0.626} & \textbf{0.518} & \textbf{0.507} & \textbf{0.577} \\
			\cline{1-10}
		\end{tabular}}
		\label{tab.4.}
	\end{table*}

\noindent{\bfseries Coarse missing part prediction.} We try to use the predicted missing seed feature to generate coarse missing part and did ablation experiments with this. The quantitative results on PCN dataset \cite{yuan2018pcn} are listed in Tables \ref{tab.3.} and \ref{tab.4.}. Through this experiment, it can be proved that the features extracted from the partial input can better predict the approximate position of the missing part (that is, the position of the center point of the coarse missing part). However, as analyzed in the main text, it is not enough to use only the features extracted from the partial input for the prediction of the missing details. Using Missing Feature Generator can better generates features that incorporate the missing details.

\begin{figure*}
      \centering
      \begin{subfigure}{0.33\linewidth}
        \includegraphics[width=6.5cm]{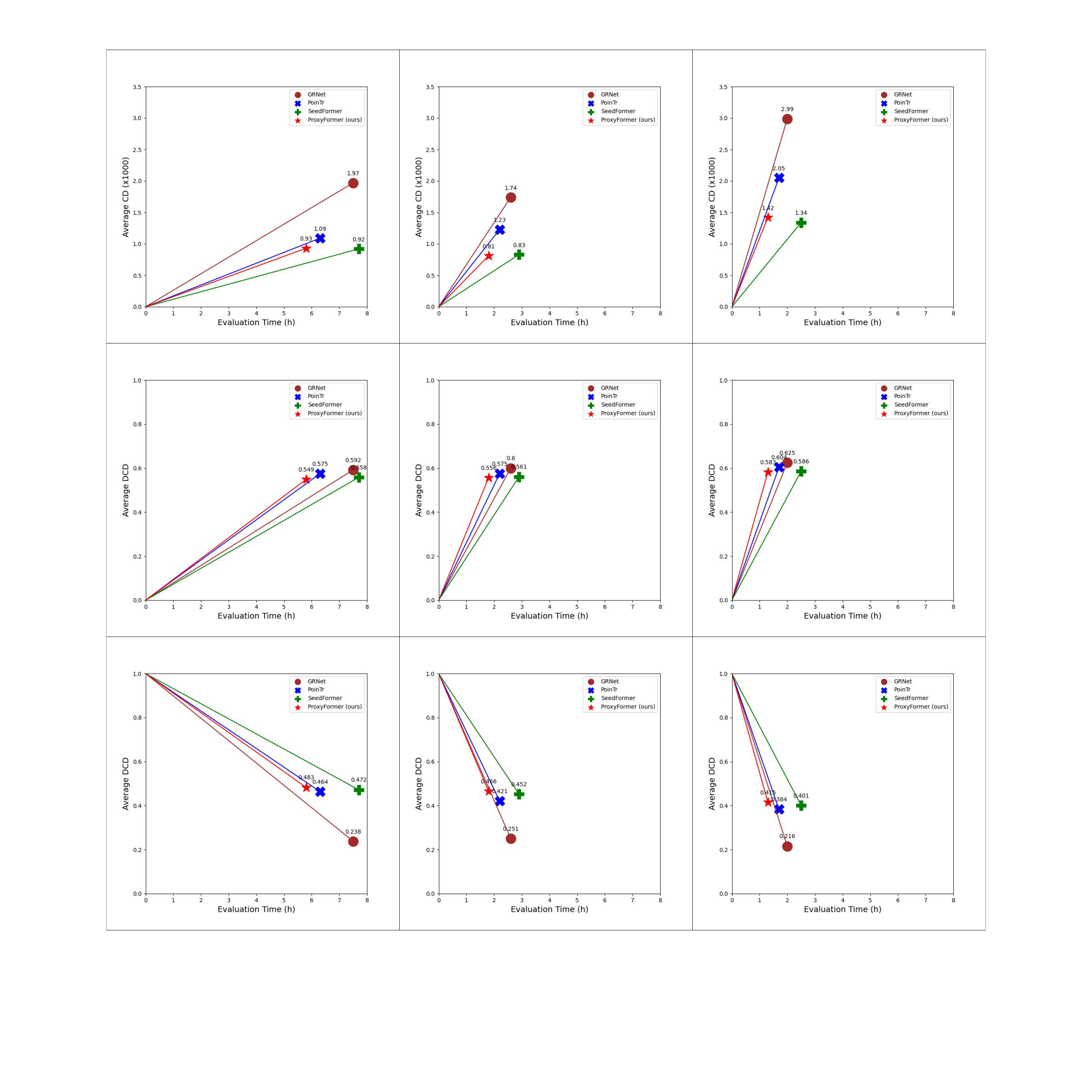}
        \caption{CD vs. Eval. time on ShapeNet-55.}
        \label{Fig.4-1.}
      \end{subfigure}
      \begin{subfigure}{0.33\linewidth}
        \includegraphics[width=6.5cm]{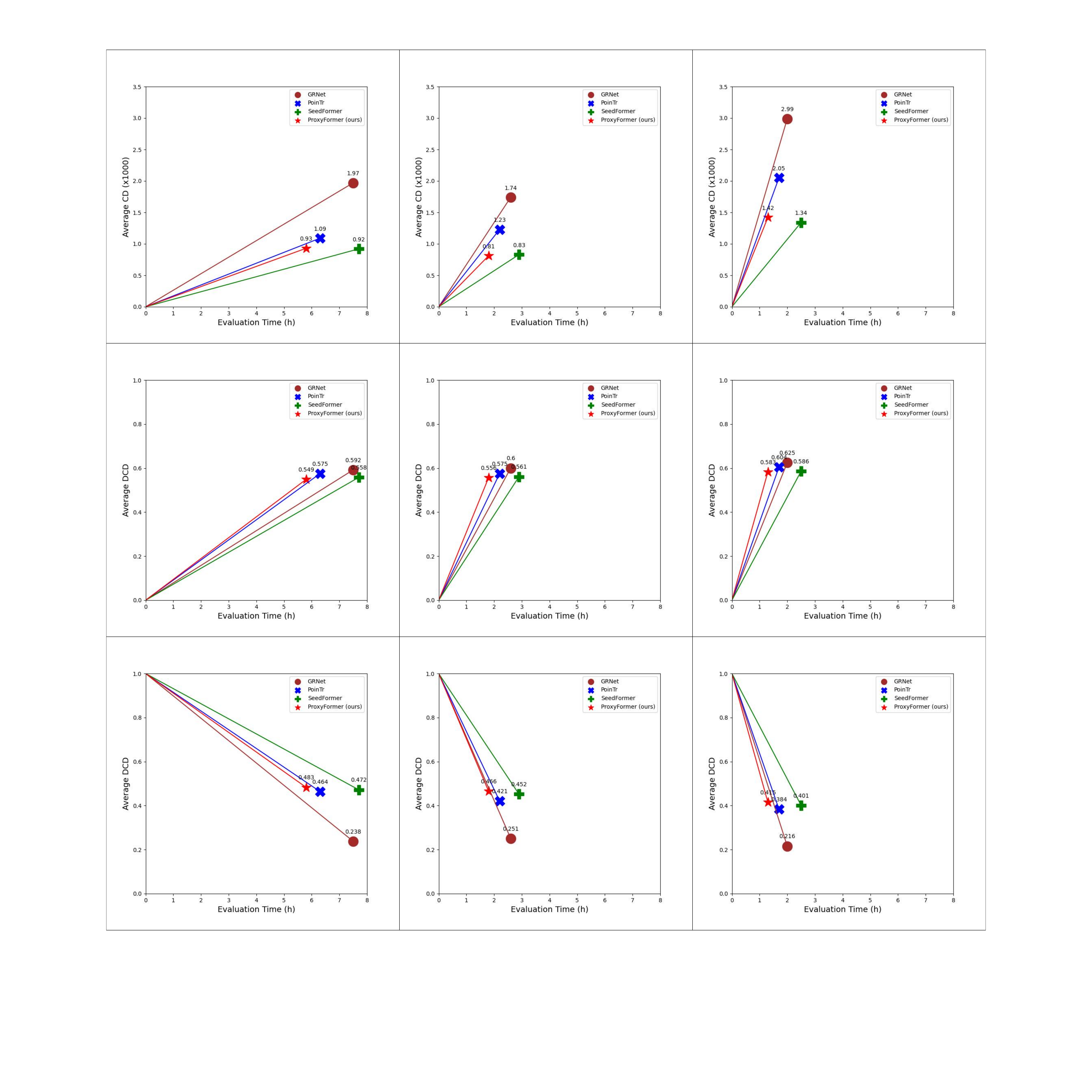}
        \caption{CD vs. Eval. time on 34 seen categories.}
        \label{Fig.4-2.}
      \end{subfigure}
      \begin{subfigure}{0.33\linewidth}
        \includegraphics[width=6.5cm]{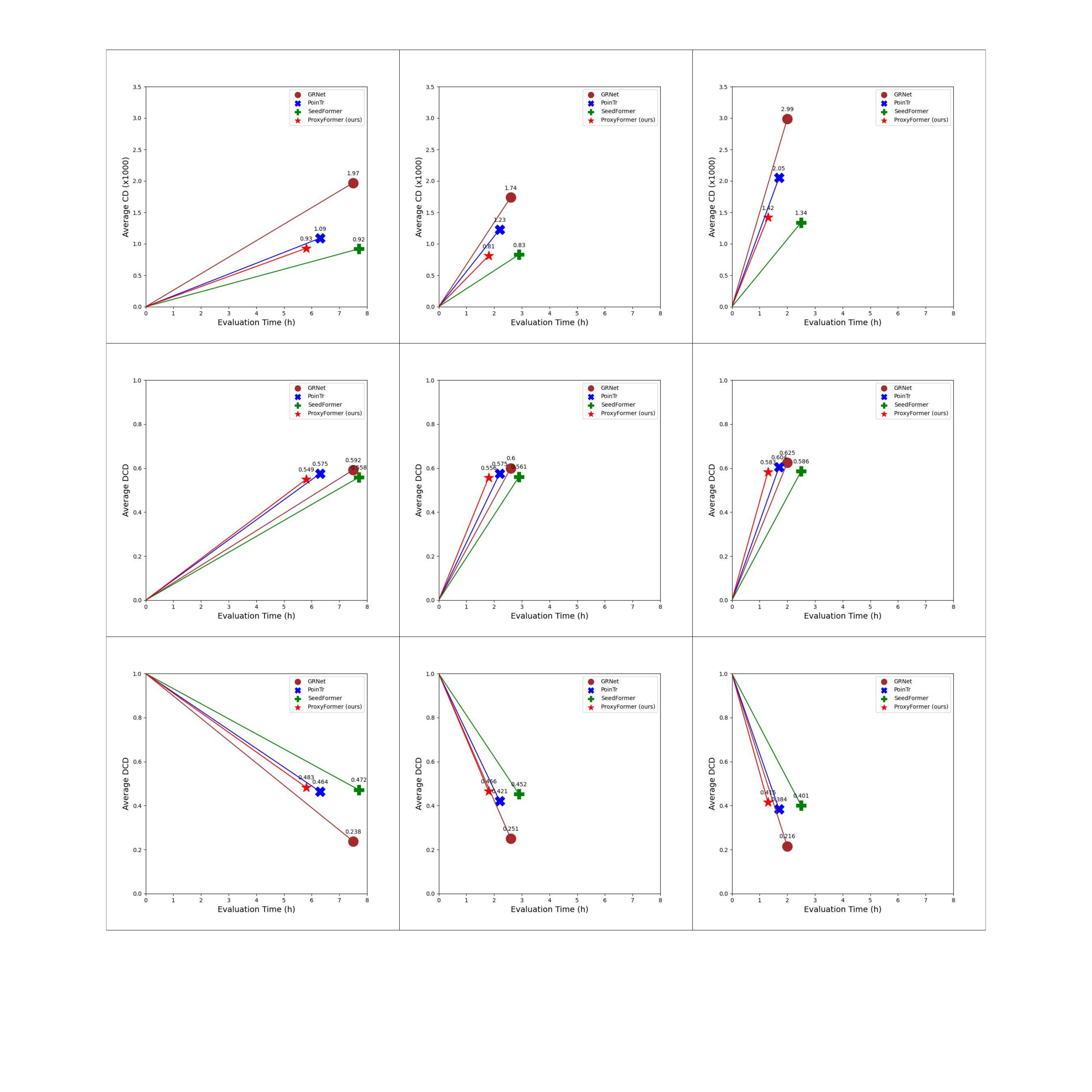}
        \caption{CD vs. Eval. time on 21 novel categories.}
        \label{Fig.4-3.}
      \end{subfigure}
      \\
      \begin{subfigure}{0.33\linewidth}
        \includegraphics[width=6.5cm]{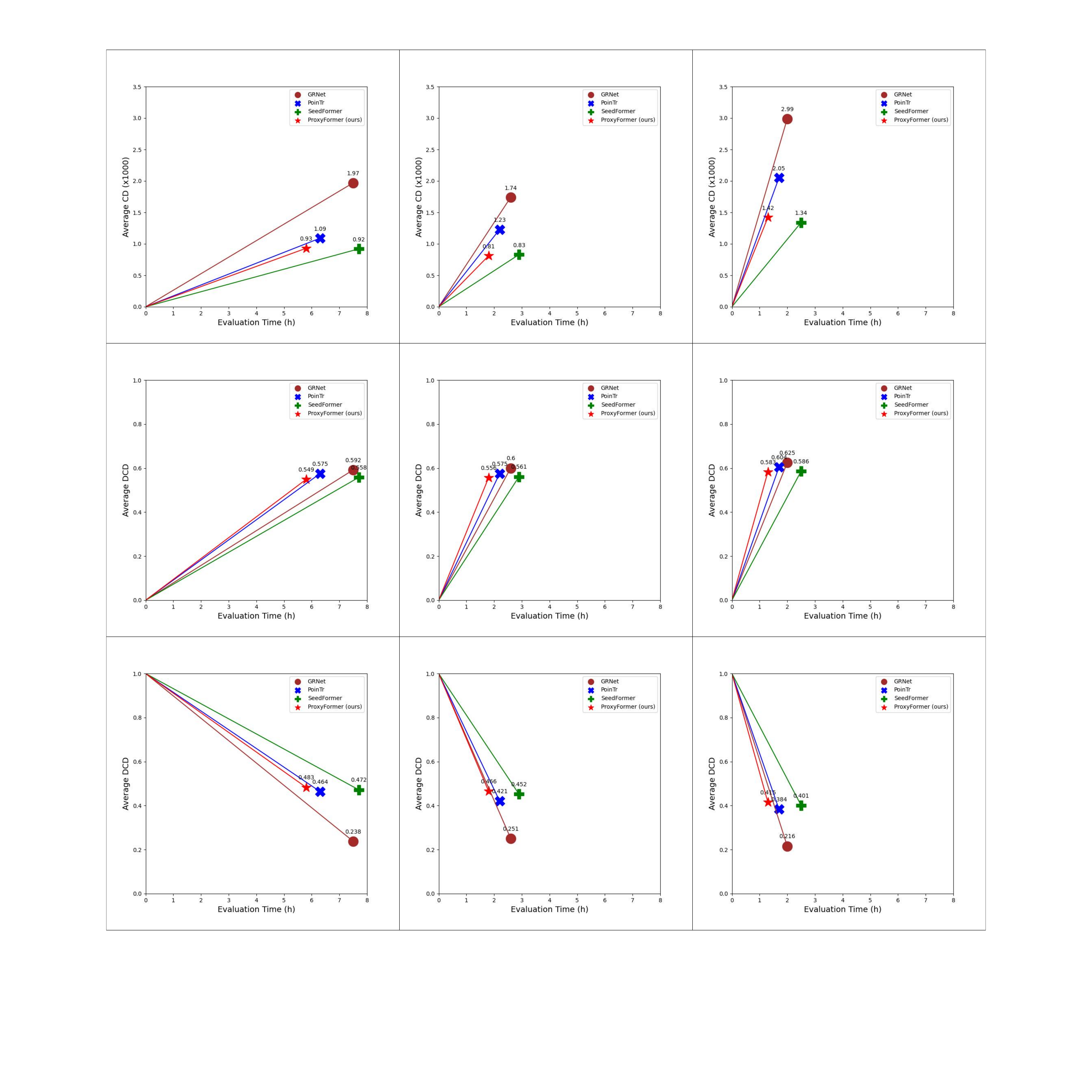}
        \caption{DCD vs. Eval. time on ShapeNet-55.}
        \label{Fig.4-4.}
      \end{subfigure}
      \begin{subfigure}{0.33\linewidth}
        \includegraphics[width=6.5cm]{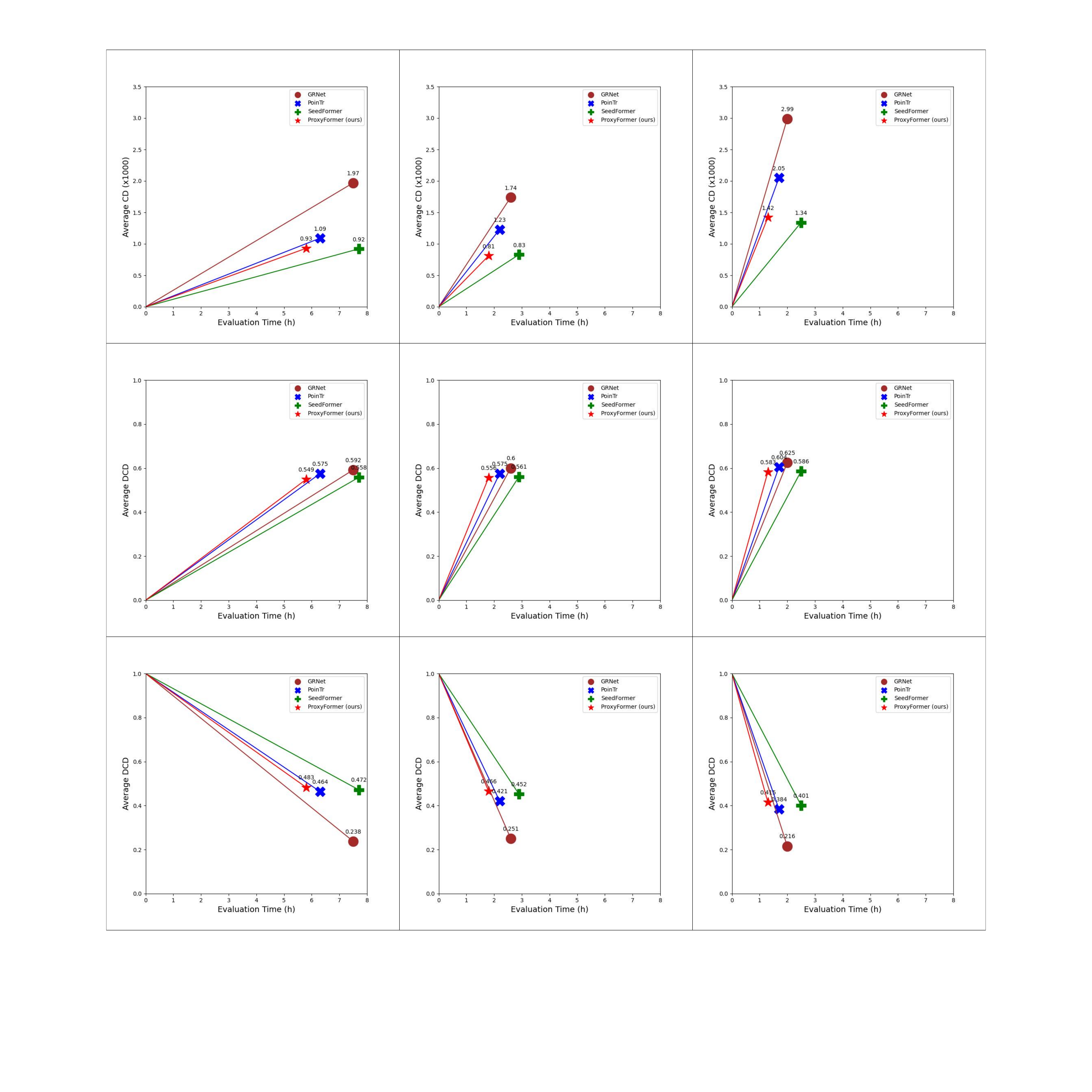}
        \caption{DCD vs. Eval. time on 34 seen categories.}
        \label{Fig.4-5.}
      \end{subfigure}
      \begin{subfigure}{0.33\linewidth}
        \includegraphics[width=6.5cm]{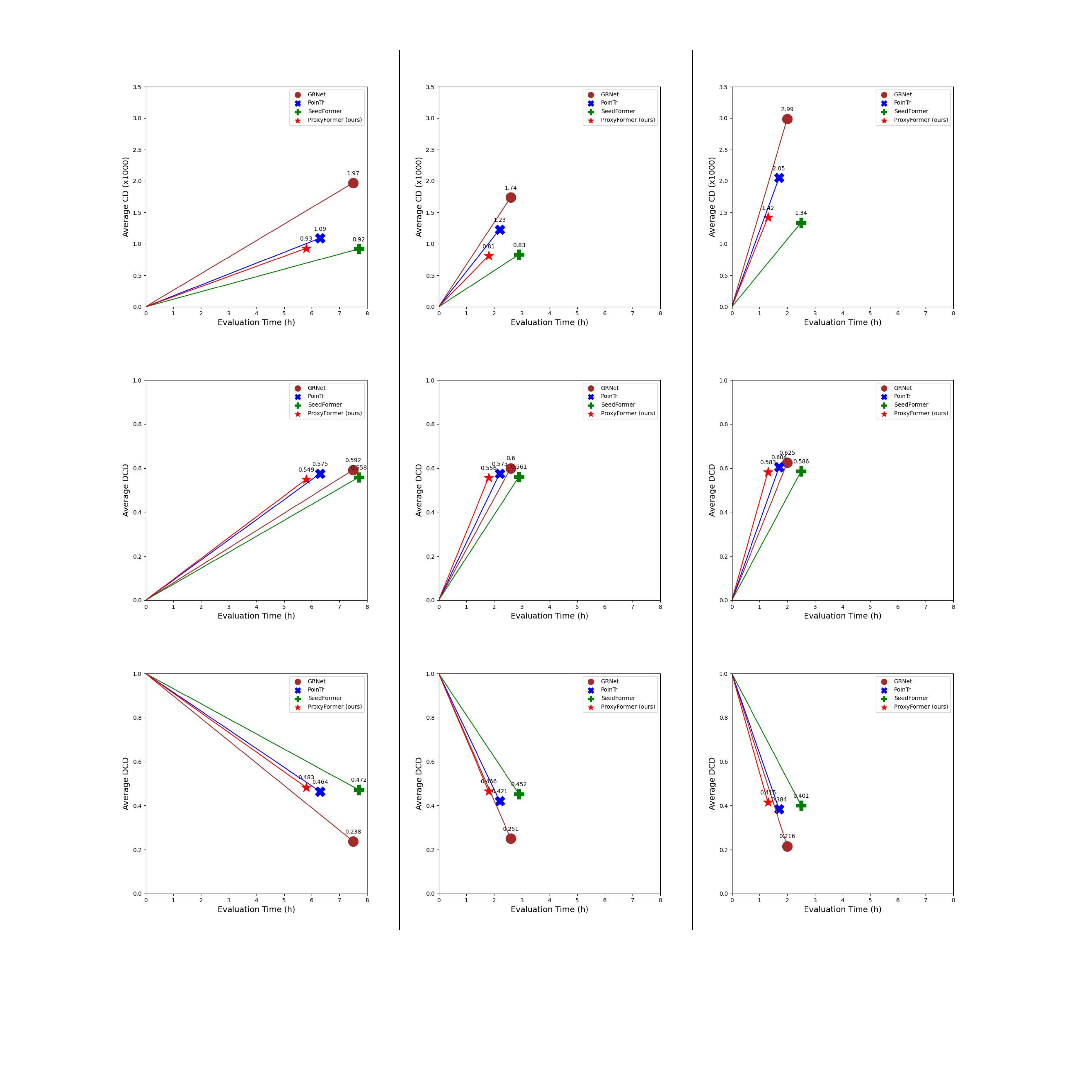}
        \caption{DCD vs. Eval. time on 21 novel categories.}
        \label{Fig.4-6.}
      \end{subfigure}
      \\
      \begin{subfigure}{0.33\linewidth}
        \includegraphics[width=6.5cm]{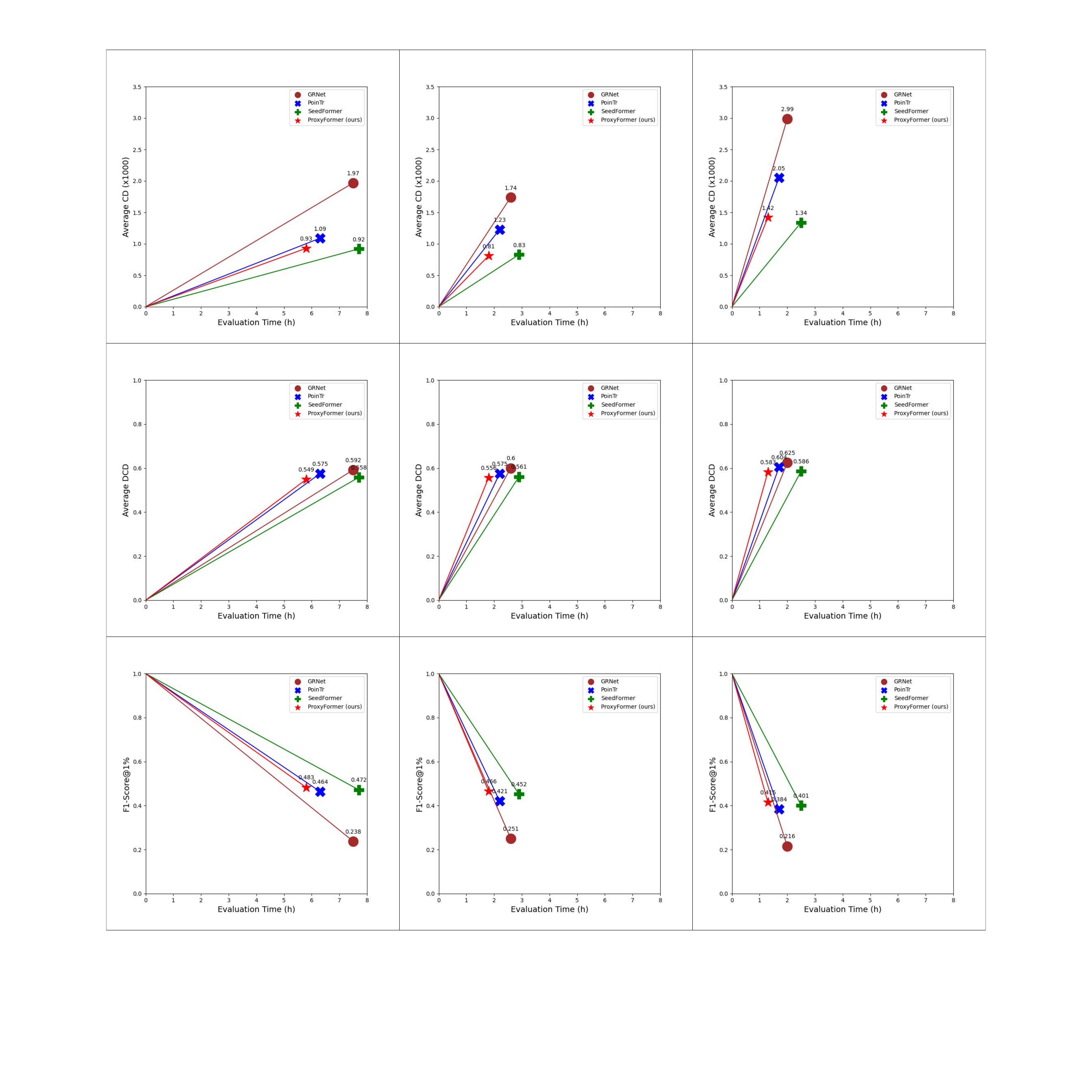}
        \caption{F1 vs. Eval. time on ShapeNet-55.}
        \label{Fig.4-7.}
      \end{subfigure}
      \begin{subfigure}{0.33\linewidth}
        \includegraphics[width=6.5cm]{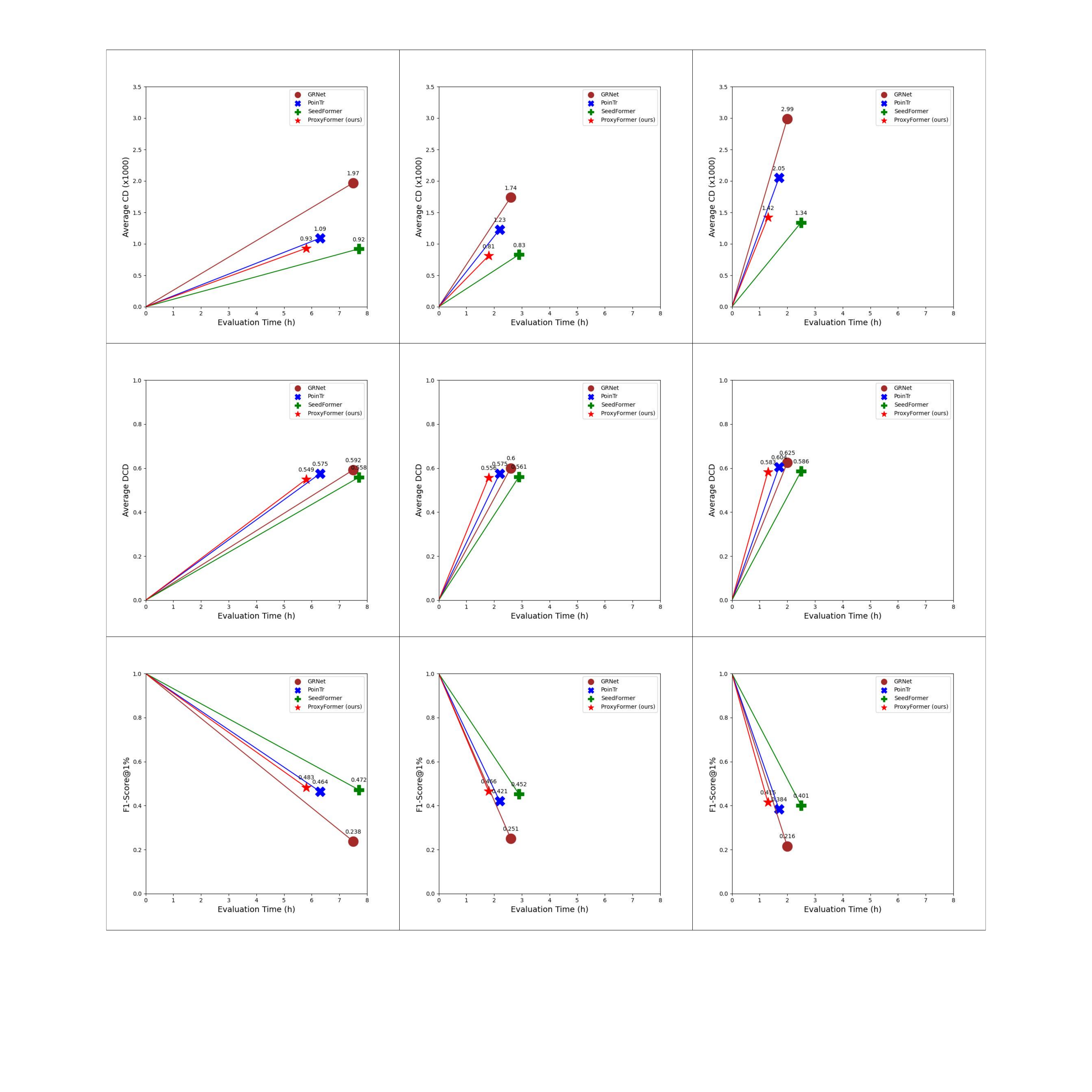}
        \caption{F1 vs. Eval. time on 34 seen categories.}
        \label{Fig.4-8.}
      \end{subfigure}
      \begin{subfigure}{0.33\linewidth}
        \includegraphics[width=6.5cm]{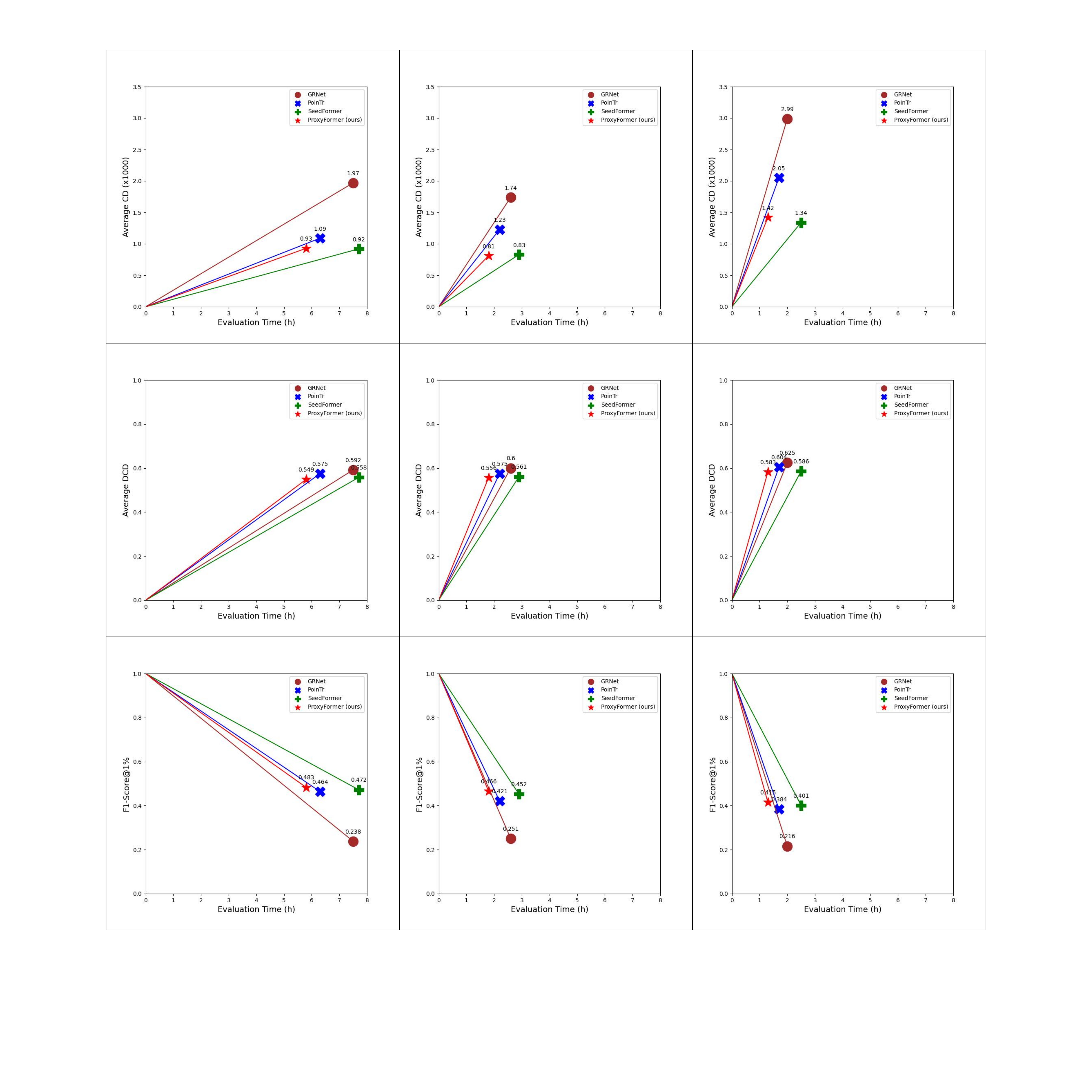}
        \caption{F1 vs. Eval. time on 21 novel categories.}
        \label{Fig.4-9.}
      \end{subfigure}
      \caption{Performance comparison. The first column represents the comparison results on ShapeNet-55. The second column represents the comparison results on 34 seen categories in ShapeNet-34, and the third column represents the comparison results on 21 novel categories in ShapeNet-34. The first row represents Eval. time \emph{vs.} CD. The second row represents Eval. time \emph{vs.} DCD. The third row represents Eval. time \emph{vs.} F1-Score@1\%. In order to observe the distance between the values and the coordinates (0, 0) or (0, 1) more clearly, we connect them, so the lines in the figure have no practical meaning and are only used for comparison.}
      \label{Fig.4.}
    \end{figure*}

\noindent{\bfseries Model performance Analysis.} In the main text, we have compared the complexity of our method with other methods on the PCN dataset, and here we show the evaluation time of each method on ShapeNet-55/34 (all methods are tested on the same device, \emph{i.e.} NVIDIA GeForce RTX 3090 graphics card). The ShapeNet-55 dataset contains $10518$ test models. The ShapeNet-34 contains $3400$ models in $34$ visible categories and $2305$ models in $21$ novel categories. Since $8$ viewpoints are fixed for each model during the testing process, and the average of the results of the $8$ viewpoints is used as the final CD value, the testing process is also time consuming. In Fig. \ref{Fig.4.}, we show the comparison of \emph{ProxyFormer} with other methods on ShapeNet-55/34 with 
evaluation time (Eval. time) \emph{vs.} CD, DCD and F1-Score. For Eval. time \emph{vs.} CD and DCD, the closer the value is to the origin of the coordinates, the better the model performance (that is, the smaller CD and DCD can be obtained while the inference speed is fast). For Eval. time \emph{vs.} F1-Score, the closer the value is to the coordinates $(0, 1)$, the better the model performance (that is, the higher the F1-Score can be obtained while the inference speed is fast). From the pictures, we can observe that our proposed \emph{ProxyFormer} has the fastest inference speed while achieving the smallest DCD and the highest F1-Score on ShapeNet-55/34.

    \begin{figure*}[h]
        \centering
        \includegraphics[width=18cm]{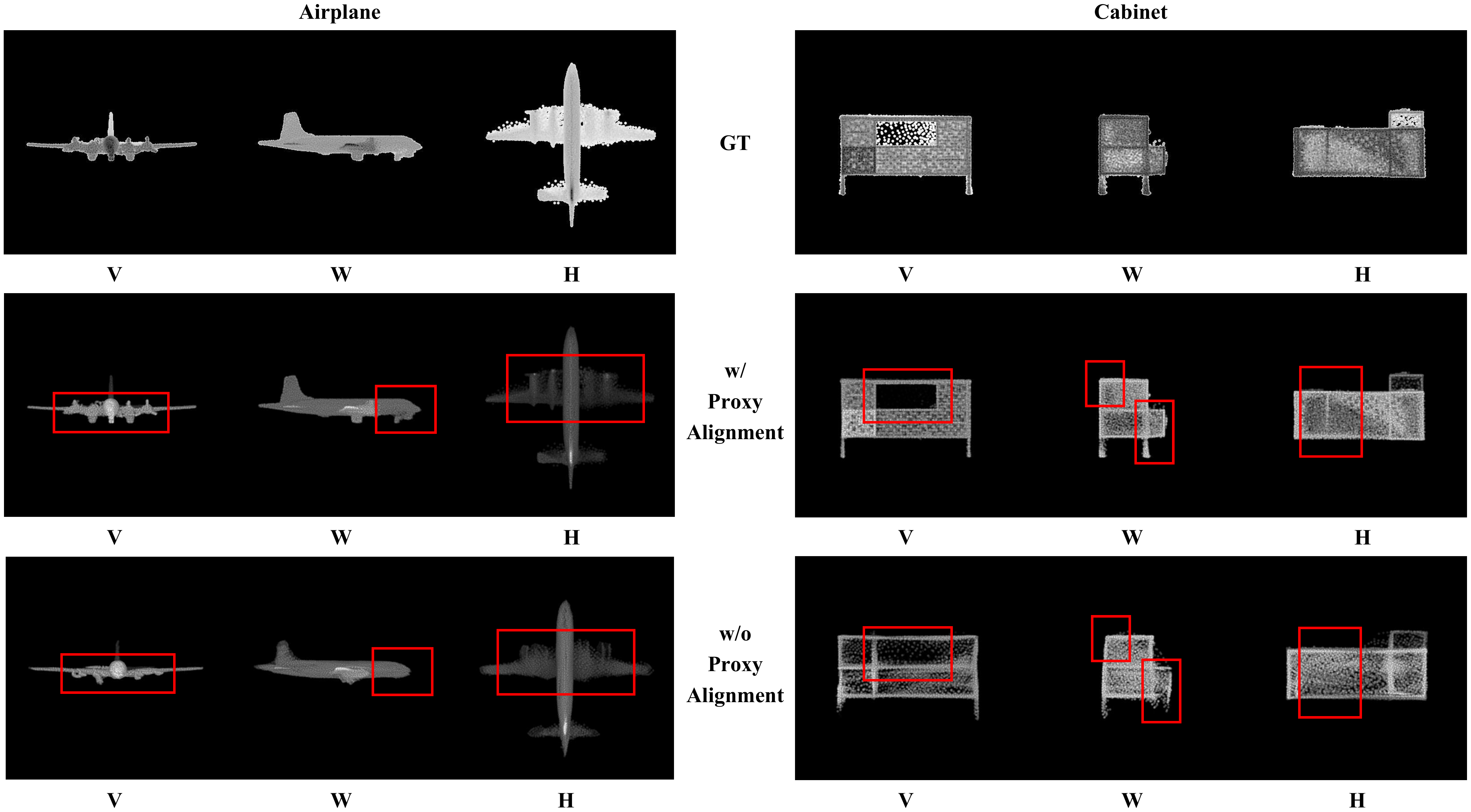}
        \\
        \includegraphics[width=18cm]{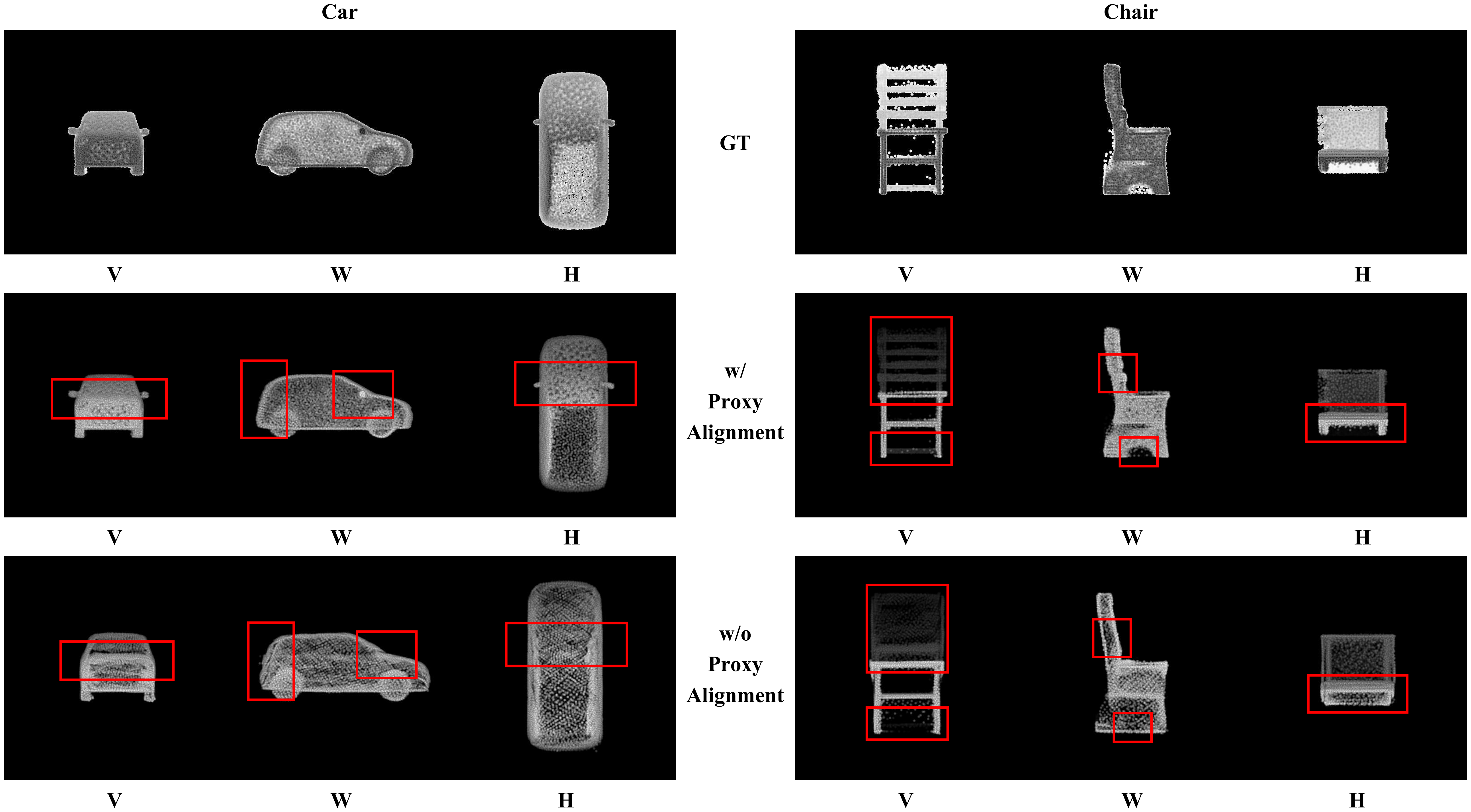}
        \\
        \centering
        \caption{Three-view drawings of GT, results with proxy alignment and results without proxy alignment. Each figure represents vertical plane (V), width plane (W) and horizontal plane (H) from left to right. The result six categories of objects in the PCN are listed separately, and the red box indicates where Proxy Alignment contributes the most.}
        \label{Fig.5.}
    \end{figure*}

    \begin{figure*}[h]
        \centering
        \includegraphics[width=18cm]{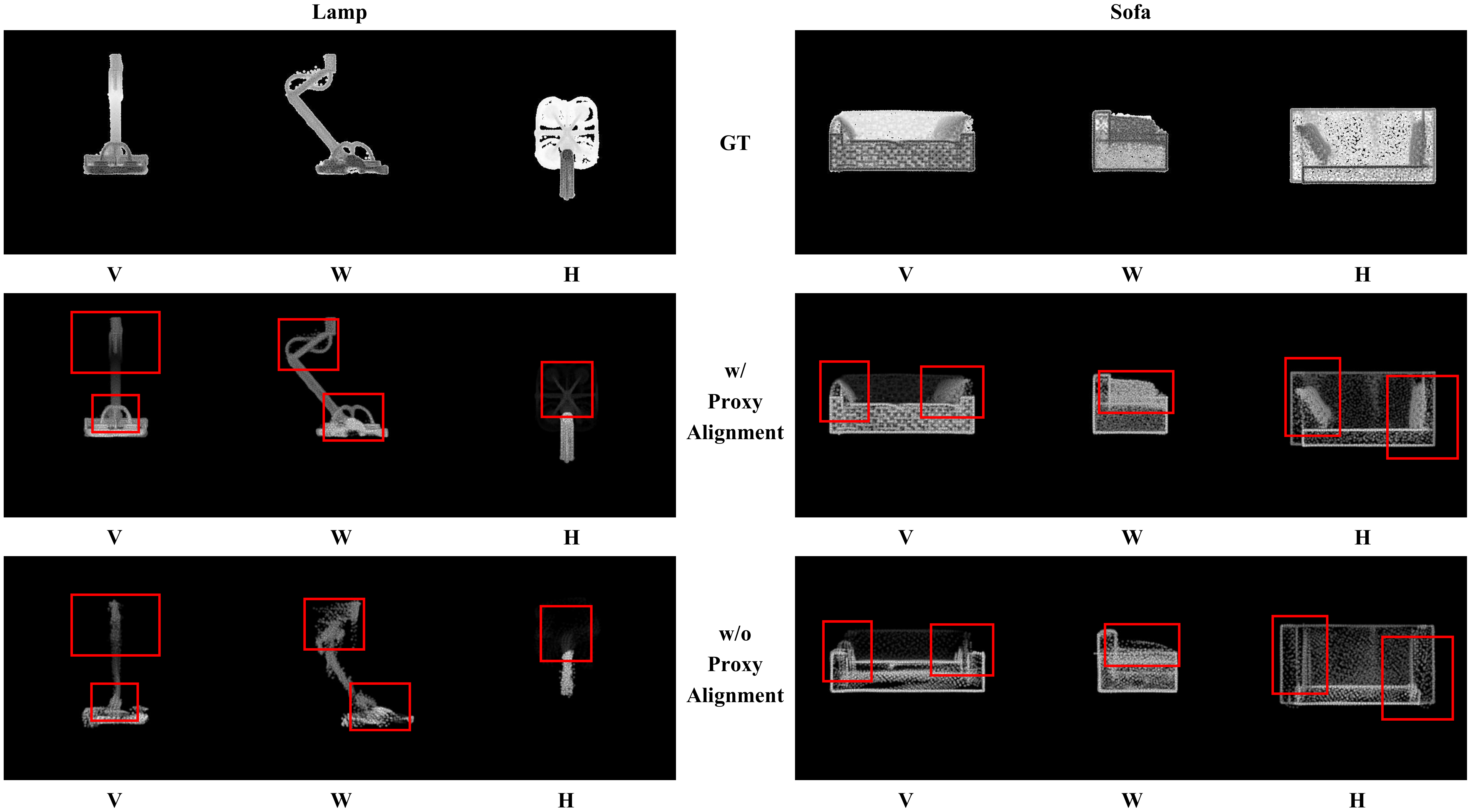}
        \\
        \includegraphics[width=18cm]{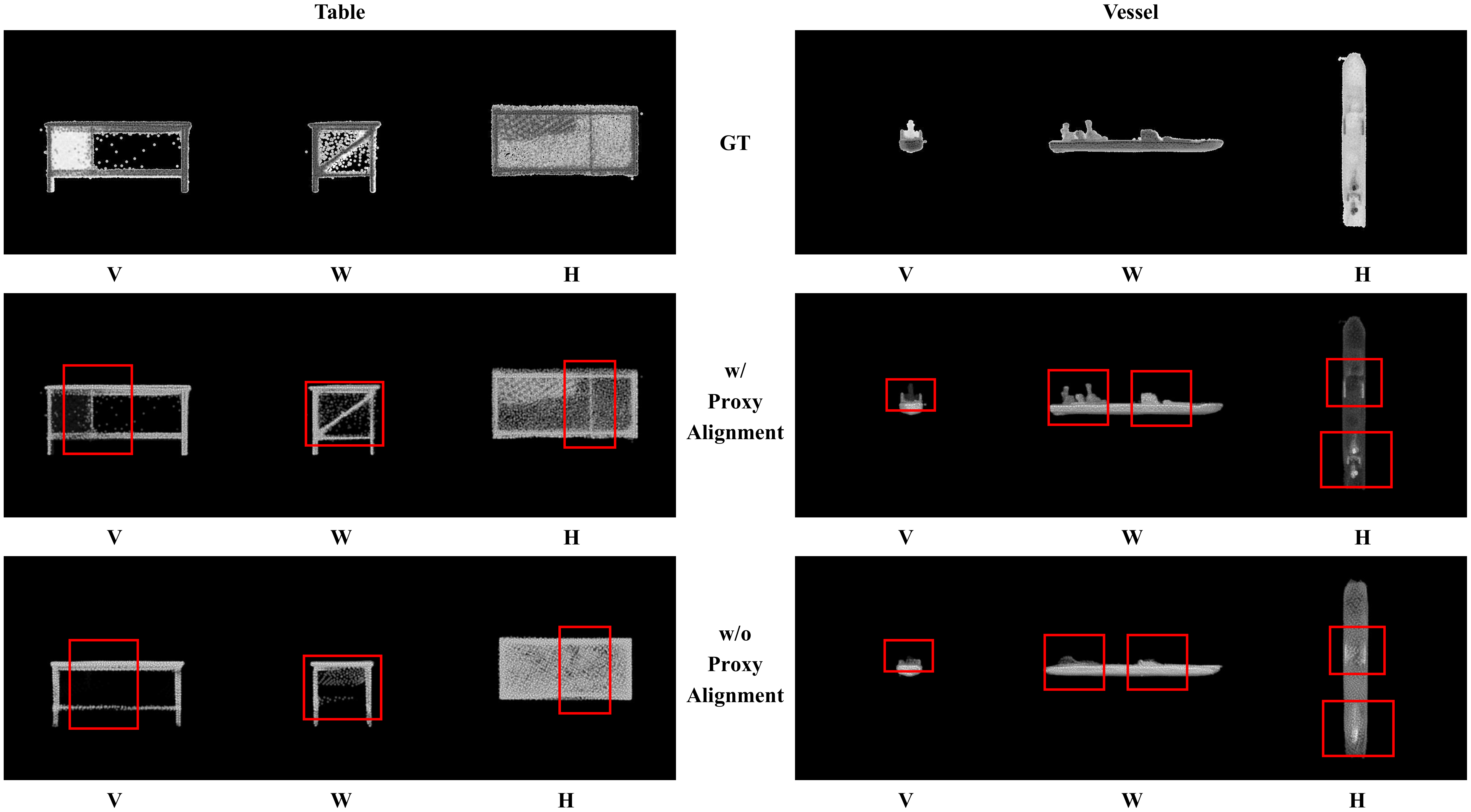}
        \\
        \centering
        \caption{Three-view drawings of GT, results with proxy alignment and results without proxy alignment. Each figure represents vertical plane (V), width plane (W) and horizontal plane (H) from left to right. The result six categories of objects in the PCN are listed separately, and the red box indicates where Proxy Alignment contributes the most.}
        \label{Fig.6.}
    \end{figure*}

% Our method achieves the best performance on many metrics on PCN dataset, ShapeNet-55, ShapeNet-34 and KITTI datasets. In Table \ref{tab.3.}, we list the number of parameters (Params), theoretical computation cost (FLOPs), the average chamfer distances (CD-Avg) and the average density-aware chamfer distances (DCD-Avg) of our method and other six methods. It can be seen that our method can obtain the lowest DCD-Avg while having the smallest FLOPs, which is second only to SeedFormer \cite{zhou2022seedformer} in terms of CD, and since the transformer decoder part was removed in \emph{ProxyFormer}, the number of parameters is also greatly reduced compared to PoinTr \cite{yu2021pointr}, which also shows that our method can better balance cost and performance.

\subsection{More experimental results}

\noindent{\bfseries Three-views drawings.} In Figs. \ref{Fig.5.} and \ref{Fig.6.}, we have drawn three views of \emph{ProxyFormer} on the PCN dataset, and we can see that proxy alignment plays a significant role in the point cloud completion task.

\begin{figure*}
        \centering
        \includegraphics[width=18cm]{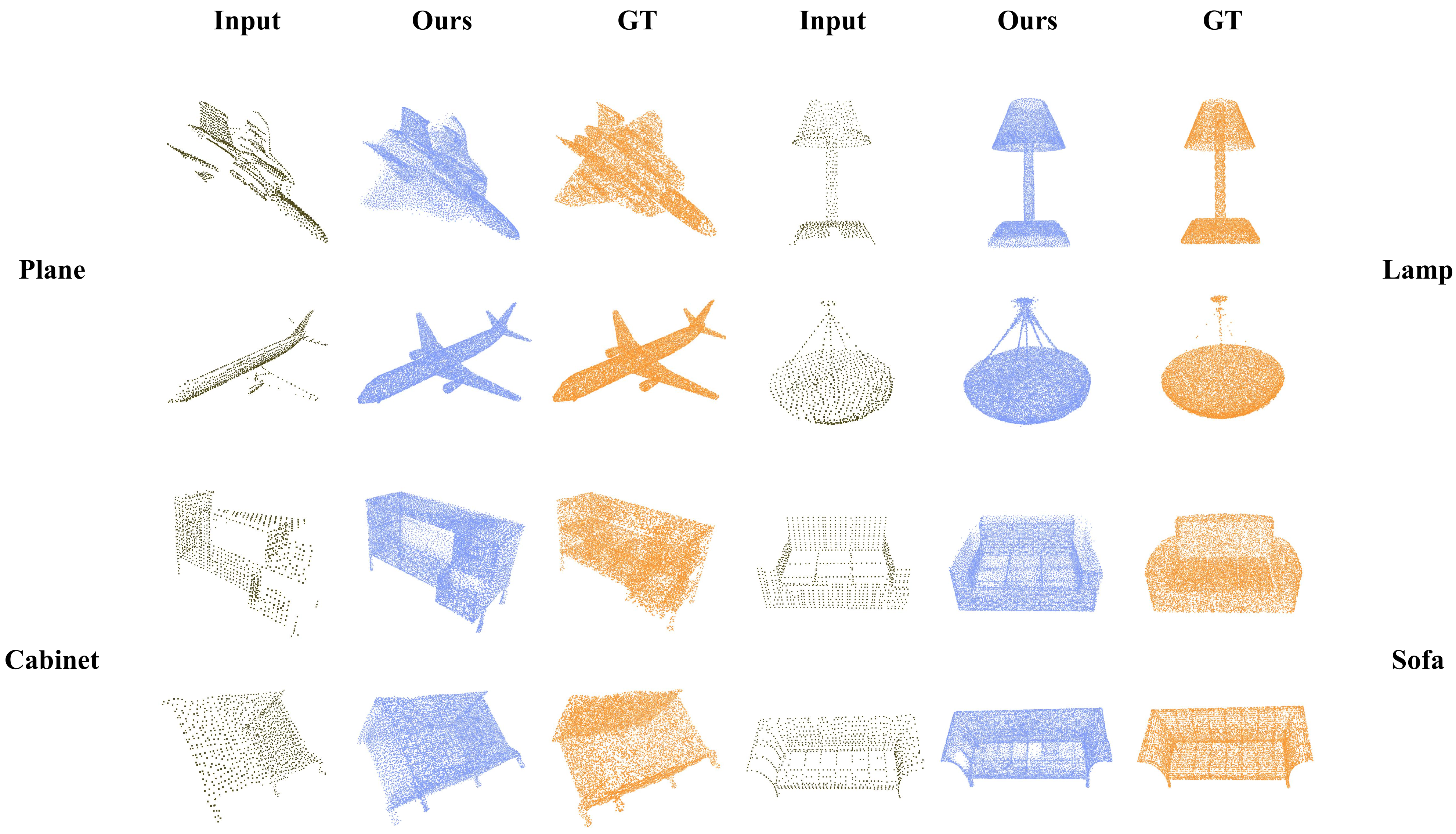}
        \\
        \includegraphics[width=18cm]{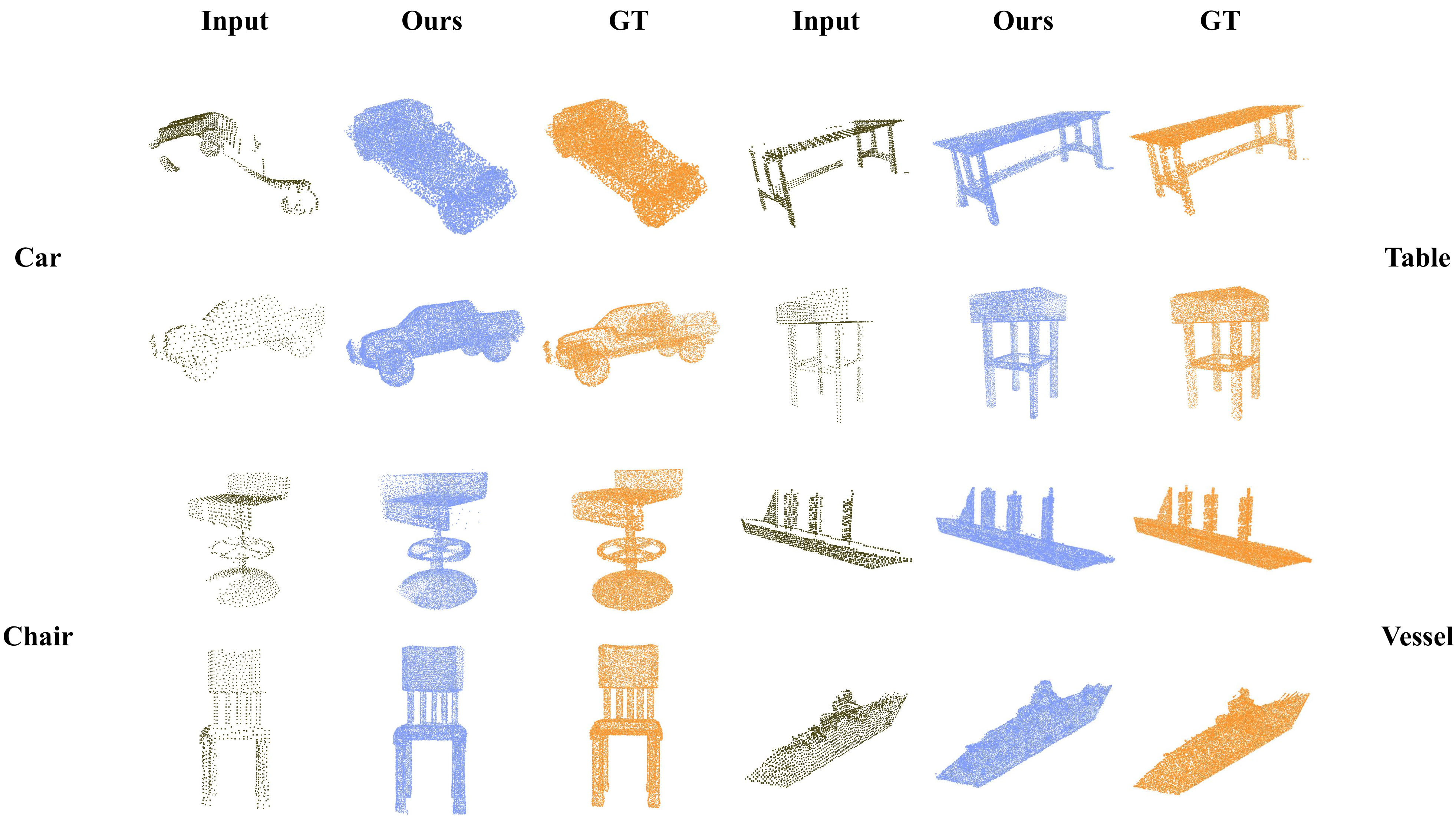}
        \centering
        \caption{More visual completion results of \emph{ProxyFormer} on PCN dataset. For each category in PCN, we show two results.}
        \label{Fig.7.}
    \end{figure*}

\noindent{\bfseries More qualitative results on PCN dataset.} We show more visualization results of \emph{ProxyFormer} on PCN dataset in Fig. \ref{Fig.7.}. From this, we can find that our method is more sensitive to the concentrated distribution of missing part, which makes it impossible to easily distinguish the shape of objects from the existing point clouds, such as the second cabinet and the first car. But our method can complete its shape well, and the final result is basically consistent with GT.

\begin{figure*}
		\centering
		\includegraphics[width=16cm]{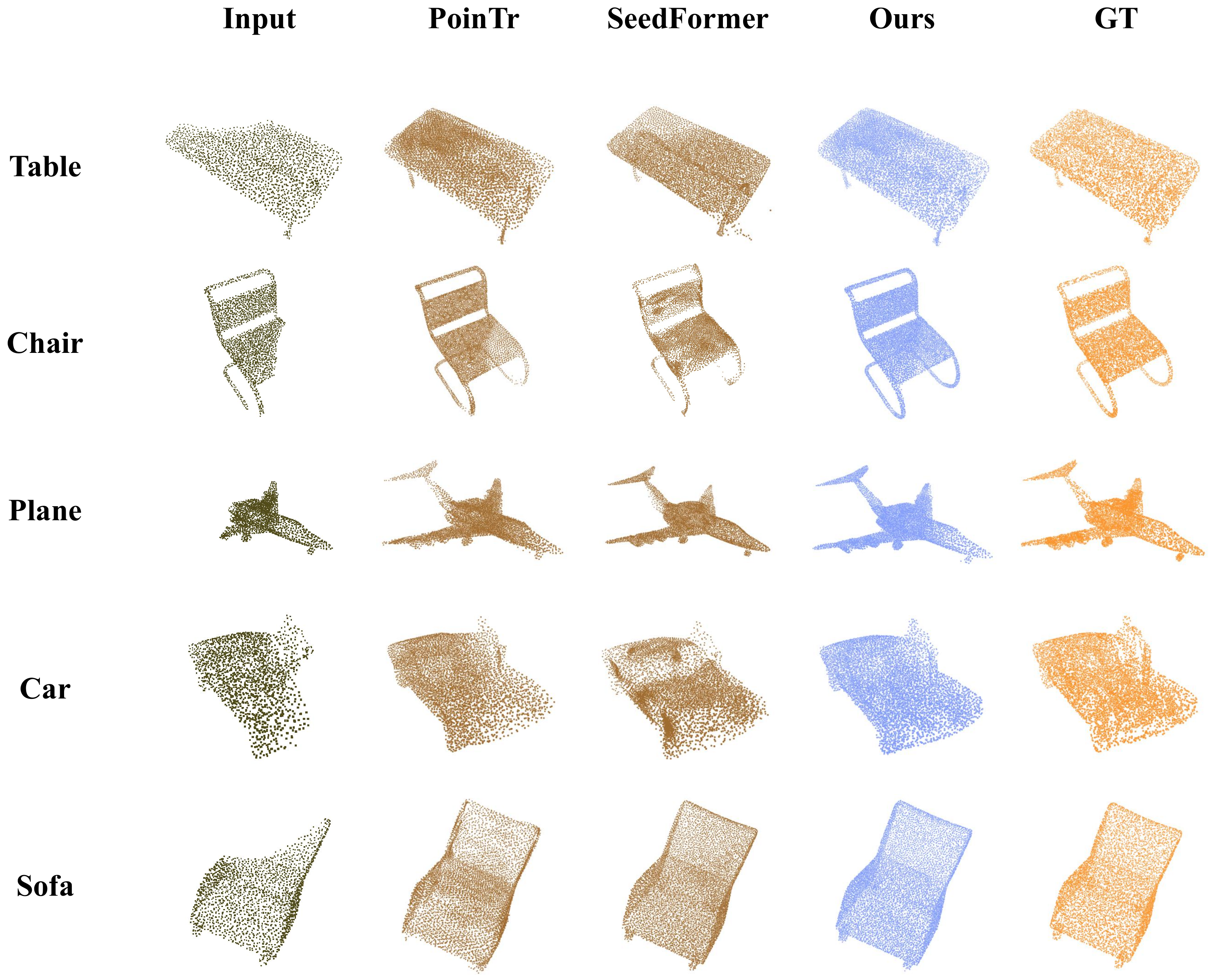}\\
		\caption{The visualization results of each method on ShapeNet-55, showing Table, Chair, Plane, Car and Sofa from top to bottom.}
		\label{Fig.8.}
	\end{figure*}

\begin{figure*}
        \centering
        \includegraphics[width=18cm]{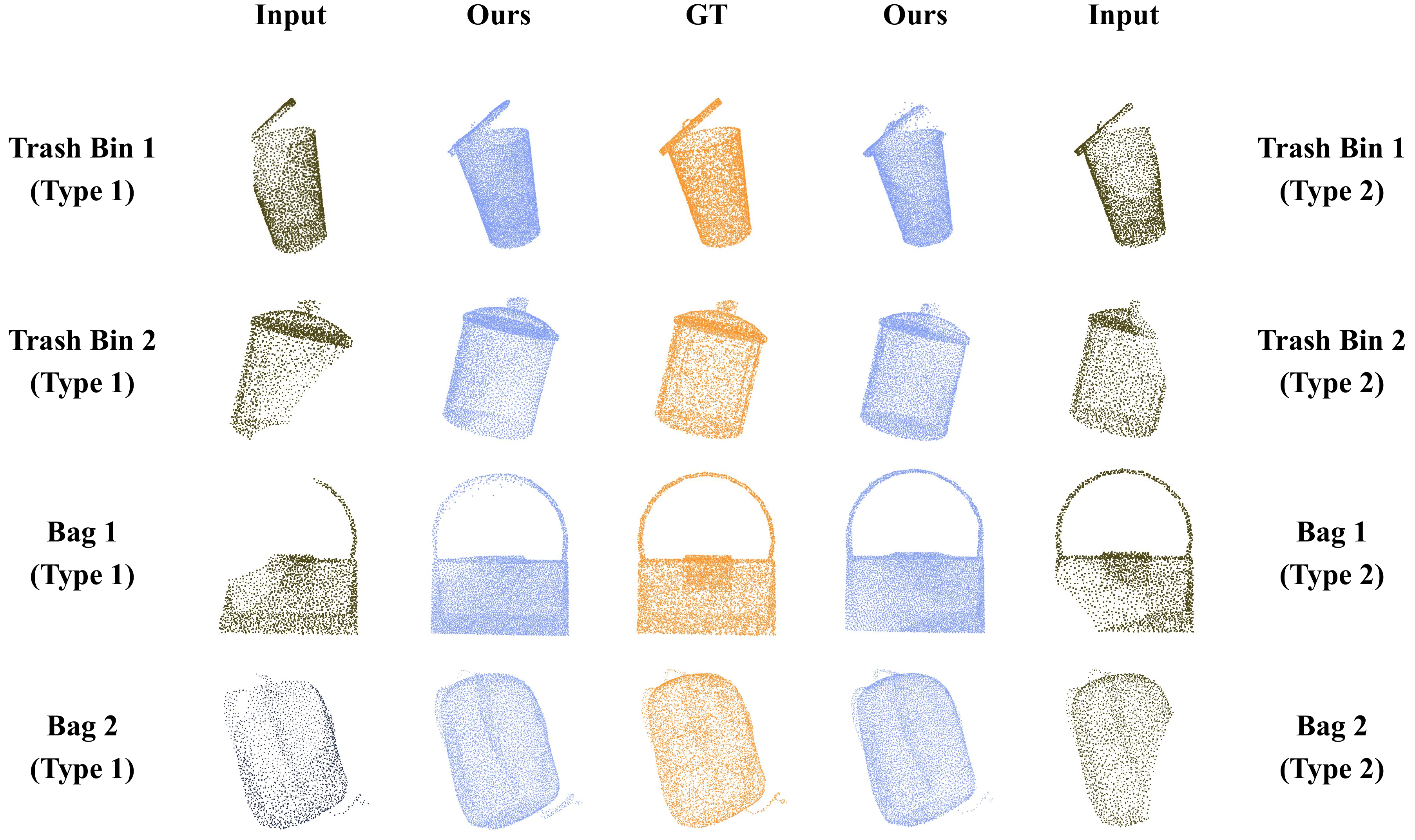}
        \\
        \includegraphics[width=18cm]{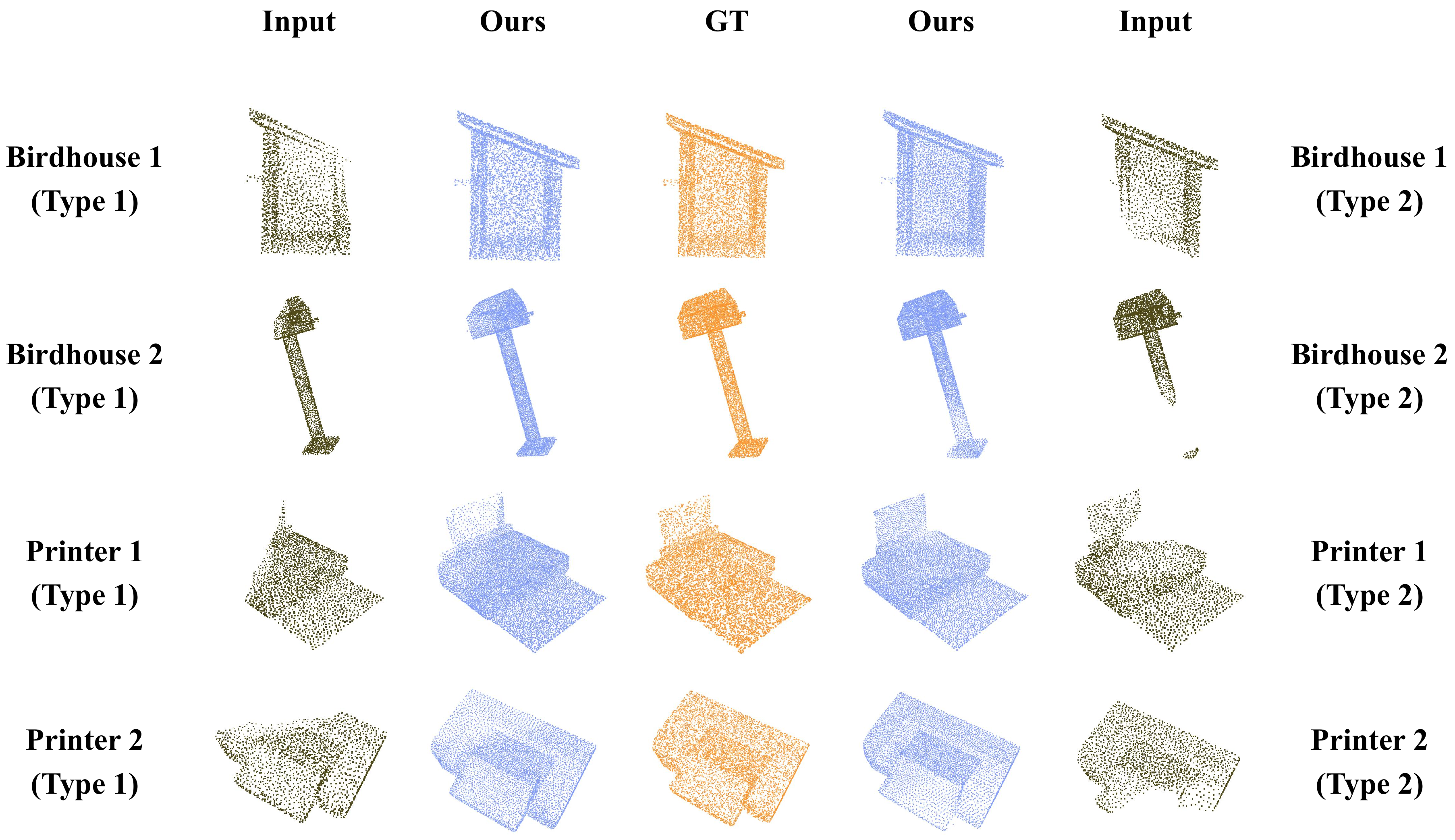}
        \centering
        \caption{More visual completion results of \emph{ProxyFormer} on ShapeNet-55. Each object contains two missing angles.}
        \label{Fig.9.}
    \end{figure*}

\noindent{\bfseries More qualitative results on ShapeNet-55/34.} In Figure \ref{Fig.8.}, we show the visualization results of point cloud completion on ShapNet-55 by PoinTr \cite{yu2021pointr}, SeedFormer \cite{zhou2022seedformer}, and \emph{ProxyFormer}. We can intuitively see that \emph{ProxyFormer} can better complete the point cloud completion task. For example, for the table in the first row, although the resulting shape of PoinTr is complete, it fails to generate the details of the bottom of the table well, and SeedFormer introduces some noise points in the completion process, which affects the final result. But our method is able to take both shape and detail into account. Another example is the car in the fourth row. The point cloud contains not only a car, but also a person. There is a problem with the results of PoinTr and SeedFormer completion, that is, the distribution of point clouds is uneven, while \emph{ProxyFormer} can better perceive the distribution of missing points, thereby generating a more reasonable complete point cloud. In Fig. \ref{Fig.9.}, we also show more visualization results of \emph{ProxyFormer} on the ShapNet-55. To better demonstrate the point cloud completion capability of our method, for each object, we show two different parts missing. For example, for the second bird house, we show both cases where the bottom and top are missing, and our method can easily get complete point clouds with local details.

\noindent{\bfseries Detailed quantitative results on ShapeNet-55/34.} We report complete results of our method on ShapeNet-55 in Tables \ref{tab.5.} and \ref{tab.6.} and results of novel categories on ShapeNet-34 in
Tables \ref{tab.7.} and \ref{tab.8.}. The models are tested under three difficulty levels: simple, moderate and hard. We can see that \emph{ProxyFormer} achieves lowest DCD on almost all categories on the three settings.

\begin{table*}
    \centering
    \caption{Detailed results on the ShapeNet-55 dataset, including Simple (S), Moderate (M) and Hard (H) three difficulties. For CD-$l_2$ ($\times 1000$), lower is better.}
    \label{tab.5.}
    \scalebox{0.83}{
    \begin{tabular}{l|ccc|ccc|ccc|ccc} 
    \hline
            & \multicolumn{3}{c|}{GRNet} & \multicolumn{3}{c|}{PoinTr}    & \multicolumn{3}{c|}{SeedFormer}                  & \multicolumn{3}{c}{Ours}                          \\
            & CD-S  & CD-M  & CD-H       & CD-S           & CD-M  & CD-H  & CD-S           & CD-M           & CD-H           & CD-S           & CD-M           & CD-H            \\ 
            \hline
            airplane    & 0.87~ & 0.87~ & 1.27~      & 0.27~          & 0.38~ & 0.69~ & 0.23~          & 0.35~          & 0.61~          & \textbf{0.21}~ & \textbf{0.28~} & \textbf{0.54~}  \\
            trash bin   & 1.69~ & 2.01~ & 3.48~      & 0.80~          & 1.15~ & 2.15~ & 0.73~          & 1.08~          & \textbf{1.94~} & \textbf{0.70~} & \textbf{1.04~} & 1.98~           \\
            bag         & 1.41~ & 1.70~ & 2.97~      & 0.53~          & 0.74~ & 1.51~ & 0.43~          & 0.67~          & \textbf{1.28~} & \textbf{0.39}~ & \textbf{0.65~} & 1.34~           \\
            basket      & 1.65~ & 1.84~ & 3.15~      & 0.73~          & 0.88~ & 1.82~ & 0.65~          & 0.83~          & 1.54~          & \textbf{0.64~} & \textbf{0.74~} & \textbf{1.40~}  \\
            bathtub     & 1.46~ & 1.73~ & 2.73~      & 0.64~          & 0.94~ & 1.68~ & \textbf{0.52~} & \textbf{0.82~} & \textbf{1.45~} & \textbf{0.52~} & 0.85~          & 1.47~           \\
            bed         & 1.64~ & 2.03~ & 3.70~      & 0.76~          & 1.10~ & 2.26~ & 0.63~          & \textbf{0.91~} & \textbf{1.89~} & \textbf{0.62}~ & \textbf{0.91~} & 2.04~           \\
            bench       & 1.03~ & 1.09~ & 1.71~      & 0.38~          & 0.52~ & 0.94~ & 0.32~          & 0.42~          & 0.84~          & \textbf{0.30~} & \textbf{0.39~} & \textbf{0.80~}  \\
            birdhouse   & 1.87~ & 2.40~ & 4.71~      & 0.98~          & 1.49~ & 3.13~ & \textbf{0.76~} & 1.30~          & \textbf{2.46~} & 0.83~          & \textbf{1.22~} & 2.67~           \\
            bookshelf   & 1.42~ & 1.71~ & 2.78~      & 0.71~          & 1.06~ & 1.93~ & 0.57~          & \textbf{0.84~} & \textbf{1.57~} & \textbf{0.55}~ & 0.92~          & 1.73~           \\
            bottle      & 1.05~ & 1.44~ & 2.67~      & 0.37~          & 0.74~ & 1.50~ & \textbf{0.31~} & 0.63~          & \textbf{1.21~} & \textbf{0.31~} & \textbf{0.60~} & 1.27~           \\
            bowl        & 1.60~ & 1.77~ & 2.99~      & 0.68~          & 0.78~ & 1.44~ & 0.56~          & 0.65~          & 1.18~          & \textbf{0.54~} & \textbf{0.62~} & \textbf{1.14~}  \\
            bus         & 1.06~ & 1.16~ & 1.48~      & 0.42~          & 0.55~ & 0.79~ & 0.42~          & 0.55~          & 0.73~          & \textbf{0.40~} & \textbf{0.46~} & \textbf{0.59~}  \\
            cabinet     & 1.27~ & 1.41~ & 2.09~      & 0.55~          & 0.66~ & 1.16~ & \textbf{0.57~} & 0.69~          & 1.05~          & 0.64~          & \textbf{0.59~} & \textbf{1.00~}  \\
            camera      & 2.14~ & 3.15~ & 6.09~      & 1.10~          & 2.03~ & 4.34~ & 0.83~          & \textbf{1.68~} & \textbf{3.45~} & \textbf{0.82~} & 1.77~          & 4.04~           \\
            can         & 1.58~ & 2.11~ & 3.81~      & 0.68~          & 1.19~ & 2.14~ & 0.58~          & 1.03~          & \textbf{1.79~} & \textbf{0.55}~ & \textbf{1.02~} & 1.86~           \\
            cap         & 1.17~ & 1.37~ & 3.05~      & 0.46~          & 0.62~ & 1.64~ & 0.33~          & \textbf{0.45~} & \textbf{1.18~} & \textbf{0.31}~ & 0.51~          & 1.28~           \\
            car         & 1.29~ & 1.48~ & 2.14~      & 0.64~          & 0.86~ & 1.25~ & 0.65~          & 0.86~          & 1.17~          & \textbf{0.61~} & \textbf{0.69~} & \textbf{1.03~}  \\
            cellphone   & 0.82~ & 0.91~ & 1.18~      & 0.32~          & 0.39~ & 0.60~ & 0.31~          & 0.40~          & 0.54~          & \textbf{0.26~} & \textbf{0.38~} & \textbf{0.45~}  \\
            chair       & 1.24~ & 1.56~ & 2.73~      & 0.49~          & 0.74~ & 1.63~ & 0.41~          & 0.65~          & \textbf{1.38~} & \textbf{0.40}~ & \textbf{0.61~} & 1.48~           \\
            clock       & 1.46~ & 1.66~ & 2.67~      & 0.62~          & 0.84~ & 1.65~ & 0.53~          & 0.74~          & \textbf{1.35~} & \textbf{0.50~} & \textbf{0.67~} & 1.45~           \\
            keyboard    & 0.74~ & 0.81~ & 1.09~      & 0.30~          & 0.39~ & 0.45~ & 0.28~          & 0.36~          & 0.45~          & \textbf{0.27}~ & \textbf{0.33~} & \textbf{0.43~}  \\
            dishwasher  & 1.43~ & 1.59~ & 2.53~      & \textbf{0.55~} & 0.69~ & 1.42~ & 0.56~          & 0.69~          & 1.30~          & 0.57~          & \textbf{0.61~} & \textbf{1.21~}  \\
            display     & 1.13~ & 1.38~ & 2.29~      & 0.48~          & 0.67~ & 1.33~ & \textbf{0.39~} & \textbf{0.59~} & \textbf{1.10~} & 0.46~          & 0.62~          & 1.18~           \\
            earphone    & 1.78~ & 2.18~ & 5.33~      & 0.81~          & 1.38~ & 3.78~ & 0.64~          & \textbf{1.04~} & \textbf{2.75~} & \textbf{0.62}~ & 1.12~          & 3.36~           \\
            faucet      & 1.81~ & 2.32~ & 4.91~      & 0.71~          & 1.42~ & 3.49~ & 0.55~          & \textbf{1.15~} & \textbf{2.63~} & \textbf{0.49~} & 1.16~          & 3.01~           \\
            filecabinet & 1.46~ & 1.71~ & 2.89~      & 0.63~          & 0.84~ & 1.69~ & \textbf{0.63~} & 0.84~          & 1.49~          & 0.74~          & \textbf{0.79~} & \textbf{1.45~}  \\
            guitar      & 0.44~ & 0.48~ & 0.76~      & 0.14~          & 0.21~ & 0.42~ & \textbf{0.13~} & \textbf{0.19~} & 0.32~          & 0.14~          & 0.20~          & \textbf{0.28~}  \\
            helmet      & 2.33~ & 3.18~ & 6.03~      & 0.99~          & 1.93~ & 4.22~ & 0.79~          & \textbf{1.52~} & \textbf{3.61~} & \textbf{0.76}~ & 1.60~          & 3.91~           \\
            jar         & 1.72~ & 2.37~ & 4.37~      & 0.77~          & 1.33~ & 2.87~ & 0.63~          & \textbf{1.13~} & \textbf{2.36~} & \textbf{0.62~} & \textbf{1.13~} & 2.69~           \\
            knife       & 0.72~ & 0.66~ & 0.96~      & 0.20~          & 0.33~ & 0.56~ & \textbf{0.15~} & 0.28~          & 0.45~          & \textbf{0.15}~ & \textbf{0.21~} & \textbf{0.43~}  \\
            lamp        & 1.68~ & 2.43~ & 5.17~      & 0.64~          & 1.40~ & 3.58~ & 0.45~          & 1.06~          & \textbf{2.67~} & \textbf{0.42~} & \textbf{1.05~} & 3.03~           \\
            laptop      & 0.83~ & 0.87~ & 1.28~      & 0.32~          & 0.34~ & 0.60~ & 0.32~          & 0.37~          & 0.55~          & \textbf{0.30}~ & \textbf{0.31~} & \textbf{0.42~}  \\
            loudspeaker & 1.75~ & 2.08~ & 3.45~      & 0.78~          & 1.16~ & 2.17~ & 0.67~          & \textbf{1.01~} & \textbf{1.80~} & \textbf{0.66}~ & 1.03~          & 1.89~           \\
            mailbox     & 1.15~ & 1.59~ & 3.42~      & 0.39~          & 0.78~ & 2.56~ & \textbf{0.30~} & 0.67~          & \textbf{2.04~} & \textbf{0.30}~ & \textbf{0.65~} & 2.17~           \\
            microphone  & 2.09~ & 2.76~ & 5.70~      & 0.70~          & 1.66~ & 4.48~ & 0.62~          & 1.61~          & \textbf{3.66~} & \textbf{0.59}~ & \textbf{1.58~} & 3.98~           \\
            microwaves  & 1.51~ & 1.72~ & 2.76~      & 0.67~          & 0.83~ & 1.82~ & 0.63~          & 0.79~          & 1.47~          & \textbf{0.61}~ & \textbf{0.69~} & \textbf{1.35~}  \\
            motorbike   & 1.38~ & 1.52~ & 2.26~      & 0.75~          & 1.10~ & 1.92~ & 0.68~          & \textbf{0.96~} & \textbf{1.44~} & \textbf{0.67}~ & \textbf{0.96~} & 1.49~           \\
            mug         & 1.75~ & 2.16~ & 3.79~      & 0.91~          & 1.17~ & 2.35~ & 0.79~          & 1.03~          & 2.06~          & \textbf{0.75~} & \textbf{0.92~} & \textbf{2.00~}  \\
            piano       & 1.53~ & 1.82~ & 3.21~      & 0.76~          & 1.06~ & 2.23~ & 0.62~          & 0.87~          & 1.79~          & \textbf{0.55}~ & \textbf{0.85~} & \textbf{1.73~}  \\
            pillow      & 1.42~ & 1.67~ & 3.04~      & 0.61~          & 0.82~ & 1.56~ & \textbf{0.48~} & 0.75~          & 1.41~          & 0.49~          & \textbf{0.71~} & \textbf{1.35~}  \\
            pistol      & 1.11~ & 1.06~ & 1.76~      & 0.43~          & 0.66~ & 1.30~ & 0.37~          & 0.56~          & 0.96~          & \textbf{0.34}~ & \textbf{0.52~} & \textbf{0.90~}  \\
            flowerpot   & 2.02~ & 2.48~ & 4.19~      & 1.01~          & 1.51~ & 2.77~ & 0.93~          & \textbf{1.30~} & \textbf{2.32~} & \textbf{0.89}~ & 1.39~          & 2.52~           \\
            printer     & 1.56~ & 2.38~ & 4.24~      & 0.73~          & 1.21~ & 2.47~ & 0.58~          & 1.11~          & 2.13~          & \textbf{0.53~} & \textbf{1.09~} & \textbf{2.08~}  \\
            remote      & 0.89~ & 1.05~ & 1.29~      & 0.36~          & 0.53~ & 0.71~ & 0.29~          & 0.46~          & 0.62~          & \textbf{0.20~} & \textbf{0.33~} & \textbf{0.54~}  \\
            rifle       & 0.83~ & 0.77~ & 1.16~      & 0.30~          & 0.45~ & 0.79~ & 0.27~          & 0.41~          & 0.66~          & \textbf{0.21~} & \textbf{0.32~} & \textbf{0.50~}  \\
            rocket      & 0.78~ & 0.92~ & 1.44~      & 0.23~          & 0.48~ & 0.99~ & 0.21~          & 0.46~          & 0.83~          & \textbf{0.21~} & \textbf{0.38~} & \textbf{0.80~}  \\
            skateboard  & 0.82~ & 0.87~ & 1.24~      & 0.28~          & 0.38~ & 0.62~ & 0.23~          & 0.32~          & 0.62~          & \textbf{0.19~} & \textbf{0.28~} & \textbf{0.56~}  \\
            sofa        & 1.35~ & 1.45~ & 2.32~      & 0.56~          & 0.67~ & 1.14~ & 0.50~          & 0.62~          & 1.02~          & \textbf{0.49~} & \textbf{0.57~} & \textbf{1.01~}  \\
            stove       & 1.46~ & 1.72~ & 3.22~      & 0.63~          & 0.92~ & 1.73~ & \textbf{0.59~} & \textbf{0.87~} & \textbf{1.49~} & 0.68~          & 0.88~          & 1.67~           \\
            table       & 1.15~ & 1.33~ & 2.33~      & 0.46~          & 0.64~ & 1.31~ & 0.41~          & 0.58~          & 1.18~          & \textbf{0.39~} & \textbf{0.48~} & \textbf{1.06~}  \\
            telephone   & 0.81~ & 0.89~ & 1.18~      & 0.31~          & 0.38~ & 0.59~ & 0.31~          & 0.39~          & 0.55~          & \textbf{0.28~} & \textbf{0.32~} & \textbf{0.48~}  \\
            tower       & 1.26~ & 1.69~ & 3.06~      & 0.55~          & 0.90~ & 1.95~ & 0.47~          & 0.84~          & \textbf{1.65~} & \textbf{0.46~} & \textbf{0.82~} & 1.67~           \\
            train       & 1.09~ & 1.14~ & 1.61~      & 0.50~          & 0.70~ & 1.12~ & 0.51~          & 0.66~          & 1.01~          & \textbf{0.49~} & \textbf{0.61~} & \textbf{0.97~}  \\
            watercraft  & 1.09~ & 1.12~ & 1.65~      & 0.41~          & 0.62~ & 1.07~ & \textbf{0.35~} & \textbf{0.56~} & \textbf{0.92~} & 0.44~          & 0.62~          & 1.04~           \\
            washer      & 1.72~ & 2.05~ & 4.19~      & 0.75~          & 1.06~ & 2.44~ & 0.64~          & \textbf{0.91~} & \textbf{2.04~} & \textbf{0.63~} & 0.94~          & 2.26~           \\ 
            \hline
            mean        & 1.35~ & 1.63~ & 2.86~      & 0.58~          & 0.88~ & 1.80~ & 0.50~          & 0.77~          & \textbf{1.49~} & \textbf{0.49~} & \textbf{0.75}~ & 1.55~           \\
            \hline
            \end{tabular}}
        \end{table*}

\begin{table*}
    \centering
    \caption{Detailed results on the ShapeNet-55 dataset, including Simple (S), Moderate (M) and Hard (H) three difficulties,  For DCD, lower is better.}
    \label{tab.6.}
    \scalebox{0.8}{
    \begin{tabular}{l|ccc|ccc|ccc|ccc} 
    \hline
                & \multicolumn{3}{c|}{GRNet} & \multicolumn{3}{c|}{PoinTr} & \multicolumn{3}{c|}{SeedFormer}   & \multicolumn{3}{c}{Ours}                             \\
                & DCD-S  & DCD-M  & DCD-H    & DCD-S  & DCD-M  & DCD-H     & DCD-S  & DCD-M  & DCD-H           & DCD-S           & DCD-M           & DCD-H            \\ 
    \hline
    airplane    & 0.520~ & 0.559~ & 0.614~   & 0.487~ & 0.524~ & 0.608~    & 0.478~ & 0.505~ & 0.574~          & \textbf{0.475~} & \textbf{0.496~} & \textbf{0.565~}  \\
    trash bin   & 0.586~ & 0.616~ & 0.705~   & 0.557~ & 0.599~ & 0.679~    & 0.547~ & 0.588~ & \textbf{0.648~} & \textbf{0.544~} & \textbf{0.578~} & 0.653~           \\
    bag         & 0.518~ & 0.563~ & 0.653~   & 0.510~ & 0.549~ & 0.628~    & 0.503~ & 0.546~ & 0.590~          & \textbf{0.483~} & \textbf{0.515~} & \textbf{0.588~}  \\
    basket      & 0.559~ & 0.593~ & 0.648~   & 0.533~ & 0.573~ & 0.630~    & 0.530~ & 0.558~ & \textbf{0.609~} & \textbf{0.522~} & \textbf{0.548~} & 0.616~           \\
    bathtub     & 0.503~ & 0.553~ & 0.646~   & 0.499~ & 0.544~ & 0.619~    & 0.498~ & 0.523~ & 0.609~          & \textbf{0.492~} & \textbf{0.519~} & \textbf{0.589~}  \\
    bed         & 0.553~ & 0.620~ & 0.694~   & 0.544~ & 0.589~ & 0.661~    & 0.537~ & 0.583~ & 0.637~          & \textbf{0.528~} & \textbf{0.554~} & \textbf{0.622~}  \\
    bench       & 0.528~ & 0.546~ & 0.607~   & 0.507~ & 0.531~ & 0.589~    & 0.506~ & 0.504~ & 0.543~          & \textbf{0.483~} & \textbf{0.494~} & \textbf{0.539~}  \\
    birdhouse   & 0.576~ & 0.617~ & 0.722~   & 0.562~ & 0.597~ & 0.697~    & 0.554~ & 0.585~ & \textbf{0.650~} & \textbf{0.538~} & \textbf{0.579~} & 0.658~           \\
    bookshelf   & 0.563~ & 0.589~ & 0.678~   & 0.541~ & 0.569~ & 0.655~    & 0.523~ & 0.562~ & \textbf{0.623~} & \textbf{0.519~} & \textbf{0.554~} & 0.631~           \\
    bottle      & 0.505~ & 0.546~ & 0.635~   & 0.485~ & 0.533~ & 0.628~    & 0.458~ & 0.516~ & 0.606~          & \textbf{0.455~} & \textbf{0.508~} & \textbf{0.605~}  \\
    bowl        & 0.545~ & 0.591~ & 0.676~   & 0.531~ & 0.560~ & 0.641~    & 0.524~ & 0.529~ & 0.613~          & \textbf{0.507~} & \textbf{0.529~} & \textbf{0.602~}  \\
    bus         & 0.533~ & 0.559~ & 0.606~   & 0.525~ & 0.547~ & 0.599~    & 0.521~ & 0.525~ & 0.584~          & \textbf{0.499~} & \textbf{0.523~} & \textbf{0.570~}  \\
    cabinet     & 0.590~ & 0.593~ & 0.643~   & 0.557~ & 0.577~ & 0.626~    & 0.532~ & 0.548~ & \textbf{0.595~} & \textbf{0.521~} & \textbf{0.544~} & 0.596~           \\
    camera      & 0.583~ & 0.647~ & 0.728~   & 0.558~ & 0.615~ & 0.701~    & 0.547~ & 0.608~ & \textbf{0.670~} & \textbf{0.535~} & \textbf{0.584~} & 0.671~           \\
    can         & 0.549~ & 0.599~ & 0.651~   & 0.527~ & 0.571~ & 0.649~    & 0.511~ & 0.550~ & 0.621~          & \textbf{0.500~} & \textbf{0.542~} & \textbf{0.616~}  \\
    cap         & 0.528~ & 0.544~ & 0.640~   & 0.495~ & 0.541~ & 0.638~    & 0.491~ & 0.539~ & \textbf{0.609~} & \textbf{0.474~} & \textbf{0.510~} & 0.612~           \\
    car         & 0.614~ & 0.660~ & 0.683~   & 0.581~ & 0.631~ & 0.673~    & 0.566~ & 0.607~ & 0.661~          & \textbf{0.565~} & \textbf{0.592~} & \textbf{0.644~}  \\
    cellphone   & 0.508~ & 0.509~ & 0.568~   & 0.483~ & 0.506~ & 0.541~    & 0.482~ & 0.497~ & \textbf{0.520~} & \textbf{0.458~} & \textbf{0.480~} & 0.521~           \\
    chair       & 0.526~ & 0.559~ & 0.623~   & 0.505~ & 0.555~ & 0.620~    & 0.501~ & 0.527~ & 0.609~          & \textbf{0.488~} & \textbf{0.523~} & \textbf{0.606~}  \\
    clock       & 0.543~ & 0.578~ & 0.657~   & 0.528~ & 0.561~ & 0.634~    & 0.515~ & 0.554~ & \textbf{0.594~} & \textbf{0.502~} & \textbf{0.531~} & 0.595~           \\
    keyboard    & 0.515~ & 0.528~ & 0.553~   & 0.499~ & 0.511~ & 0.541~    & 0.477~ & 0.498~ & \textbf{0.510~} & \textbf{0.473~} & \textbf{0.483~} & 0.515~           \\
    dishwasher  & 0.566~ & 0.600~ & 0.643~   & 0.540~ & 0.568~ & 0.629~    & 0.514~ & 0.536~ & \textbf{0.593~} & \textbf{0.508~} & \textbf{0.535~} & 0.596~           \\
    display     & 0.532~ & 0.544~ & 0.595~   & 0.517~ & 0.544~ & 0.587~    & 0.483~ & 0.519~ & \textbf{0.554~} & \textbf{0.483~} & \textbf{0.508~} & 0.560~           \\
    earphone    & 0.579~ & 0.609~ & 0.708~   & 0.555~ & 0.605~ & 0.694~    & 0.539~ & 0.603~ & 0.669~          & \textbf{0.529~} & \textbf{0.571~} & \textbf{0.675~}  \\
    faucet      & 0.512~ & 0.602~ & 0.681~   & 0.502~ & 0.573~ & 0.687~    & 0.495~ & 0.565~ & \textbf{0.663~} & \textbf{0.483~} & \textbf{0.548~} & 0.665~           \\
    filecabinet & 0.580~ & 0.615~ & 0.662~   & 0.559~ & 0.579~ & 0.644~    & 0.529~ & 0.562~ & 0.619~          & \textbf{0.529~} & \textbf{0.556~} & \textbf{0.616~}  \\
    guitar      & 0.499~ & 0.538~ & 0.616~   & 0.488~ & 0.514~ & 0.602~    & 0.459~ & 0.495~ & \textbf{0.571~} & \textbf{0.456~} & \textbf{0.493~} & 0.578~           \\
    helmet      & 0.591~ & 0.613~ & 0.699~   & 0.567~ & 0.608~ & 0.695~    & 0.551~ & 0.597~ & \textbf{0.668~} & \textbf{0.538~} & \textbf{0.586~} & 0.676~           \\
    jar         & 0.574~ & 0.598~ & 0.679~   & 0.541~ & 0.590~ & 0.669~    & 0.516~ & 0.567~ & \textbf{0.650~} & \textbf{0.512~} & \textbf{0.565~} & 0.660~           \\
    knife       & 0.486~ & 0.571~ & 0.651~   & 0.470~ & 0.544~ & 0.643~    & 0.464~ & 0.533~ & \textbf{0.606~} & \textbf{0.461~} & \textbf{0.518~} & 0.613~           \\
    lamp        & 0.526~ & 0.587~ & 0.717~   & 0.520~ & 0.581~ & 0.689~    & 0.510~ & 0.578~ & 0.677~          & \textbf{0.488~} & \textbf{0.553~} & \textbf{0.657~}  \\
    laptop      & 0.501~ & 0.538~ & 0.558~   & 0.492~ & 0.510~ & 0.557~    & 0.488~ & 0.506~ & 0.548~          & \textbf{0.473~} & \textbf{0.488~} & \textbf{0.532~}  \\
    loudspeaker & 0.561~ & 0.589~ & 0.669~   & 0.548~ & 0.583~ & 0.649~    & 0.544~ & 0.566~ & 0.622~          & \textbf{0.526~} & \textbf{0.556~} & \textbf{0.620~}  \\
    mailbox     & 0.499~ & 0.543~ & 0.674~   & 0.477~ & 0.540~ & 0.667~    & 0.465~ & 0.520~ & \textbf{0.620~} & \textbf{0.458~} & \textbf{0.518~} & 0.628~           \\
    microphone  & 0.546~ & 0.622~ & 0.709~   & 0.511~ & 0.587~ & 0.701~    & 0.503~ & 0.579~ & 0.681~          & \textbf{0.490~} & \textbf{0.560~} & \textbf{0.674~}  \\
    microwaves  & 0.559~ & 0.580~ & 0.634~   & 0.546~ & 0.566~ & 0.621~    & 0.541~ & 0.544~ & 0.614~          & \textbf{0.522~} & \textbf{0.544~} & \textbf{0.597~}  \\
    motorbike   & 0.653~ & 0.690~ & 0.714~   & 0.623~ & 0.657~ & 0.709~    & 0.614~ & 0.652~ & \textbf{0.687~} & \textbf{0.596~} & \textbf{0.627~} & 0.695~           \\
    mug         & 0.582~ & 0.618~ & 0.706~   & 0.560~ & 0.588~ & 0.678~    & 0.536~ & 0.585~ & 0.657~          & \textbf{0.535~} & \textbf{0.569~} & \textbf{0.644~}  \\
    piano       & 0.550~ & 0.585~ & 0.658~   & 0.538~ & 0.573~ & 0.632~    & 0.537~ & 0.552~ & 0.611~          & \textbf{0.519~} & \textbf{0.544~} & \textbf{0.602~}  \\
    pillow      & 0.495~ & 0.529~ & 0.665~   & 0.494~ & 0.529~ & 0.633~    & 0.485~ & 0.518~ & 0.613~          & \textbf{0.474~} & \textbf{0.510~} & \textbf{0.595~}  \\
    pistol      & 0.557~ & 0.586~ & 0.668~   & 0.529~ & 0.571~ & 0.655~    & 0.509~ & 0.570~ & \textbf{0.620~} & \textbf{0.506~} & \textbf{0.545~} & 0.627~           \\
    flowerpot   & 0.608~ & 0.626~ & 0.701~   & 0.596~ & 0.622~ & 0.697~    & 0.591~ & 0.609~ & \textbf{0.667~} & \textbf{0.570~} & \textbf{0.601~} & 0.671~           \\
    printer     & 0.560~ & 0.595~ & 0.648~   & 0.543~ & 0.578~ & 0.640~    & 0.529~ & 0.561~ & 0.619~          & \textbf{0.513~} & \textbf{0.544~} & \textbf{0.617~}  \\
    remote      & 0.510~ & 0.538~ & 0.564~   & 0.479~ & 0.506~ & 0.559~    & 0.468~ & 0.483~ & 0.538~          & \textbf{0.454~} & \textbf{0.483~} & \textbf{0.536~}  \\
    rifle       & 0.534~ & 0.572~ & 0.626~   & 0.519~ & 0.562~ & 0.646~    & 0.500~ & 0.551~ & 0.632~          & \textbf{0.497~} & \textbf{0.539~} & \textbf{0.616~}  \\
    rocket      & 0.512~ & 0.567~ & 0.670~   & 0.511~ & 0.553~ & 0.650~    & 0.503~ & 0.546~ & \textbf{0.610~} & \textbf{0.484~} & \textbf{0.524~} & 0.616~           \\
    skateboard  & 0.517~ & 0.525~ & 0.586~   & 0.484~ & 0.507~ & 0.589~    & 0.461~ & 0.503~ & 0.580~          & \textbf{0.461~} & \textbf{0.483~} & \textbf{0.566~}  \\
    sofa        & 0.554~ & 0.588~ & 0.603~   & 0.532~ & 0.559~ & 0.596~    & 0.525~ & 0.541~ & \textbf{0.563~} & \textbf{0.507~} & \textbf{0.523~} & 0.567~           \\
    stove       & 0.562~ & 0.577~ & 0.650~   & 0.528~ & 0.554~ & 0.634~    & 0.521~ & 0.550~ & \textbf{0.601~} & \textbf{0.516~} & \textbf{0.546~} & 0.606~           \\
    table       & 0.513~ & 0.529~ & 0.588~   & 0.499~ & 0.529~ & 0.579~    & 0.491~ & 0.517~ & 0.576~          & \textbf{0.481~} & \textbf{0.502~} & \textbf{0.553~}  \\
    telephone   & 0.506~ & 0.532~ & 0.564~   & 0.483~ & 0.498~ & 0.554~    & 0.467~ & 0.495~ & 0.529~          & \textbf{0.457~} & \textbf{0.476~} & \textbf{0.516~}  \\
    tower       & 0.571~ & 0.622~ & 0.676~   & 0.536~ & 0.593~ & 0.684~    & 0.513~ & 0.567~ & 0.659~          & \textbf{0.512~} & \textbf{0.562~} & \textbf{0.649~}  \\
    train       & 0.530~ & 0.589~ & 0.660~   & 0.529~ & 0.565~ & 0.629~    & 0.520~ & 0.563~ & 0.618~          & \textbf{0.518~} & \textbf{0.541~} & \textbf{0.599~}  \\
    watercraft  & 0.525~ & 0.551~ & 0.652~   & 0.507~ & 0.547~ & 0.641~    & 0.496~ & 0.524~ & 0.605~          & \textbf{0.488~} & \textbf{0.524~} & \textbf{0.599~}  \\
    washer      & 0.553~ & 0.591~ & 0.649~   & 0.540~ & 0.570~ & 0.647~    & 0.533~ & 0.567~ & 0.621~          & \textbf{0.522~} & \textbf{0.551~} & \textbf{0.612~}  \\ 
    \hline
    mean        & 0.545~ & 0.581~ & 0.650~   & 0.525~ & 0.562~ & 0.637~    & 0.513~ & 0.549~ & 0.612~          & \textbf{0.502~} & \textbf{0.536~} & \textbf{0.608~}  \\
    \hline
    \end{tabular}}
\end{table*}

\begin{table*}
\centering
\caption{Detailed results on the 21 unseen categories of ShapeNet-34 dataset, including Simple (S), Moderate (M) and Hard (H) three difficulties. For CD-$l_2$ ($\times 1000$), lower is better.}
\label{tab.7.}
\begin{tabular}{l|ccc|ccc|ccc|ccc} 
\hline
           & \multicolumn{3}{c|}{GRNet} & \multicolumn{3}{c|}{PoinTr} & \multicolumn{3}{c|}{SeedFormer}                  & \multicolumn{3}{c}{Ours}                          \\
           & CD-S  & CD-M  & CD-H       & CD-S  & CD-M  & CD-H        & CD-S           & CD-M           & CD-H           & CD-S           & CD-M           & CD-H            \\ 
\hline
bag        & 1.47~ & 1.88~ & 3.45~      & 0.96~ & 1.34~ & 2.08~       & 0.49~          & 0.82~          & 1.45~          & \textbf{0.48~} & \textbf{0.80~} & \textbf{1.43~}  \\
basket     & 1.78~ & 1.94~ & 4.18~      & 1.04~ & 1.40~ & 2.90~       & 0.60~          & \textbf{0.85~} & \textbf{1.98~} & \textbf{0.57~} & 0.91~          & 2.08~           \\
birdhouse  & 1.89~ & 2.34~ & 5.16~      & 1.22~ & 1.79~ & 3.45~       & 0.72~          & 1.19~          & 2.31~          & \textbf{0.51~} & \textbf{1.03~} & \textbf{2.00~}  \\
bowl       & 1.77~ & 1.97~ & 3.90~      & 1.05~ & 1.32~ & 2.40~       & \textbf{0.60~} & \textbf{0.77~} & \textbf{1.50~} & 0.63~          & 0.96~          & 1.79~           \\
camera     & 2.31~ & 3.38~ & 7.20~      & 1.63~ & 2.67~ & 4.97~       & 0.89~          & \textbf{1.77~} & \textbf{3.75~} & \textbf{0.88~} & 1.94~          & 4.04~           \\
can        & 1.53~ & 1.80~ & 3.08~      & 0.80~ & 1.17~ & 2.85~       & 0.56~          & 0.89~          & \textbf{1.57~} & \textbf{0.50~} & \textbf{0.64~} & 1.73~           \\
cap        & 3.29~ & 4.87~ & 13.02~     & 1.40~ & 2.74~ & 8.35~       & \textbf{0.50~} & \textbf{1.34~} & \textbf{5.19~} & 0.51~          & 1.57~          & 6.38~           \\
keyboard   & 0.73~ & 0.77~ & 1.11~      & 0.43~ & 0.45~ & 0.63~       & 0.32~          & 0.41~          & 0.60~          & \textbf{0.29~} & \textbf{0.34~} & \textbf{0.59~}  \\
dishwasher & 1.79~ & 1.70~ & 3.27~      & 0.93~ & 1.05~ & 2.04~       & \textbf{0.63~} & \textbf{0.78~} & \textbf{1.44~} & 0.72~          & 0.90~          & 1.49~           \\
earphone   & 4.29~ & 4.16~ & 10.30~     & 2.03~ & 5.10~ & 10.69~      & 1.18~          & \textbf{2.78~} & \textbf{6.71~} & \textbf{1.09~} & 2.96~          & 8.98~           \\
helmet     & 3.06~ & 4.38~ & 10.27~     & 1.86~ & 3.30~ & 6.96~       & \textbf{1.10~} & \textbf{2.27~} & \textbf{4.78~} & 1.32~          & 2.44~          & 5.74~           \\
mailbox    & 1.52~ & 1.90~ & 4.33~      & 1.03~ & 1.47~ & 3.34~       & \textbf{0.56~} & \textbf{0.99~} & \textbf{2.06~} & 0.74~          & 1.09~          & 2.14~           \\
microphone & 2.29~ & 3.23~ & 8.41~      & 1.25~ & 2.27~ & 5.47~       & 0.80~          & \textbf{1.61~} & 4.21~          & \textbf{0.73~} & 1.73~          & \textbf{3.70~}  \\
microwaves & 1.74~ & 1.81~ & 3.82~      & 1.01~ & 1.18~ & 2.14~       & 0.64~          & \textbf{0.83~} & 1.69~          & \textbf{0.60~} & 0.90~          & \textbf{1.58~}  \\
pillow     & 1.43~ & 1.69~ & 3.43~      & 0.92~ & 1.24~ & 2.39~       & \textbf{0.43~} & \textbf{0.66~} & 1.45~          & \textbf{0.43~} & 0.73~          & \textbf{1.07~}  \\
printer    & 1.82~ & 2.41~ & 5.09~      & 1.18~ & 1.76~ & 3.10~       & 0.69~          & \textbf{1.25~} & \textbf{2.33~} & \textbf{0.59~} & 1.40~          & 2.56~           \\
remote     & 0.82~ & 1.02~ & 1.29~      & 0.44~ & 0.58~ & 0.78~       & \textbf{0.27~} & \textbf{0.42~} & \textbf{0.61~} & 0.29~          & 0.51~          & 0.67~           \\
rocket     & 0.97~ & 0.79~ & 1.60~      & 0.39~ & 0.72~ & 1.39~       & 0.28~          & 0.51~          & 1.02~          & \textbf{0.26~} & \textbf{0.46~} & \textbf{0.82~}  \\
skateboard & 0.93~ & 1.07~ & 1.83~      & 0.52~ & 0.80~ & 1.31~       & \textbf{0.35~} & \textbf{0.56~} & 0.92~          & \textbf{0.35~} & 0.61~          & \textbf{0.78~}  \\
tower      & 1.35~ & 1.80~ & 3.85~      & 0.82~ & 1.35~ & 2.48~       & \textbf{0.51~} & 0.92~          & 1.87~          & 0.59~          & \textbf{0.83~} & \textbf{1.79~}  \\
washer     & 1.83~ & 1.97~ & 5.28~      & 1.04~ & 1.39~ & 2.73~       & \textbf{0.61~} & \textbf{0.87~} & 1.94~          & 0.62~          & 0.89~          & \textbf{1.90~}  \\
mean       & 1.84~ & 2.23~ & 4.95~      & 1.05~ & 1.67~ & 3.45~       & 0.61~          & \textbf{1.07~} & \textbf{2.35~} & \textbf{0.60~} & 1.13~          & 2.54~           \\
\hline
\end{tabular}
\end{table*}

\begin{table*}
\centering
\caption{Detailed results on the 21 unseen categories of ShapeNet-34 dataset, including Simple (S), Moderate (M) and Hard (H) three difficulties. For DCD, lower is better.}
\label{tab.8.}
\scalebox{0.85}{
\begin{tabular}{l|ccc|ccc|ccc|ccc} 
\hline
           & \multicolumn{3}{c|}{GRNet} & \multicolumn{3}{c|}{PoinTr} & \multicolumn{3}{c|}{SeedFormer}                     & \multicolumn{3}{c}{Ours}                             \\
           & DCD-S  & DCD-M  & DCD-H    & DCD-S  & DCD-M  & DCD-H     & DCD-S           & DCD-M           & DCD-H           & DCD-S           & DCD-M           & DCD-H            \\ 
\hline
bag        & 0.521~ & 0.597~ & 0.617~   & 0.506~ & 0.573~ & 0.593~    & 0.489~          & \textbf{0.556~} & \textbf{0.572~} & \textbf{0.492~} & \textbf{0.556~} & 0.573~           \\
basket     & 0.599~ & 0.631~ & 0.633~   & 0.574~ & 0.621~ & 0.626~    & 0.555~          & \textbf{0.605~} & \textbf{0.613~} & \textbf{0.548~} & 0.608~          & 0.614~           \\
birdhouse  & 0.610~ & 0.678~ & 0.708~   & 0.580~ & 0.658~ & 0.680~    & 0.561~          & 0.632~          & 0.659~          & \textbf{0.554~} & \textbf{0.615~} & \textbf{0.641~}  \\
bowl       & 0.603~ & 0.637~ & 0.659~   & 0.581~ & 0.624~ & 0.631~    & 0.558~          & \textbf{0.595~} & \textbf{0.614~} & \textbf{0.553~} & 0.597~          & 0.616~           \\
camera     & 0.624~ & 0.640~ & 0.708~   & 0.594~ & 0.629~ & 0.689~    & \textbf{0.574~} & \textbf{0.618~} & 0.666~          & 0.587~          & 0.619~          & \textbf{0.665~}  \\
can        & 0.570~ & 0.628~ & 0.637~   & 0.550~ & 0.617~ & 0.582~    & 0.525~          & 0.587~          & \textbf{0.568~} & \textbf{0.518~} & \textbf{0.571~} & 0.569~           \\
cap        & 0.632~ & 0.737~ & 0.783~   & 0.598~ & 0.715~ & 0.757~    & 0.568~          & \textbf{0.689~} & 0.743~          & \textbf{0.560~} & 0.692~          & \textbf{0.742~}  \\
keyboard   & 0.484~ & 0.513~ & 0.535~   & 0.461~ & 0.501~ & 0.519~    & 0.450~          & 0.484~          & 0.505~          & \textbf{0.446~} & \textbf{0.477~} & \textbf{0.493~}  \\
dishwasher & 0.597~ & 0.631~ & 0.701~   & 0.564~ & 0.609~ & 0.681~    & \textbf{0.554~} & \textbf{0.590~} & 0.658~          & 0.566~          & 0.591~          & \textbf{0.656~}  \\
earphone   & 0.695~ & 0.687~ & 0.803~   & 0.666~ & 0.671~ & 0.773~    & \textbf{0.655~} & \textbf{0.645~} & \textbf{0.739~} & 0.660~          & 0.648~          & 0.751~           \\
helmet     & 0.669~ & 0.730~ & 0.781~   & 0.639~ & 0.717~ & 0.761~    & \textbf{0.627~} & 0.706~          & \textbf{0.738~} & 0.632~          & \textbf{0.699~} & \textbf{0.738~}  \\
mailbox    & 0.563~ & 0.571~ & 0.627~   & 0.547~ & 0.552~ & 0.605~    & 0.526~          & \textbf{0.536~} & 0.582~          & \textbf{0.518~} & 0.541~          & \textbf{0.581~}  \\
microphone & 0.618~ & 0.717~ & 0.764~   & 0.585~ & 0.700~ & 0.746~    & 0.573~          & 0.672~          & \textbf{0.732~} & \textbf{0.566~} & \textbf{0.666~} & 0.734~           \\
microwaves & 0.609~ & 0.622~ & 0.660~   & 0.575~ & 0.604~ & 0.634~    & 0.557~          & \textbf{0.582~} & 0.623~          & \textbf{0.549~} & 0.584~          & \textbf{0.622~}  \\
pillow     & 0.595~ & 0.620~ & 0.626~   & 0.575~ & 0.606~ & 0.603~    & 0.562~          & \textbf{0.577~} & 0.578~          & \textbf{0.556~} & 0.578~          & \textbf{0.568~}  \\
printer    & 0.631~ & 0.675~ & 0.725~   & 0.607~ & 0.665~ & 0.700~    & 0.593~          & \textbf{0.650~} & 0.687~          & \textbf{0.588~} & 0.651~          & \textbf{0.683~}  \\
remote     & 0.496~ & 0.516~ & 0.542~   & 0.480~ & 0.494~ & 0.526~    & 0.462~          & 0.480~          & 0.508~          & \textbf{0.456~} & \textbf{0.470~} & \textbf{0.507~}  \\
rocket     & 0.499~ & 0.510~ & 0.560~   & 0.476~ & 0.501~ & 0.542~    & 0.457~          & 0.488~          & 0.518~          & \textbf{0.449~} & \textbf{0.469~} & \textbf{0.516~}  \\
skateboard & 0.493~ & 0.539~ & 0.607~   & 0.474~ & 0.527~ & 0.580~    & 0.463~          & \textbf{0.511~} & 0.567~          & \textbf{0.456~} & 0.514~          & \textbf{0.566~}  \\
tower      & 0.545~ & 0.602~ & 0.685~   & 0.526~ & 0.590~ & 0.667~    & 0.500~          & 0.563~          & \textbf{0.655~} & \textbf{0.494~} & \textbf{0.544~} & 0.656~           \\
washer     & 0.586~ & 0.612~ & 0.717~   & 0.562~ & 0.591~ & 0.698~    & 0.551~          & \textbf{0.566~} & \textbf{0.686~} & \textbf{0.544~} & 0.569~          & \textbf{0.686~}  \\ 
\hline
mean       & 0.583~ & 0.623~ & 0.670~   & 0.558~ & 0.608~ & 0.647~    & 0.541~          & 0.587~          & 0.629~          & \textbf{0.538~} & \textbf{0.584~} & \textbf{0.627~}  \\
\hline
\end{tabular}}
\end{table*}

\end{appendices}

\end{document}